\documentclass[10pt,twocolumn,letterpaper]{article}
\usepackage{cvpr}
\usepackage{times}
\usepackage{epsfig}
\usepackage{graphicx}
\graphicspath{{Figures/}}
\usepackage{amsmath, amssymb}
\usepackage{relsize}
\usepackage{amssymb}
\usepackage{tabularx}
\usepackage{multirow}
\usepackage{blindtext}
\usepackage{booktabs}
\usepackage{bm}
\usepackage{mmstyles}
\usepackage[section]{placeins}
\usepackage[numbers]{natbib}
\usepackage{diagbox}
\usepackage{caption}
\usepackage{float}
\usepackage{enumitem}
\usepackage{tabularx}
\usepackage{makecell}
\newcolumntype{Y}{>{\centering\arraybackslash}X}

\usepackage{cuted}
\usepackage{capt-of}
\usepackage{flushend}

\renewcommand{\arraystretch}{1.1}

\newcolumntype{L}{>{\centering\arraybackslash}X}
\newcolumntype{C}{>{\centering\arraybackslash}X}

\newcommand{\ignore}[1]{}

\usepackage{xcolor}
\definecolor{todo}{rgb}{1,.5,0} 

\usepackage[pagebackref=true,breaklinks=true,colorlinks,bookmarks=false]{hyperref}
\usepackage[british,UKenglish,USenglish,american]{babel}

\cvprfinalcopy 

\def\cvprPaperID{3032} 

\ifcvprfinal\fi

\begin{document}

\newenvironment{absolutelynopagebreak}
{\par\nobreak\vfil\penalty0\vfilneg
	\vtop\bgroup}
{\par\xdef\tpd{\the\prevdepth}\egroup
	\prevdepth=\tpd}

\title{Cross-domain Correspondence Learning for Exemplar-based Image Translation}
\author{Pan Zhang$^1$
	\thanks{Author did this work during the internship at Microsoft Research Asia.}
	, Bo Zhang$^2$, Dong Chen$^2$, Lu Yuan$^3$, Fang Wen$^2$
	\\
	$^1$University of Science and Technology of China \quad
	$^2$Microsoft Research Asia \quad $^3$ Microsoft Cloud+AI
}

\maketitle
\thispagestyle{empty}

\maketitle
\begin{abstract}
	\vspace{-1em}
    We present a general framework for exemplar-based image translation, which synthesizes a photo-realistic image from the input in a distinct domain (\eg, semantic segmentation mask, or edge map, or pose keypoints), given an exemplar image. The output has the style (\eg, color, texture) in consistency with the semantically corresponding objects in the exemplar.
    We propose to jointly learn the cross-domain correspondence and the image translation, where both tasks facilitate each other and thus can be learned with weak supervision.
    The images from distinct domains are first aligned to an intermediate domain where dense correspondence is established. Then, the network synthesizes images based on the appearance of semantically corresponding patches in the exemplar.
    We demonstrate the effectiveness of our approach in several image translation tasks. Our method is superior to state-of-the-art methods in terms of image quality significantly, with the image style faithful to the exemplar with semantic consistency. Moreover, we show the utility of our method for several  applications.
    
\end{abstract}
\section{Introduction}
\label{sec:introduction}

\begin{figure}[t!]
\centering
\small
\setlength\tabcolsep{0pt}
\includegraphics[width=1.0\columnwidth]{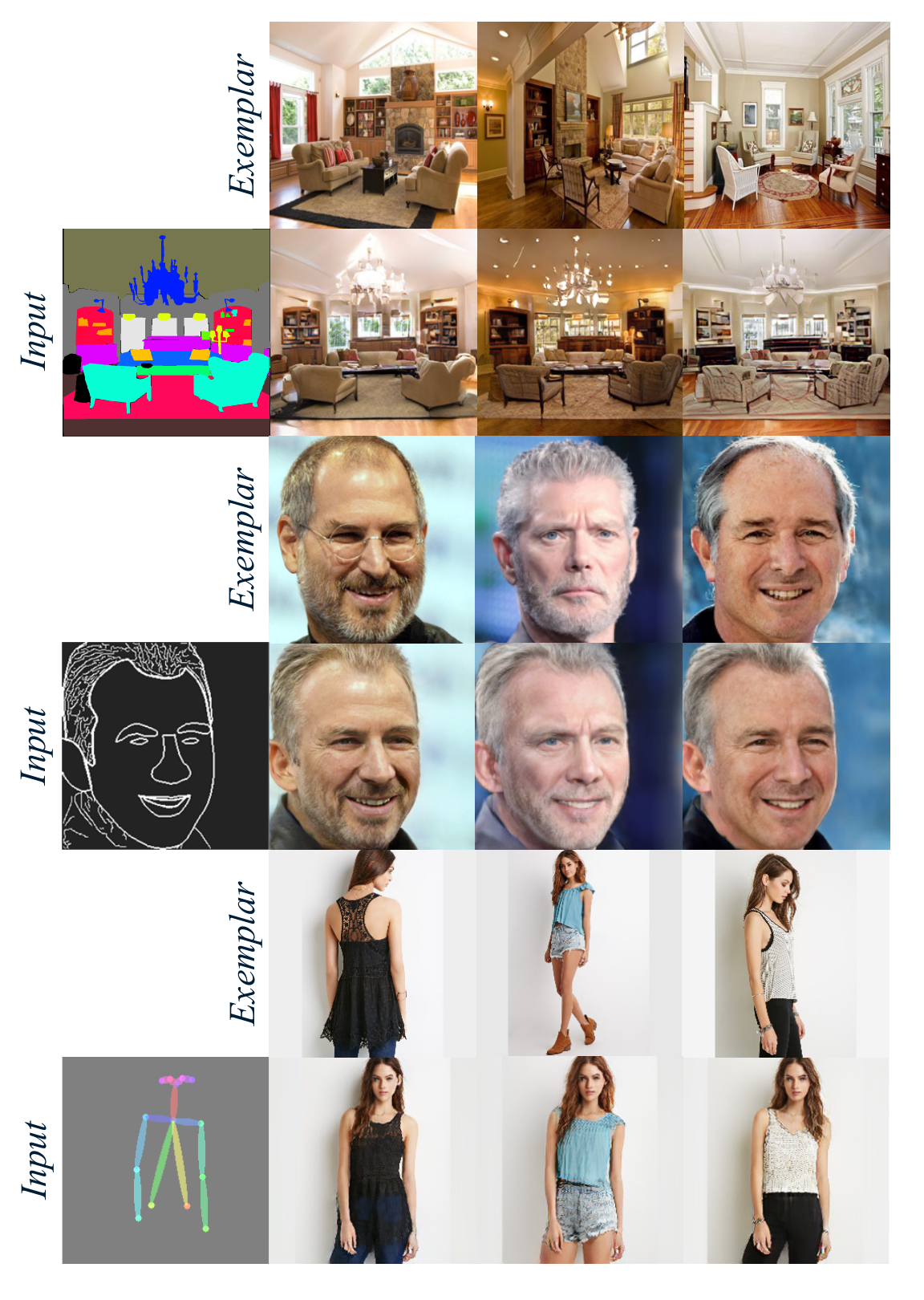}
\caption{\textbf{Exemplar-based image synthesis.} Given the exemplar images (1st row), our network translates the inputs, in the form of segmentation mask, edge and pose, to photo-realistic images (2nd row). Please refer to \emph{supplementary material} for more results.}
\label{fig:teaser1}
\end{figure}

Conditional image synthesis aims to generate photo-realistic images based on certain input data~\cite{isola2017image,wang2018high,zhu2017unpaired,chen2017photographic}. We are interested in a specific form of conditional image synthesis, which converts a semantic segmentation mask, an edge map, and pose keypoints to a photo-realistic image, given an exemplar image, as shown in Figure~\ref{fig:teaser1}. We refer to this form as \emph{exemplar-based image translation}. It allows more flexible control for multi-modal generation according to a user-given exemplar.

Recent methods directly learn the mapping from a semantic segmentation mask to an exemplar image using neural networks~\cite{huang2018multimodal,park2019semantic,ma2018exemplar,wang2019example}. Most of these methods encode the style of the exemplar into a latent style vector, from which the network synthesizes images with the desired style similar to the examplar. However, the style code only characterizes the global style of the exemplar, regardless of spatial relevant information. Thus, it causes some local style ``wash away" in the ultimate image.

To address this issue, the \emph{cross-domain correspondence} between the input and the exemplar has to be established before image translation. As an extension of Image Analogies~\cite{hertzmann2001image}, Deep Analogy~\cite{liao2017visual} attempts to find a dense semantically-meaningful correspondence between the image pair. It leverages deep features of VGG pretrained on real image classification tasks for matching. We argue such representation may fail to handle a more challenging mapping from mask (or edge, keypoints) to photo since the pretrained network does not recognize such images. In order to consider the mask (or edge) in the training, some methods~\cite{gu2019mask,yi2019apdrawinggan,chang2018pairedcyclegan} explicitly separate the exemplar image into semantic regions and learns to synthesize different parts individually. In this way, it successfully generates high-quality results. However, these approaches are task specific, and are unsuitable for general translation.

How to find a more general solution for \emph{exemplar-based image translation} is non-trivial. We aim to learn the dense semantic correspondence for cross-domain images (\eg, mask-to-image, edge-to-image, keypoints-to-image, etc.), and then use it to guide the image translation. It is weakly supervise learning, since we have neither the correspondence annotations nor the synthesis ground truth given a random exemplar.

In this paper, we propose a \emph{{C}r{O}ss-domain {C}{O}rre{S}pondence network} (\emph{CoCosNet}) that learns cross-domain correspondence and image translation simultaneously. The network architecture comprises two sub-networks: 1) \emph{Cross-domain correspondence Network} transforms the inputs from distinct domains to an intermediate feature domain where reliable dense correspondence can be established; 2) \emph{Translation network}, employs a set of spatially-variant de-normalization blocks~\cite{park2019semantic} to progressively synthesizes the output, using the style details from a warped exemplar which is semantically aligned to the mask (or edge, keypoints map) according to the estimated correspondence. Two sub-networks facilitate each other and are learned end-to-end with novel loss functions. Our method outperforms previous methods in terms of image quality by a large margin, with instance-level appearance being faithful to the exemplar. Moreover, the cross-domain correspondence implicitly learned enables some intriguing applications, such as image editing and makeup transfer. Our contribution can be summarized as follows:
\begin{itemize}[leftmargin=*]
\itemsep0em
\item We address the problem of learning dense cross-domain correspondence with weak supervision---joint learning with image translation.
\item With the cross-domain correspondence, we present a general solution to exemplar-based image translation, that for the first time, outputs images resembling the fine structures of the exemplar at instance level.
\item Our method outperforms state-of-the-art methods in terms of image quality by a large margin in various application tasks.
\end{itemize}

\section{Related Work}
\label{sec:related_work}

\noindent\textbf{Image-to-image translation}~~
The goal of image translation is to learn the mapping between different image domains. Most prominent contemporary approaches solve this problem through conditional generative adversarial network~\cite{mirza2014conditional} that leverages either paired data~\cite{isola2017image,wang2018high,park2019semantic} or unpaired data~\cite{zhu2017unpaired,yi2017dualgan,kim2017learning,liu2017unsupervised,royer2017xgan}. Since the mapping from one image domain to another is inherently multi-modal, following works promote the synthesis diversity by performing stochastic sampling from the latent space~\cite{zhu2017toward,huang2018multimodal,lee2018diverse}. However, none of these methods allow delicate control of the output since the latent representation is rather complex and does not have an explicit correspondence to image style. In contrast, our method supports customization of the result according to a user-given exemplar, which allows more flexible control for multi-modal generation.

\noindent\textbf{Exemplar-based image synthesis}~~
Very recently, a few works~\cite{qi2018semi,wang2019example,ma2018exemplar,riviere2019inspirational,bansal2019shapes} propose to synthesize photorealistic images from semantic layout under the guidance of exemplars. Non-parametric or semi-parametric approaches~\cite{qi2018semi,bansal2019shapes} synthesize images by compositing the image fragments retrieved from a large database. 
Mainstream works, however, formulate the  problem as image-to-image translation. \citet{huang2018multimodal} and \citet{ma2018exemplar} propose to employ Adaptive Instance Normalization (AdaIN)~\cite{huang2017arbitrary} to transfer the style code from the exemplar to the source image. \citet{park2019semantic} learn an encoder to map the exemplar image into a vector from which the images are further synthesized. The style consistency discriminator is proposed in~\cite{wang2019example} to examine whether the image pairs exhibit a similar style. However, this method requires to constitute style consistency image pairs from video clips, which makes it unsuitable for general image translation. Unlike all of the above methods that only transfer the global style, our method transfers the fine style from a semantically corresponding region of the exemplar. Our work is inspired by the recent exemplar-based image colorization~\cite{zhang2019deep,he2018deep}, but we solve a more general problem: translating images between distinct domains.

\noindent\textbf{Semantic correspondence}~~
Early studies~\cite{lowe2004distinctive,dalal2005histograms,tola2009daisy} on semantic correspondence focus on matching hand-crafted features. With the advent of the convolutional neural network, deep features are proven powerful to represent the high-level semantics. \citet{long2014convnets} first propose to establish semantic correspondence by matching deep features extracted from a pretrained classification model. Following works further improve the correspondence quality by incorporating additional annotations~\cite{zhou2016learning,choy2016universal,ham2017proposal,han2017scnet,kim2017fcss,lee2019sfnet}, adopting coarse-to-fine strategy~\cite{liao2017visual} or retaining reliable sparse matchings~\cite{aberman2018neural}.
However, all these methods can only handle the correspondence between natural images instead of cross-domain images, \eg, edge and photorealistic images. We explore this new scenario and implicitly learns the task with weak supervision.

\begin{figure*}[t!]
	\centering
	\includegraphics[width=1.0\linewidth]{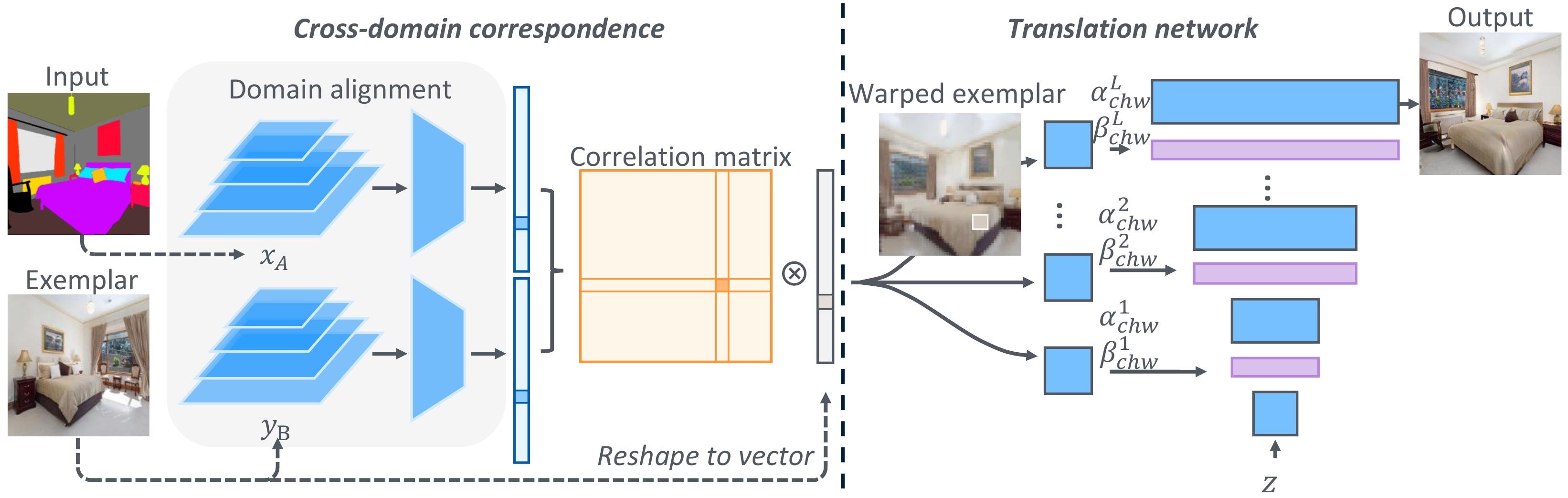}
	\caption{\textbf{The illustration of the \emph{CoCosNet} architecture.} Given the input $x_A\in A$ and the exemplar $y_B \in B$, the correspondence submodule adapts them into the same domain $S$, where dense correspondence can be established. Then, the translation network generates the final output based on the warped exemplar $r_{y\to x}$ according to the correspondence, yielding an exemplar-based translation output.}
	\label{fig:overview}
\end{figure*}
\section{Approach}
\label{sec:method}

We aim to learn the translation from the source domain $\cA$ to the target domain $B$ given an input image $x_A \in A$ and an exemplar image $y_B \in B$. The generated output is desired to conform to the content as $x_A$ while resembling the style from semantically similar parts in $y_B$. For this purpose, the correspondence between $x_A$ and $y_B$, which lie in different domains, is first established, and the exemplar image is warped accordingly so that its semantics is aligned with $x_A$ (Section~\ref{sec:cross_domain}). Thereafter, an image is synthesized according to the warped exemplar (Section~\ref{sec:translation}). The whole network architecture is illustrated in Figure~\ref{fig:overview}, by the example of mask to image synthesis.

\subsection{Cross-domain correspondence network}
\label{sec:cross_domain}
Usually the semantic correspondence is found by matching patches~\cite{liao2017visual,lee2019sfnet} in the feature domain with a pre-trained classification model. However, pre-trained models are typically trained on a specific type of images, \eg, natural images, so the extracted features cannot generalize to depict the semantics for another domain. Hence, prior works cannot establish the correspondence between heterogeneous images, \eg, edge and photo-realistic images. To tackle this, we propose a novel cross-domain correspondence network, mapping the input domains to a shared domain $S$ in which the representation is capable to represent the semantics for both input domains. As a result, reliable semantic correspondence can be found within domain $S$.

\vspace{0.4em}
\noindent\textbf{Domain alignment} As shown in Figure~\ref{fig:overview}, we first adapt the input image and the exemplar to a shared domain $S$. To be specific, $x_A$ and $y_B$ are fed into the feature pyramid network that extracts multi-scale deep features by leveraging both local and global image context~\cite{ronneberger2015u,lin2017feature}. The extracted feature maps are further transformed to the representations in $S$, denoted by $x_S\in \Rbb^{HW\times C}$ and $y_S\in \Rbb^{HW \times C}$ respectively ($H$,$W$ are feature spatial size; $C$ is the channel-wise dimension). Let $\cF_{A\to S} $ and $\cF_{B\to S}$ be the domain transformation from the two input domains respectively, so the adapted representation can be formulated as,
\begin{align}
    x_S &= \cF_{A\to S}(x_A;\theta_{\cF,A\to S}), \\ 
    y_S &= \cF_{B\to S}(y_B;\theta_{\cF,B\to S}).
    \label{eq:domain_adaptation}
\end{align}
where $\theta$ denotes the learnable parameter. The representation $x_S$ and $y_S$ comprise discriminative features that characterize the semantics of inputs. Domain alignment is, in practice, essential for correspondence in that only when $x_S$ and $y_S$ reside in the same domain can they be further matched with some similarity measure. 
 
\vspace{0.4em}
\noindent\textbf{Correspondence within shared domain}
We propose to match the features of $x_S$ and $y_S$ with the correspondence layer proposed in~\cite{zhang2019deep}. Concretely, we compute a correlation matrix $\cM\in \Rbb^{HW\times HW}$ of which each element is a pairwise feature correlation,
\begin{equation}
    \cM(u, v) = \frac{\hat{x}_S(u)^T\hat{y}_S(v)}{\Vert{\hat{x}_S(u)} \Vert  \ \Vert{\hat{y}_S(v)}\Vert},
    \label{eq:correlation_matrix}
\end{equation}
where $\hat{x}_S(u)$ and $\hat{y}_S(v) \in \Rbb^{C}$ represent the channel-wise centralized feature of $x_S$ and $y_S$ in position $u$ and $v$, \ie, $\hat{x}_S(u) = x_S(u) - \text{mean}(x_S(u))$ and $\hat{y}_S(v) = y_S(v) - \text{mean}(y_S(v))$.
$\cM(u,v)$ indicates a higher semantic similarity between $x_S(u)$ and $y_S(v)$. 

Now the challenge is how to learn the correspondence without direct supervision. Our idea is to jointly train with image translation. The translation network may find it easier to generate high-quality outputs only by referring to the correct corresponding regions in the exemplar, which implicitly pushes the network to learn the accurate correspondence. In light of this, we warp $y_B$ according to $\cM$ and obtain the warped exemplar $r_{y\to x}\in \Rbb^{HW}$. Specifically, we obtain $r_{y\to x}$ by selecting the most correlated pixels in $y_B$ and calculating their weighted average,
\begin{equation}
    r_{y\to x}(u) = \sum_v \softmax_v (\alpha \cM(u,v)) \cdot y_B(v).
    \label{eq:warping}
\end{equation}
Here, $\alpha$ is the coefficient that controls the sharpness of the softmax
and we set its default value as $100$. In the following, images will be synthesized conditioned on $r_{y\to x}$ and the correspondence network, in this way, learns its assignment with indirect supervision.

\subsection{Translation network}
\label{sec:translation}
Under the guidance of $r_{y\to x}$, the translation network $\cG$ transforms the constant code $z$ to the desired output $\hat{x}_B \in B$. In order to preserve the structural information of $r_{y\to x}$, we employ the spatially-adaptive denormalization (SPADE) block~\cite{park2019semantic} to project the spatially variant exemplar style to different activation locations. As shown in Figure~\ref{fig:overview}, the translation network has $L$ layers with the exemplar style progressively injected. As opposed to~\cite{park2019semantic} which computes layer-wise statistics for batch normalization (BN), we empirically find the normalization that computes the statistics at each spatial position, the positional normalization (PN)~\cite{2019arXiv190704312L}, better preserves the structure information synthesized in prior layers. Hence, we propose to marry positional normalization and spatially-variant denormalization for high-fidelity texture transfer from the exemplar. 

Formally, given the activation $F^i\in \Rbb^{C_i \times H_i \times W_i}$ before the $i^{th}$ normalization layer, we inject the exemplar style through,
\begin{equation}
    \alpha^i_{h,w}(r_{y\to x}) \times \frac{F^i_{c,h,w} - \mu^i_{h,w}}{\sigma^i_{h,w}} + \beta^i_{h,w}(r_{y\to x}),
    \label{eq:normalization}
\end{equation}
where the statistic value $\mu^i_{h,w}$ and $\sigma^i_{h,w}$ are calculated exclusively across channel direction compared to BN. The denormalization parameter $\alpha^i$ and $\beta^i$ characterize the style of the exemplar, which is mapped from $r_{y\to x}$ with the projection $\cT$ parameterized by $\theta_\mathcal{T}$, \ie,
\begin{equation}
    \alpha^i,\beta^i  = \cT_i(r_{y\to x}; \theta_{\mathcal{T}}).
\end{equation}
We use two plain convolutional layers to implement $\cT$ so $\alpha$ and $\beta$ have the same spatial size as $r_{y\to x}$. With the style modulation for each normalization layer, the overall image translation can be formulated as
\begin{align}
    \hat{x}_B 
    = \cG(z,\cT_i(r_{y\to x};\theta_{\mathcal{T}});\theta_{\mathcal{G}}),
\end{align}
where $\theta_{\mathcal{G}}$ denotes the learnable parameter.

\subsection{Losses for exemplar-based translation}
We jointly train the cross-domain correspondence along with image synthesis with following loss functions, hoping the two tasks benefit each other. 

\vspace{0.4em}
\noindent\textbf{Losses for pseudo exemplar pairs}~~We construct exemplar training pairs by utilizing paired data $\{x_A,x_B\}$ that are semantically aligned but differ in domains.
Specifically, we apply random geometric distortion to $x_B$ and get the distorted image $x'_B=h(x_B)$, where $h$ denotes the augmentation operation like image warping or random flip. When $x'_B$ is regarded as the exemplar, the translation of $x_A$ is expected to be its counterpart $x_B$. In this way, we obtain pseudo exemplar pairs. We propose to penalize the difference between the translation output and the ground truth $x_B$ by minimizing the \emph{feature matching loss}~\cite{johnson2016perceptual,isola2017image,chen2017photographic}
\begin{equation}
    \cL_{feat} = \sum_l \lambda_{l}\norm{\phi_l(\cG(x_A,x'_B))-\phi_l(x_B)}_1,
\end{equation}
where $\phi_l$ represents the activation of layer $l$ in the pre-trained VGG-19 model and $\lambda_l$ balance the terms.

\vspace{0.4em}
\noindent\textbf{Domain alignment loss}~We need to make sure the transformed embedding $x_S$ and $y_S$ lie in the same domain. To achieve this, we once again make use of the image pair $\{x_A,x_B\}$, whose feature embedding should be aligned exactly after domain transformation: 
\begin{equation}
\cL_{domain}^{\ell_1} = \norm{\cF_{A\to S}(x_A)-\cF_{B\to S}(x_B)}_1.
\end{equation}
Note that we perform channel-wise normalization as the last layer of $\cF_{A\to S}$ and $\cF_{B\to S}$ so minimizing this domain discrepancy will not lead to a trivial solution (\ie, small magnitude of activations).  

\vspace{0.4em}
\noindent\textbf{Exemplar translation losses}~~The learning with pair or pseudo exemplar pair is hard to generalize to general cases where the semantic layout of exemplar differs significantly from the source image. To tackle this, we propose the following losses.

First, the ultimate output should be consistent with the semantics of the input $x_A$, or its counterpart $x_B$. We thereby penalize the \emph{perceptual loss} to minimize the semantic discrepancy:
\begin{equation}
    \cL_{perc} = \norm{\phi_l(\hat{x}_B)-\phi_l(x_B)}_1.
\end{equation}
Here we choose $\phi_l$ to be the activation after $relu$4\_2 layer in the VGG-19 network since this layer mainly contains high-level semantics. 

On the other hand, we need a loss function that encourages $\hat{x}_B$ to adopt the appearance from the semantically corresponding patches from $y_B$. To this end, we employ the \emph{contextual loss} proposed in ~\cite{mechrez2018contextual} to match the statistics between $\hat{x}_B$ and $y_B$, which is
\begin{equation}
    \thinmuskip=1mu
    \thickmuskip=1mu
\begin{split}
    &\cL_{context} =\\ 
    &\sum_l \omega_{l}\left[-\log\left(\frac{1}{n_l}\sum_i \max_j A^l(\phi^l_i(\hat{x}_B),\phi^l_j(y_B))\right) \right],
\end{split}
\end{equation}
where $i$ and $j$ index the feature map of layer $\phi^l$ that contains $n_l$ features, and $\omega_{l}$ controls the relative importance of different layers. Still, we rely on pretrained VGG features. As opposed to $\cL_{perc}$ which mainly utilizes high-level features, the contextual loss uses $relu$2\_2 up to $relu$5\_2 layers since low-level features capture richer style information (\eg, color or textures) useful for transferring the exemplar appearance.

\vspace{0.4em}
\noindent\textbf{Correspondence regularization}~~Besides, the learned correspondence should be cycle consistent, \ie, the image should match itself after forward-backward warping, which is
\begin{equation}
    \cL_{reg} = \norm{r_{y\to x \to y}-y_B}_1,
\end{equation}  
where $r_{y\to x \to y}(v) = \sum_u \softmax_u (\alpha \cM(u,v)) \cdot r_{y\to x}(u)$ is the forward-backward warping image. Indeed, this objective function is crucial because the rest loss functions, imposed at the end of the network, are weak supervision and cannot guarantee that the network learns a meaningful correspondence. Figure~\ref{fig:ablation} shows that without $\cL_{reg}$ the network fails to learn the cross-domain correspondence correctly although it is still capable to generate plausible translation result. The regularization $\cL_{reg}$ enforces the warped image $r_{y\to x}$ remain in domain $B$ by constraining its backward warping, implicitly encouraging the correspondence to be meaningful as desired. 

\vspace{0.4em}
\noindent\textbf{Adversarial loss}~~We train a discriminator~\cite{goodfellow2014generative} that discriminates the translation outputs and the real samples of domain $B$. Both the discriminator $\cD$ and the translation network $\cG$ are trained alternatively until synthesized images look indistinguishable to real ones. The adversarial objectives of $\cD$ and $\cG$ are respectively defined as:
\begin{equation}
\begin{split}
\cL^{\cD}_{adv} &= -\Ebb [h(\cD(y_B))] - \Ebb [h(-\cD(\cG(x_A,y_B)))]\\
\cL^{\cG}_{adv} &=  -\Ebb [\cD(\cG(x_A,y_B))],
\end{split}
\label{eq:hinge_loss}
\end{equation}
where $h(t)=\min(0,-1+t)$ is a hinge function used to regularize the discriminator~\cite{zhang2018self,brock2018large}. 

\vspace{0.4em}
\noindent\textbf{Total loss}~~In all, we optimize the following objective,
\begin{equation}
\thickmuskip=1mu
\begin{split}
    \cL_\theta =\ \min_{\cF,\cT,\cG} &\max_{\cD}\psi_1 \cL_{feat} + \psi_2\cL_{perc} + \psi_3\cL_{context} \\ &+ \psi_4\cL^{\cG}_{adv} + \psi_5\cL^{\ell_1}_{domain}+ \psi_6\cL_{reg},
\end{split}
\end{equation}
where weights $\psi$ are used  to balance the objectives.

\section{Experiments}
\label{sec:experiments}
\begin{figure*}[t]
    \center
    \small
    \setlength\tabcolsep{1pt}
    {
    \renewcommand{\arraystretch}{0.6}
    \begin{tabular}{cccccccc}
        Input & Ground truth & Pix2pixHD~\cite{wang2018high} & MUINT~\cite{huang2018multimodal} & EGSC-IT~\cite{ma2018exemplar} & SPADE~\cite{park2019semantic} & Ours & Exemplar\\[0.1ex]
        \includegraphics[width=0.25\columnwidth]{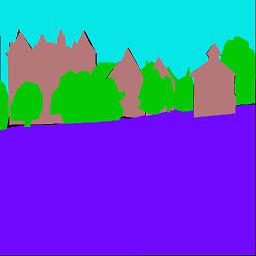} &
        \includegraphics[width=0.25\columnwidth]{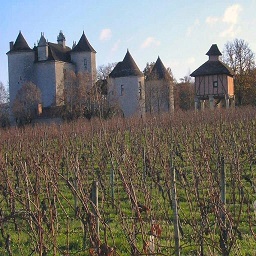} &
        \includegraphics[width=0.25\columnwidth]{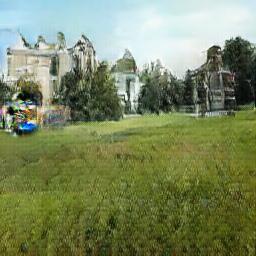} &
        \includegraphics[width=0.25\columnwidth]{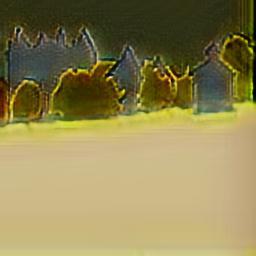} &
        \includegraphics[width=0.25\columnwidth]{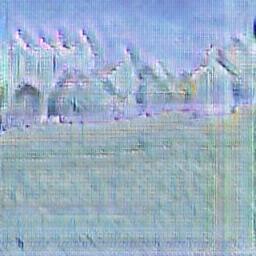} &
        \includegraphics[width=0.25\columnwidth]{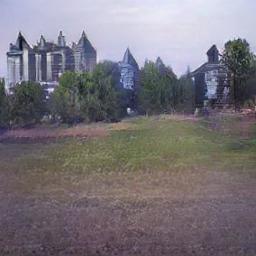} &
        \includegraphics[width=0.25\columnwidth]{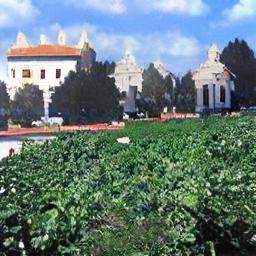} &
        \includegraphics[width=0.25\columnwidth]{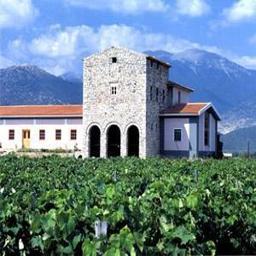}\\

        \includegraphics[width=0.25\columnwidth]{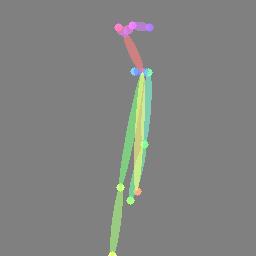} &
        \includegraphics[width=0.25\columnwidth]{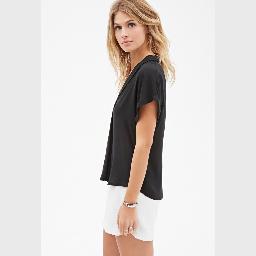} &
        \includegraphics[width=0.25\columnwidth]{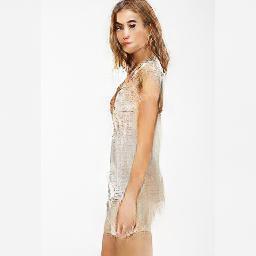} &
        \includegraphics[width=0.25\columnwidth]{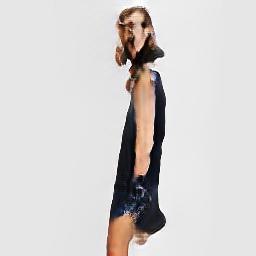} &
        \includegraphics[width=0.25\columnwidth]{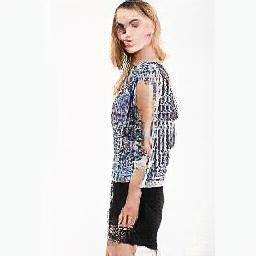} &
        \includegraphics[width=0.25\columnwidth]{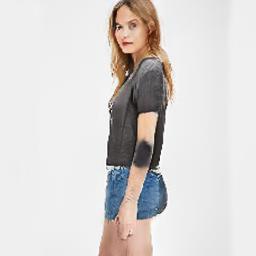} &
        \includegraphics[width=0.25\columnwidth]{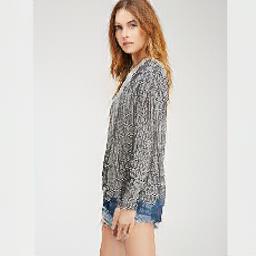} &
        \includegraphics[width=0.25\columnwidth]{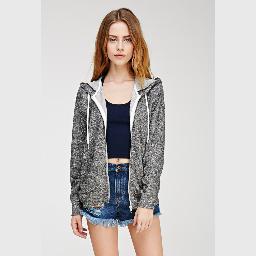}\\
        
        \includegraphics[width=0.25\columnwidth]{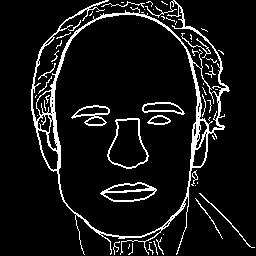} &
        \includegraphics[width=0.25\columnwidth]{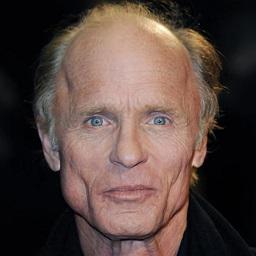} &
        \includegraphics[width=0.25\columnwidth]{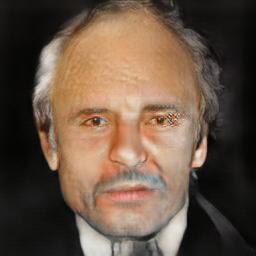} &
        \includegraphics[width=0.25\columnwidth]{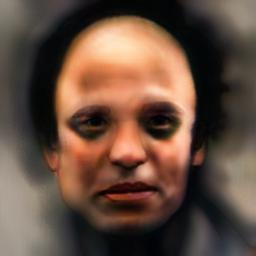} &
        \includegraphics[width=0.25\columnwidth]{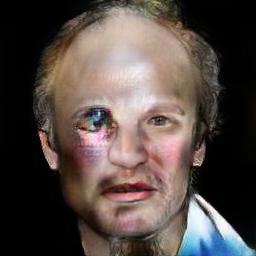} &
        \includegraphics[width=0.25\columnwidth]{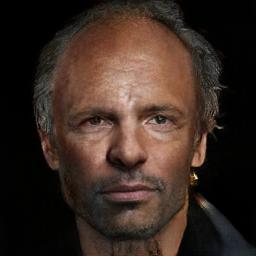} &
        \includegraphics[width=0.25\columnwidth]{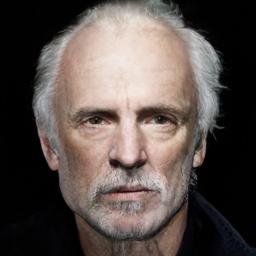} &
        \includegraphics[width=0.25\columnwidth]{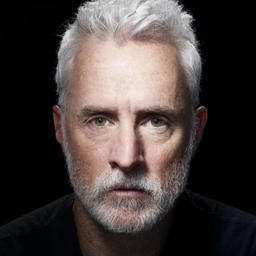}\\
    \end{tabular}
    }
    \caption{\textbf{
    Qualitative comparison of different methods.}}
    \label{fig:comparison}
\end{figure*}

\begin{figure*}[t]
\center
\small
\setlength\tabcolsep{0pt}
{
\renewcommand{\arraystretch}{0.0}
\begin{tabular}{@{}ccccccccccc@{}}
    Input & Synthesis & Exemplar & ~ & Input & Synthesis & Exemplar & ~ & Input & Synthesis & Exemplar\\[0.5ex]
    \includegraphics[width=0.23\columnwidth]{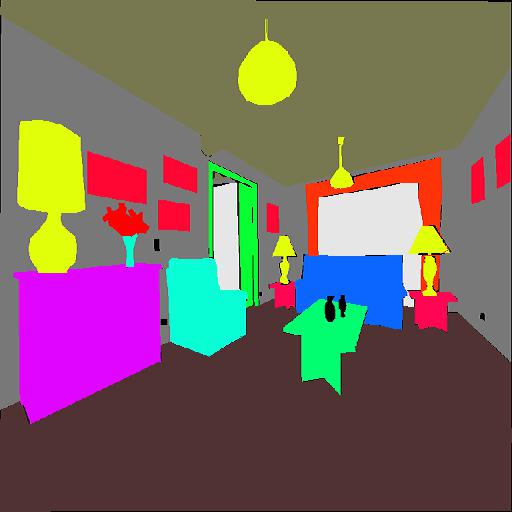}&
    \includegraphics[width=0.23\columnwidth]{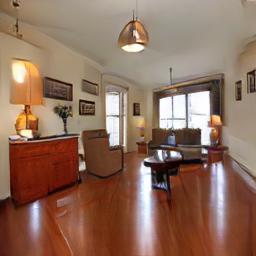}&
    \includegraphics[width=0.23\columnwidth]{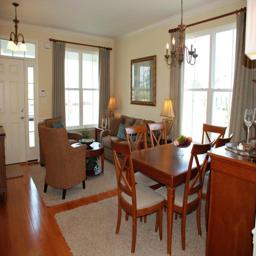}& ~ &
    \includegraphics[width=0.23\columnwidth]{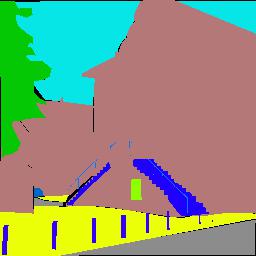}& \includegraphics[width=0.23\columnwidth]{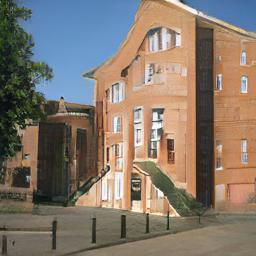} & 
    \includegraphics[width=0.23\columnwidth]{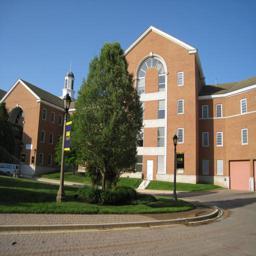}& ~ &
    \includegraphics[width=0.23\columnwidth]{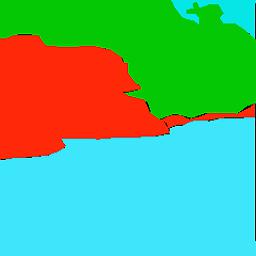}&
    \includegraphics[width=0.23\columnwidth]{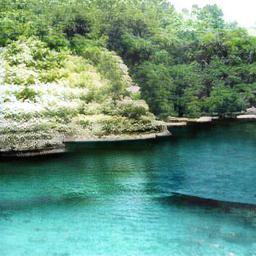}&
    \includegraphics[width=0.23\columnwidth]{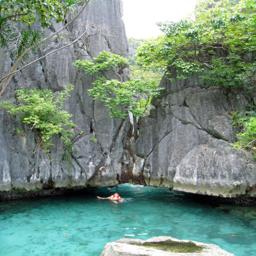}\\
    \includegraphics[width=0.23\columnwidth]{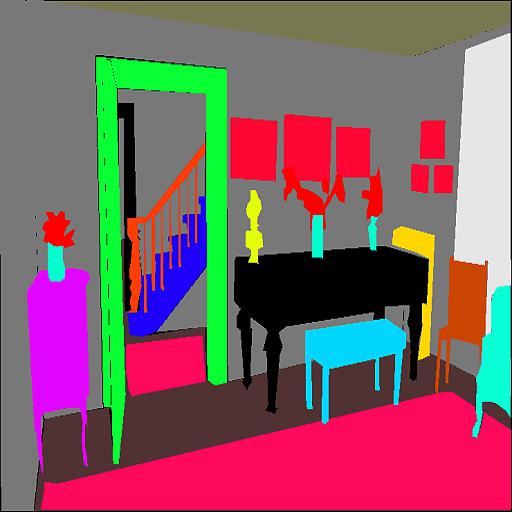}&
    \includegraphics[width=0.23\columnwidth]{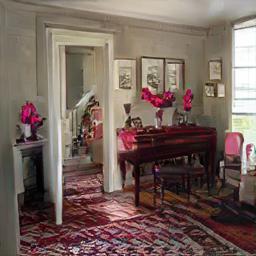}&
    \includegraphics[width=0.23\columnwidth]{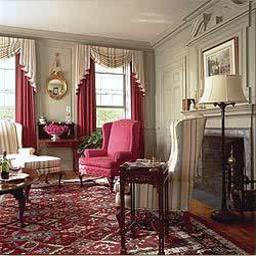}& ~ & 
    \includegraphics[width=0.23\columnwidth]{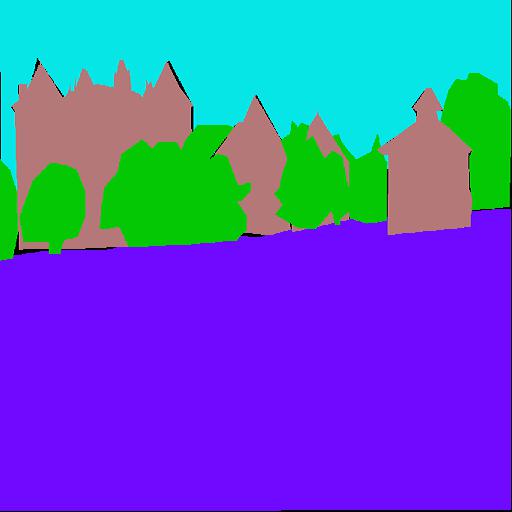}&
    \includegraphics[width=0.23\columnwidth]{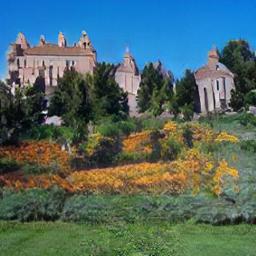}& 
    \includegraphics[width=0.23\columnwidth]{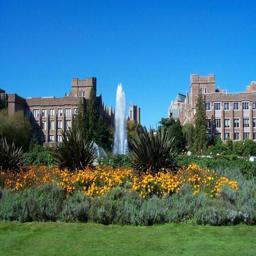}& ~ &
    \includegraphics[width=0.23\columnwidth]{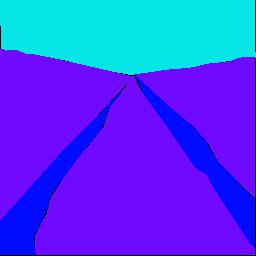}&
    \includegraphics[width=0.23\columnwidth]{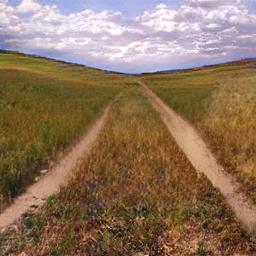}&
    \includegraphics[width=0.23\columnwidth]{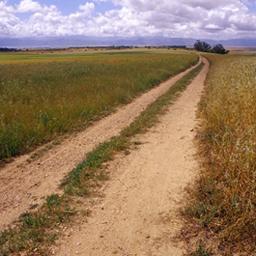}\\
\end{tabular}
}
\caption{\textbf{Our results of segmentation mask to image synthesis (ADE20k dataset).}}
\label{figure:ade20k_results}
\end{figure*}

\begin{figure*}[t]
\center
\small
\setlength\tabcolsep{0pt}
{
\renewcommand{\arraystretch}{0.0}
\begin{tabular}{@{}rrccccccc@{}}
    &
    \raisebox{0.25\height}{\rotatebox{90}{Exemplars}}&
    \includegraphics[width=0.25\columnwidth]{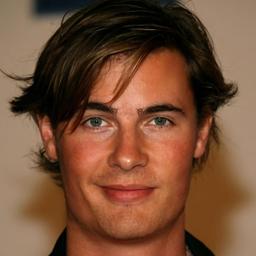}&
    \includegraphics[width=0.25\columnwidth]{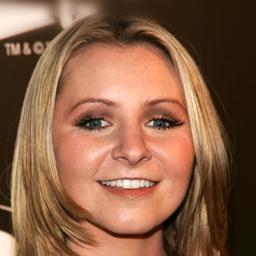}&
    \includegraphics[width=0.25\columnwidth]{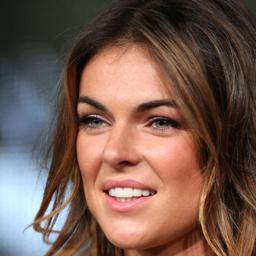}&
    \includegraphics[width=0.25\columnwidth]{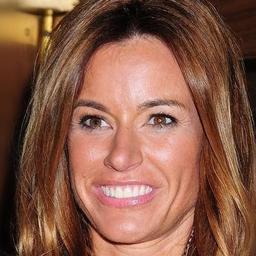}&
    \includegraphics[width=0.25\columnwidth]{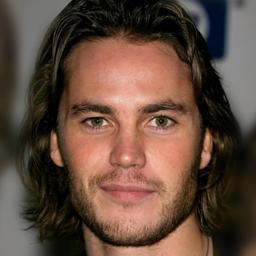}&
    \includegraphics[width=0.25\columnwidth]{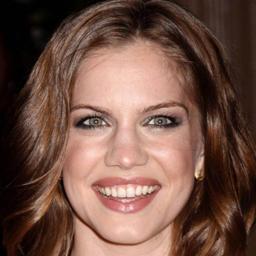}&
    \includegraphics[width=0.25\columnwidth]{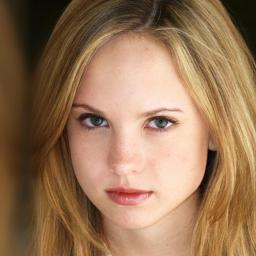}\\
    

    \raisebox{1.05\height}{\rotatebox{90}{Edge}}&
    \includegraphics[width=0.25\columnwidth]{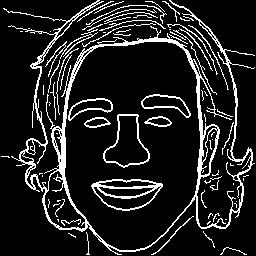}&
    \includegraphics[width=0.25\columnwidth]{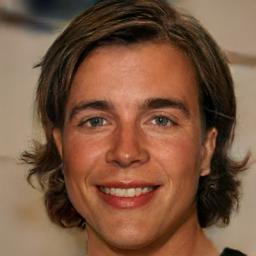}&
    \includegraphics[width=0.25\columnwidth]{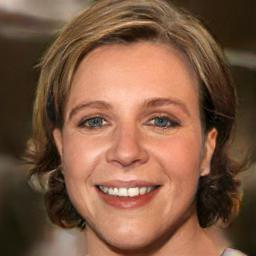}&
    \includegraphics[width=0.25\columnwidth]{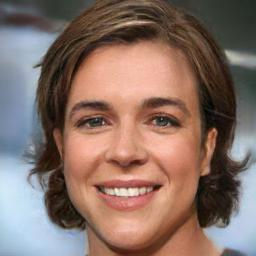}&
    \includegraphics[width=0.25\columnwidth]{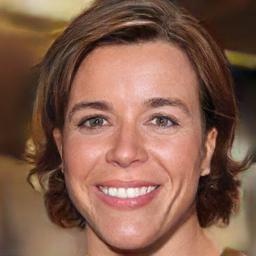}&
    \includegraphics[width=0.25\columnwidth]{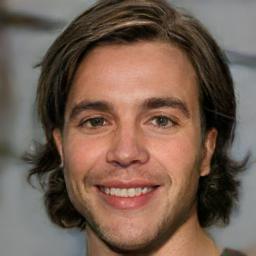}&
    \includegraphics[width=0.25\columnwidth]{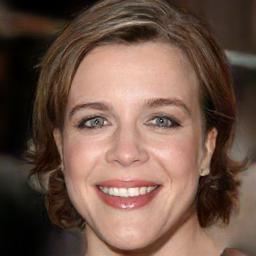}&
    \includegraphics[width=0.25\columnwidth]{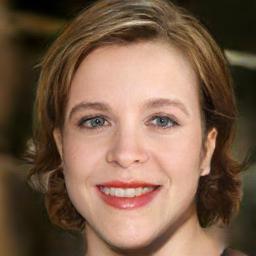}\\
\end{tabular}
}
\caption{\textbf{Our results of edge to face synthesis (CelebA-HQ dataset).} First row: exemplars. Second row: our results.}
\label{figure:cebeba_result}
\end{figure*}

\begin{figure}[t]
    \center
    \small
    \setlength\tabcolsep{0pt}
    {
    \renewcommand{\arraystretch}{0.0}
    \begin{tabular}{cccccccc}
        & 
        \includegraphics[width=0.12\columnwidth]{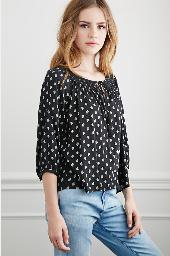} &
        \includegraphics[width=0.12\columnwidth]{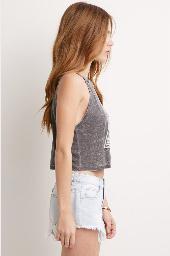} &
        \includegraphics[width=0.12\columnwidth]{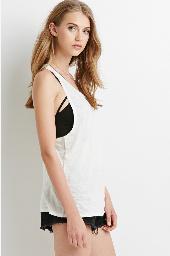}&
        &
        \includegraphics[width=0.12\columnwidth]{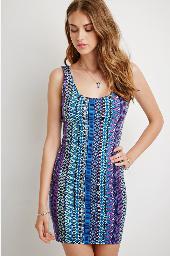} &
        \includegraphics[width=0.12\columnwidth]{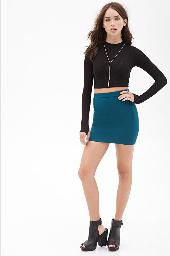} &
        \includegraphics[width=0.12\columnwidth]{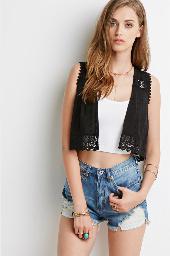}\\
        
        \includegraphics[width=0.12\columnwidth]{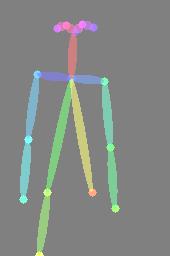} &
        \includegraphics[width=0.12\columnwidth]{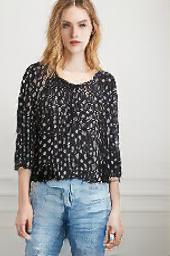} &
        \includegraphics[width=0.12\columnwidth]{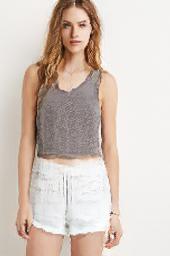} &
        \includegraphics[width=0.12\columnwidth]{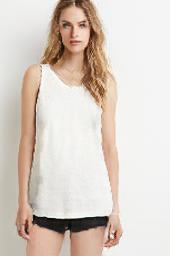} &
        \includegraphics[width=0.12\columnwidth]{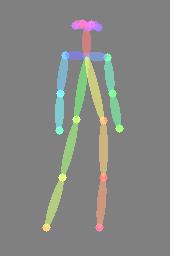} &
        \includegraphics[width=0.12\columnwidth]{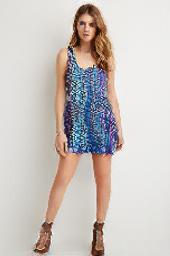} &
        \includegraphics[width=0.12\columnwidth]{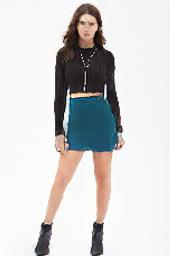} &
        \includegraphics[width=0.12\columnwidth]{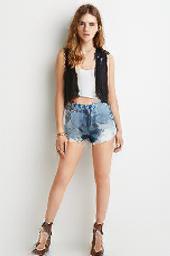}\\
    \end{tabular}
    }
    \caption{\textbf{Our results of pose to body synthesis (DeepFashion).} First row: exemplars. Second row: our results.}
    \label{figure:pose}
\end{figure}

\begin{table}[!t]
    \footnotesize
    \centering 
    \setlength\tabcolsep{2.5pt}
    \caption{\textbf{Image quality comparison.} Lower FID or SWD score indicates better image quality. The best scores are highlighted.}
    \begin{tabular}{@{}lcccccccc@{}}
    \toprule
    \multirow{2}{*}{} & \multicolumn{2}{c}{ADE20k} & \multicolumn{2}{c}{ADE20k-\emph{outdoor}} & \multicolumn{2}{c}{CelebA-HQ} & \multicolumn{2}{c}{DeepFashion}\\
    \cmidrule(lr){2-3} 
    \cmidrule(lr){4-5} 
    \cmidrule(lr){6-7} 
    \cmidrule(lr){8-9} 
     & FID & SWD & FID & SWD & FID & SWD & FID & SWD\\ \midrule
     Pix2pixHD & 81.8 & 35.7 & 97.8 & 34.5 & 62.7 & 43.3 & 25.2 & \textbf{16.4}\\
     SPADE & 33.9 & 19.7 & 63.3 & 21.9 & 31.5 & 26.9 & 36.2 & 27.8\\
     MUNIT & 129.3 & 97.8 & 168.2 & 126.3 & 56.8 & 40.8 & 74.0 & 46.2\\
     SIMS & N/A & N/A & 67.7 & 27.2 & N/A & N/A & N/A & N/A \\
     EGSC-IT & 168.3 & 94.4 & 210.0 & 104.9 & 29.5 & 23.8  & 29.0 & 39.1\\
     \emph{Ours} & \textbf{26.4} & \textbf{10.5} & \textbf{42.4} & \textbf{11.5} & \textbf{14.3} & \textbf{15.2} & \textbf{14.4} & 17.2\\
     \bottomrule
    \end{tabular}
    \label{table:image_quality}
\end{table}

\noindent\textbf{Implementation}~~We use Adam~\cite{kingma2014adam} solver with $\beta_1=0, \beta_2=0.999$. Following the TTUR~\cite{heusel2017gans}, we set imbalanced learning rates, $1e{-4}$ and $4e{-4}$ respectively, for the generator and discriminator. Spectral normalization~\cite{miyato2018spectral} is applied to all the layers for both networks to stabilize the adversarial training. Readers can refer to the supplementary material for detailed network architecture. We conduct experiments using 8 32GB Tesla V100 GPUs, and it takes roughly 4 days to train 100 epochs on the ADE20k dataset~\cite{zhou2017scene}.   

\noindent\textbf{Datasets}~~We conduct experiments on multiple datasets with different sorts of image representation. All the images are resized to 256$\times$256 during training. 
\begin{itemize}[leftmargin=*]
\itemsep0em 
\item {ADE20k}~\cite{zhou2017scene} consists of $\sim$20k training images, each image associated with a 150-class segmentation mask. This is a challenging dataset for most existing methods due to its large diversity.
\item {ADE20k-\emph{outdoor}} contains the outdoor images extracted from ADE20k, as the same protocol in SIMS~\cite{qi2018semi}.
\item {CelebA-HQ}~\cite{liu2015faceattributes} contains high quality face images. We connect the face landmarks for face region, and use Canny edge detector to detect edges in the background. We perform an edge-to-face translation on this dataset. 
\item {Deepfashion}~\cite{liuLQWTcvpr16DeepFashion} consists of 52,712 person images in fashion clothes. We extract the pose keypoints using the  OpenPose~\cite{cao2018openpose}, and learn the translation to human body.
\end{itemize}
 
\begin{table}[!t]
    \footnotesize
    \setlength\tabcolsep{4.5pt}
    \centering 
    \caption{\textbf{Comparison of semantic consistency.} The best scores are highlighted.} 
    \begin{tabular}{@{}lcccc@{}}
    \toprule
    & ADE20k & ADE20k-\emph{outdoor} & CelebA-HQ & DeepFashion\\
    \midrule
    Pix2pixHD & 0.833 & 0.848 & 0.914 & 0.943\\
    SPADE     & 0.856 & 0.867 & 0.922 & 0.936\\
    MUNIT     & 0.723 & 0.704 & 0.848 & 0.910\\
    SIMS      & N/A & 0.822 & N/A & N/A\\
    EGSC-IT   & 0.734 & 0.723 & 0.915 & 0.942\\
    \emph{Ours} & \textbf{0.862} & \textbf{0.873} & \textbf{0.949} & \textbf{0.968}\\
    \bottomrule
    \end{tabular}
    \label{table:semantic_consistency}
\end{table}
\begin{table}[!t]
    \footnotesize
    \centering 
    \setlength\tabcolsep{0.9pt}
    \caption{\textbf{Comparison of style relevance.} A higher score indicates a higher appearance similarity relative to the exemplar. The best scores are highlighted.}
    \begin{tabularx}{\columnwidth}{@{}lYYYYYY@{}}
    \toprule
    \multirow{2}{*}{} & \multicolumn{2}{c}{ADE20k} & \multicolumn{2}{c}{CelebA-HQ} & \multicolumn{2}{c}{DeepFashion}\\
    \cmidrule(lr){2-3}
    \cmidrule(lr){4-5}
    \cmidrule(lr){6-7}
    & Color & Texture & Color & Texture & Color & Texture\\ 
    \midrule
    SPADE     & 0.874 & 0.892 & 0.955 & 0.927 & 0.943 & 0.904\\
    MUNIT     & 0.745 & 0.782 & 0.939 & 0.884 & 0.893 & 0.861\\
    EGSC-IT   & 0.781 & 0.839 & 0.965 & 0.942 & 0.945 & 0.916\\
    \emph{Ours} &  \textbf{0.962} & \textbf{0.941} & \textbf{0.977} & \textbf{0.958} & \textbf{0.982} & \textbf{0.958}\\
    \bottomrule
    \end{tabularx}
    \label{table:exemplar_relevance}
\end{table}

\begin{figure}[t!]
    \begin{center}
    \includegraphics[width=1.0\linewidth]{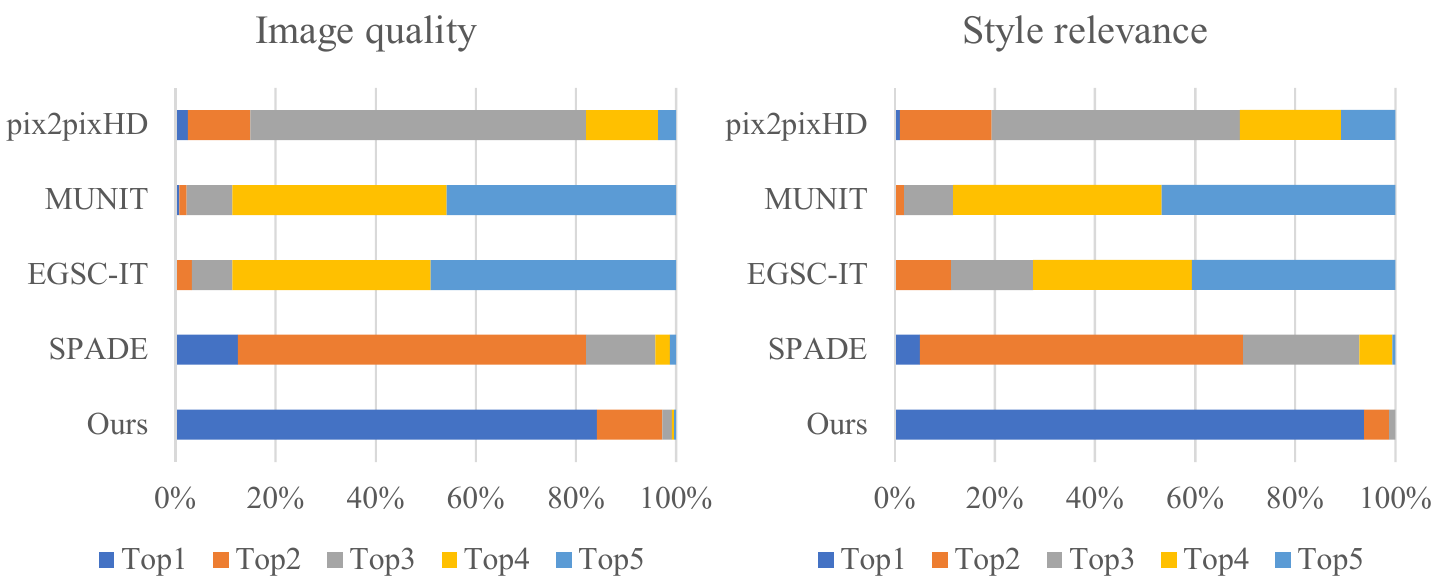}
    \end{center}
    \caption{\textbf{User study results.}}
    \label{figure:user_study}
\end{figure}

\noindent\textbf{Baselines}~~We compare our method with state-of-the-art image translation methods: 1) Pix2pixHD~\cite{wang2018high}, a leading supervised approach; 2) SPADE~\cite{park2019semantic}, a recently proposed supervised translation method which also supports the style injection from an exemplar image; 3) MUNIT~\cite{huang2018multimodal}, an unsupervised method that produces multi-modal results; 4) SIMS~\cite{qi2018semi}, which synthesizes images by compositing image segments from a memory bank; 5) EGSC-IT~\cite{ma2018exemplar}, an exemplar-based method that also considers the semantic consistency but can only mimic the global style. These methods except Pix2pixHD can generate exemplar-based results, and we use their released codes in this mode to train on several datasets. Since it is computationally prohibitive to prepare a database using SIMS, we directly use their reported figures. As we aim to propose a general translation framework, we do not include other task-specific methods. To provide the exemplar for our method, we first train a plain translation network to generate natural images and use them to retrieve the exemplars from the dataset. 

\vspace{0.4em}
\noindent\textbf{Quantitative evaluation}~~We evaluate different methods from three aspects. 
\begin{itemize}[leftmargin=*]
    \itemsep0em
    \item We use two metrics to measure image quality. First, we use the Fr\'echet Inception Score (FID)~\cite{heusel2017gans} to measure the distance between the distributions of synthesized images and real images. While FID measures the semantic realism, we also adopt sliced Wasserstein distance (SWD)~\cite{karras2017progressive} to measure their statistical distance of low-level patch distributions. Measured by these two metrics, Table~\ref{table:image_quality} shows that our method significantly outperforms prior methods in almost all the comparisons. Our method improves the FID score by 7.5 compared to previous leading methods on the challenging ADE20k dataset.
    \item The ultimate output should not alter the input semantics. To evaluate the semantic consistency, we adopt an ImageNet pretrained VGG model~\cite{brock2018large}, and use its high-level features maps, $relu3\_2$, $relu4\_2$ and $relu5\_2$, to represent high-level semantics. We calculate the cosine similarity for these layers and take the average to yield the final score. Table~\ref{table:semantic_consistency} shows that our method best maintains the semantics during translation.
    \item Style relevance. We use low level features $relu1\_2$ and $relu2\_2$ respectively to measure the color and texture distance between the semantically corresponding patches in the output and the exemplar. We do not include Pix2pixHD as it does not produce an exemplar-based translation. Still, our method achieves considerably better instance-level style relevance as shown in Table~\ref{table:exemplar_relevance}.
\end{itemize}

        
        

\noindent\textbf{Qualitative comparison}~~Figure~\ref{fig:comparison} provides a qualitative comparison of different methods. It shows that our \emph{CocosNet} demonstrates the most visually appealing quality with much fewer artifacts. Meanwhile, compared to prior exemplar-based methods, our method demonstrates the best style fidelity, with the fine structures matching the semantically corresponding regions of the exemplar. This also correlates with the quantitative results, showing the obvious advantage of our approach. We show diverse results by changing the exemplar image in Figure~\ref{figure:ade20k_results}-\ref{figure:pose}. Please refer to the supplementary material for more results.
    
    
\begin{figure}[!t]
\center
\small
\setlength\tabcolsep{0pt}
{
\renewcommand{\arraystretch}{0.0}
\begin{tabular}{cccc}
    \includegraphics[width=0.25\columnwidth]{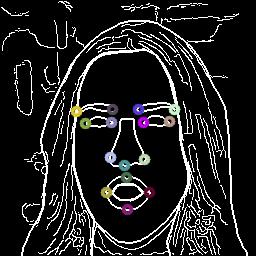} & 
    \includegraphics[width=0.25\columnwidth]{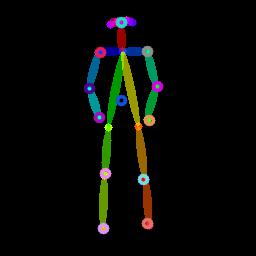}&
    \includegraphics[width=0.25\columnwidth]{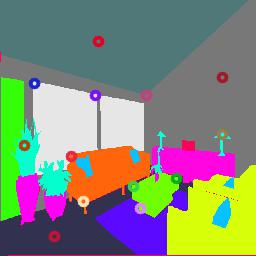}&
    \includegraphics[width=0.25\columnwidth]{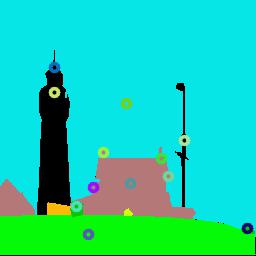}
    \\
    \includegraphics[width=0.25\columnwidth]{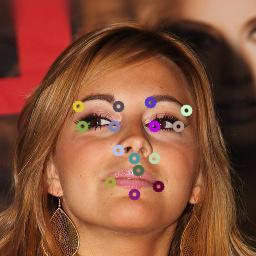} &
    \includegraphics[width=0.25\columnwidth]{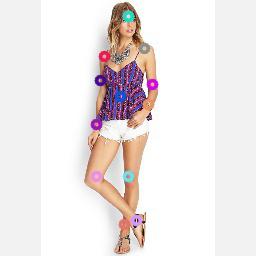}&
    \includegraphics[width=0.25\columnwidth]{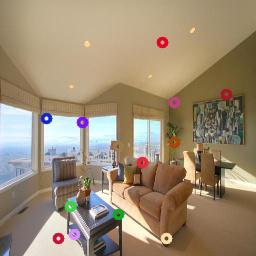} &
    \includegraphics[width=0.25\columnwidth]{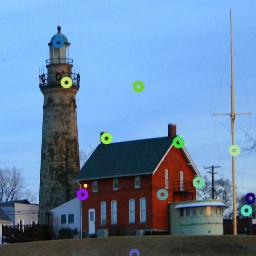}\\
\end{tabular}
}
\caption{\textbf{Sparse correspondence of different domains.} Given the manual annotation points in domain A (first row), our method finds their corresponding points in domain B (second row).}
\label{fig:correspondence}
\end{figure}

\begin{table}[!t]
    \footnotesize
    \setlength\tabcolsep{4.5pt}
    \centering 
    \caption{\textbf{Ablation study.}} 
    \begin{tabular}{@{}lccc@{}}
    \toprule
    & FID $\downarrow$ & Semantic consistency $\uparrow$ & Style (color/texture) $\uparrow$ \\
    \midrule
    w/o $\cL_{feat}$ & 14.4 & 0.948 & 0.975 / 0.955\\
    w/o $\cL_{domain}$ & 21.1 & 0.933 & \textbf{0.983} / 0.957\\
    w/o $\cL_{perc}$& 59.3 & 0.852 & 0.971 / 0.852\\
    w/o $\cL_{context}$& 28.4 & 0.931 & 0.954 / 0.948\\
    w/o $\cL_{reg}$& 19.3 & 0.929 & 0.981 / 0.951\\
    \emph{Full} & \textbf{14.3} & \textbf{0.949} & 0.977 / \textbf{0.958} \\
    \bottomrule
    \end{tabular}
    \label{table:ablation}
\end{table}

\begin{figure}[!t]
\center
\small
\setlength\tabcolsep{1pt}
{
\renewcommand{\arraystretch}{0.6}
\begin{tabular}{lccc}
    & Input/Exemplar  & Dense warping & Final output\\[0.1ex]
    
    \raisebox{3.5\height}{w/o $\cL_{domain}^{\ell_1}$} & \includegraphics[width=0.22\columnwidth]{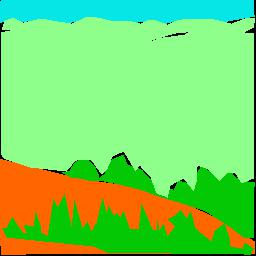} &
    \includegraphics[width=0.22\columnwidth]{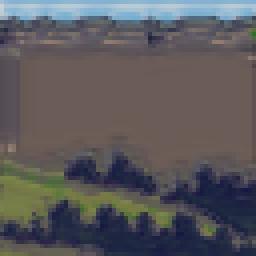} &
    \includegraphics[width=0.22\columnwidth]{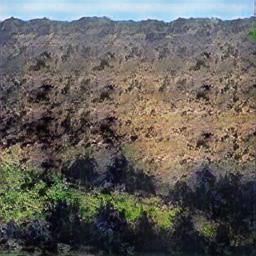}\\

    \raisebox{3.5\height}{w/ $\cL_{domain}^{\ell_1}$}  &
    \includegraphics[width=0.22\columnwidth]{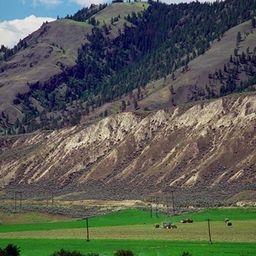} &
    \includegraphics[width=0.22\columnwidth]{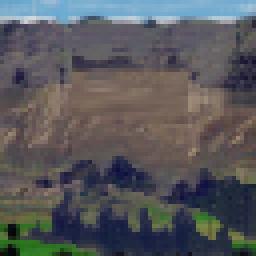} &
    \includegraphics[width=0.22\columnwidth]{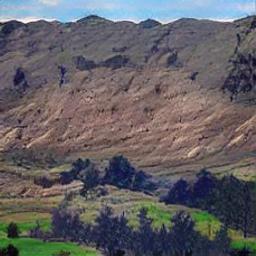} \\
    \midrule
    \raisebox{3.5\height}{w/o $\cL_{reg}$}  &
    \includegraphics[width=0.22\columnwidth]{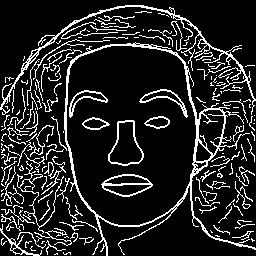} &
    \includegraphics[width=0.22\columnwidth]{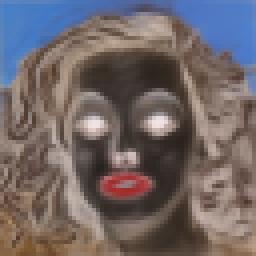} &
    \includegraphics[width=0.22\columnwidth]{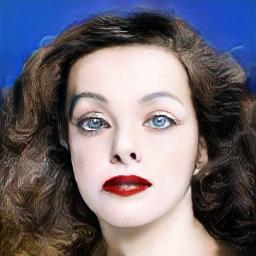}\\

    \raisebox{3.5\height}{w/ $\cL_{reg}$}  &
    \includegraphics[width=0.22\columnwidth]{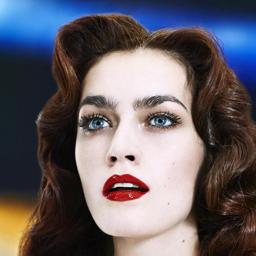} &
    \includegraphics[width=0.22\columnwidth]{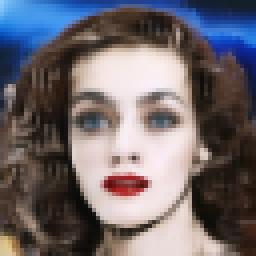} &
    \includegraphics[width=0.22\columnwidth]{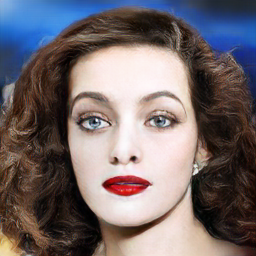} \\
\end{tabular}
}
\caption{\textbf{Ablation study of loss functions.}}
\label{fig:ablation}
\end{figure}

\noindent\textbf{Subjective evaluation}~~
We also conduct user study to compare the subjective quality. We randomly select 10 images for each task, yielding 30 images in total for comparison. We design two tasks, and let users sort all the methods in terms of the image quality and the style relevance. Figure~\ref{figure:user_study} shows the results, where our method demonstrates a clear advantage. Our method ranks the first in 84.2\% cases in evaluating the image quality, with 93.8\% chance to be the best in the style relevance comparison.


\noindent\textbf{Cross-domain correspondence}~~
Figure~\ref{fig:correspondence} shows the cross-domain correspondence. For better visualization, we just annotate the sparse points. As the first approach in doing this, our \emph{CoCosNet} successfully establishes meaningful semantic correspondence which is even difficult for manual labeling. The network is still capable to find the correspondence for sparse representation such as edge map, which captures little explicit semantic information.

\noindent\textbf{Ablation study}~~In order to validate the effectiveness of each component, we conduct comprehensive ablation studies. Here we want to emphasize two key elements (Figure~\ref{fig:ablation}). First, the domain alignment loss $\cL_{domain}^{\ell_1}$ with data pairs $x_A$ and $x_B$ is crucial. Without it, the correspondence will fail in unaligned domains, leading to oversmooth dense warping. We also ablate the correspondence regularization loss $\cL_{reg}$, which leads to incorrect dense correspondence, \eg, face to hair in Figure~\ref{fig:ablation}, though the network still yields plausible final output. With $\cL_{reg}$, the correspondence becomes meaningful, which facilitates the image synthesis as well. We also quantitatively measures the role of different losses in Table~\ref{table:ablation} where the full model demonstrates the best performance in terms of all the metrics.  

\section{Applications}
\label{sec:application}
\begin{figure}[!t]
    \center
    \small
    \setlength\tabcolsep{1pt}
    {
    \renewcommand{\arraystretch}{0.6}
    \begin{tabular}{cccc}
        \includegraphics[width=0.25\columnwidth]{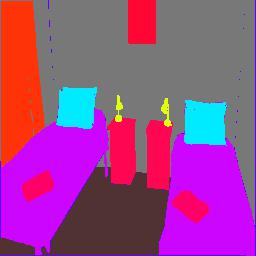} &
        \includegraphics[width=0.25\columnwidth]{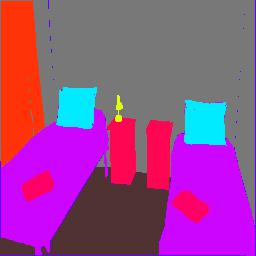} &
        \includegraphics[width=0.25\columnwidth]{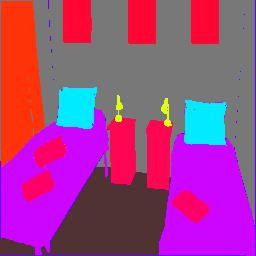} &
        \includegraphics[width=0.25\columnwidth]{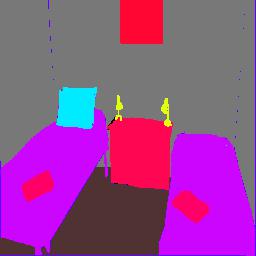}\\
        
        \includegraphics[width=0.25\columnwidth]{Figures/Application/ADE_val_00000156.jpg} &
        \includegraphics[width=0.25\columnwidth]{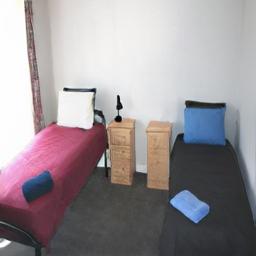} &
        \includegraphics[width=0.25\columnwidth]{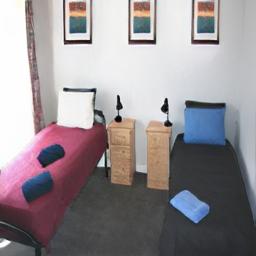} &
        \includegraphics[width=0.25\columnwidth]{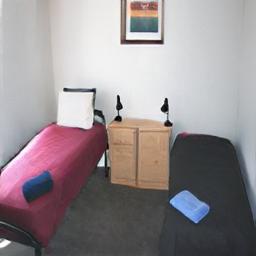}\\
    \end{tabular}
    }
    \caption{\textbf{Image editing.} Given the input image and its mask (1st column), we can semantically edit the image content through the manipulation on the mask (column 2-4).}
    \label{figure:editing}
    \end{figure}

\begin{figure}[!t]
    \center
    \small
    \setlength\tabcolsep{1pt}
    {
    \renewcommand{\arraystretch}{0.6}
    \begin{tabular}{cccc}
        \includegraphics[width=0.25\columnwidth]{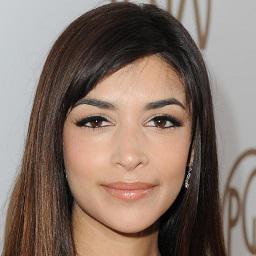} &
        \includegraphics[width=0.25\columnwidth]{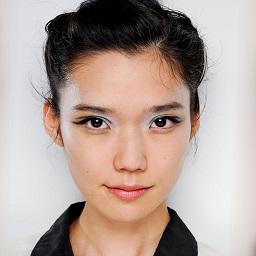} &
        \includegraphics[width=0.25\columnwidth]{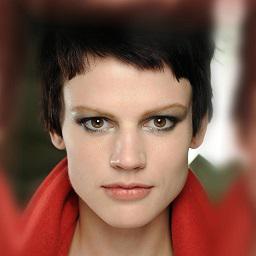} &
        \includegraphics[width=0.25\columnwidth]{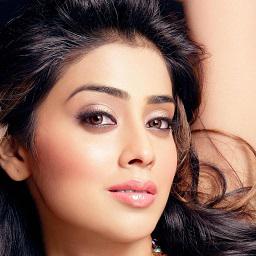}\\
        
        \includegraphics[width=0.25\columnwidth]{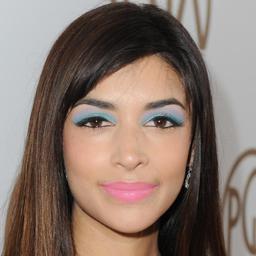} &
        \includegraphics[width=0.25\columnwidth]{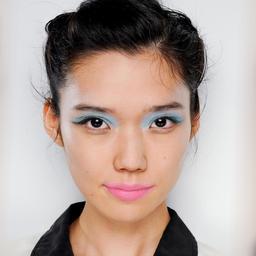} &
        \includegraphics[width=0.25\columnwidth]{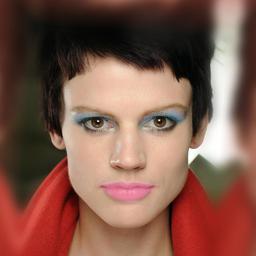} &
        \includegraphics[width=0.25\columnwidth]{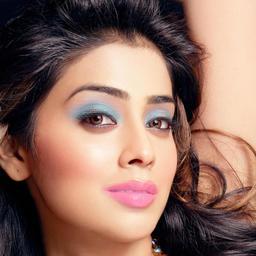}\\
    \end{tabular}
    }
    \caption{\textbf{Makeup transfer.} Given a portrait and makeup strokes (1st column), we can transfer these makeup edits to other portraits by matching the semantic correspondence. We show more examples in the supplementary material.}
    \label{figure:makeup}
\end{figure}

Our method can enable a few intriguing applications. Here we give two examples. 

\noindent\textbf{Image editing}~~
Given a natural image, we can manipulate its content by modifying the segmentation layout and synthesizing the image by using the original image as the self-exemplar. Since this is similar to the pseudo exemplar pairs we constitute for training, our \emph{CocosNet} could perfectly handle it and produce the output with high quality. Figure~\ref{figure:editing} illustrates the image editing, where one can move, add and delete instances.   

\noindent\textbf{Makeup transfer}~~
Artists usually manually adds digital makeup on portraits. Because of the dense semantic correspondence we find, we can transfer the artistic stokes to other portraits. In this way, one can manually add makeup edits on one portrait, and use our network to process a large batch of portraits automatically based on the semantic correspondence, which illustrated in Figure~\ref{figure:makeup}.


\section{Conclusion}
\label{sec:conclusion}
In this paper, we present the \emph{CocosNet}, which translates the image by relying on the cross-domain correspondence. Our method achieves preferable performance than leading approaches both quantitatively and qualitatively. Besides, our method learns the dense correspondence for cross-domain images, paving a way for several intriguing applications. Our method is computationally intensive and we leave high-resolution synthesis to future work.


    


\cleardoublepage
{\small
	\bibliographystyle{IEEEtranSN}
	\bibliography{egbib}

\begin{thebibliography}{54}
\providecommand{\natexlab}[1]{#1}
\providecommand{\url}[1]{#1}
\csname url@samestyle\endcsname
\providecommand{\newblock}{\relax}
\providecommand{\bibinfo}[2]{#2}
\providecommand{\BIBentrySTDinterwordspacing}{\spaceskip=0pt\relax}
\providecommand{\BIBentryALTinterwordstretchfactor}{4}
\providecommand{\BIBentryALTinterwordspacing}{\spaceskip=\fontdimen2\font plus
\BIBentryALTinterwordstretchfactor\fontdimen3\font minus
  \fontdimen4\font\relax}
\providecommand{\BIBforeignlanguage}[2]{{%
\expandafter\ifx\csname l@#1\endcsname\relax
\typeout{** WARNING: IEEEtranSN.bst: No hyphenation pattern has been}%
\typeout{** loaded for the language `#1'. Using the pattern for}%
\typeout{** the default language instead.}%
\else
\language=\csname l@#1\endcsname
\fi
#2}}
\providecommand{\BIBdecl}{\relax}
\BIBdecl

\bibitem[Aberman et~al.(2018)Aberman, Liao, Shi, Lischinski, Chen, and
  Cohen-Or]{aberman2018neural}
K.~Aberman, J.~Liao, M.~Shi, D.~Lischinski, B.~Chen, and D.~Cohen-Or, ``Neural
  best-buddies: Sparse cross-domain correspondence,'' \emph{ACM Transactions on
  Graphics (TOG)}, vol.~37, no.~4, p.~69, 2018.

\bibitem[Bansal et~al.(2019)Bansal, Sheikh, and Ramanan]{bansal2019shapes}
A.~Bansal, Y.~Sheikh, and D.~Ramanan, ``Shapes and context: In-the-wild image
  synthesis \& manipulation,'' in \emph{Proceedings of the IEEE Conference on
  Computer Vision and Pattern Recognition}, 2019, pp. 2317--2326.

\bibitem[Brock et~al.(2018)Brock, Donahue, and Simonyan]{brock2018large}
A.~Brock, J.~Donahue, and K.~Simonyan, ``Large scale {GAN} training for high
  fidelity natural image synthesis,'' \emph{arXiv preprint arXiv:1809.11096},
  2018.

\bibitem[Cao et~al.(2018)Cao, Hidalgo, Simon, Wei, and Sheikh]{cao2018openpose}
Z.~Cao, G.~Hidalgo, T.~Simon, S.-E. Wei, and Y.~Sheikh, ``Openpose: realtime
  multi-person 2d pose estimation using part affinity fields,'' \emph{arXiv
  preprint arXiv:1812.08008}, 2018.

\bibitem[Chang et~al.(2018)Chang, Lu, Yu, and
  Finkelstein]{chang2018pairedcyclegan}
H.~Chang, J.~Lu, F.~Yu, and A.~Finkelstein, ``Pairedcyclegan: Asymmetric style
  transfer for applying and removing makeup,'' in \emph{Proceedings of the IEEE
  Conference on Computer Vision and Pattern Recognition}, 2018, pp. 40--48.

\bibitem[Chen and Koltun(2017)]{chen2017photographic}
Q.~Chen and V.~Koltun, ``Photographic image synthesis with cascaded refinement
  networks,'' in \emph{Proceedings of the IEEE International Conference on
  Computer Vision}, 2017, pp. 1511--1520.

\bibitem[Choy et~al.(2016)Choy, Gwak, Savarese, and
  Chandraker]{choy2016universal}
C.~B. Choy, J.~Gwak, S.~Savarese, and M.~Chandraker, ``Universal correspondence
  network,'' in \emph{Advances in Neural Information Processing Systems}, 2016,
  pp. 2414--2422.

\bibitem[Dalal and Triggs(2005)]{dalal2005histograms}
N.~Dalal and B.~Triggs, ``Histograms of oriented gradients for human
  detection,'' 2005.

\bibitem[Goodfellow et~al.(2014)Goodfellow, Pouget-Abadie, Mirza, Xu,
  Warde-Farley, Ozair, Courville, and Bengio]{goodfellow2014generative}
I.~Goodfellow, J.~Pouget-Abadie, M.~Mirza, B.~Xu, D.~Warde-Farley, S.~Ozair,
  A.~Courville, and Y.~Bengio, ``Generative adversarial nets,'' in
  \emph{Advances in neural information processing systems}, 2014, pp.
  2672--2680.

\bibitem[Gu et~al.(2019)Gu, Bao, Yang, Chen, Wen, and Yuan]{gu2019mask}
S.~Gu, J.~Bao, H.~Yang, D.~Chen, F.~Wen, and L.~Yuan, ``Mask-guided portrait
  editing with conditional gans,'' in \emph{Proceedings of the IEEE Conference
  on Computer Vision and Pattern Recognition}, 2019, pp. 3436--3445.

\bibitem[Ham et~al.(2017)Ham, Cho, Schmid, and Ponce]{ham2017proposal}
B.~Ham, M.~Cho, C.~Schmid, and J.~Ponce, ``Proposal flow: Semantic
  correspondences from object proposals,'' \emph{IEEE transactions on pattern
  analysis and machine intelligence}, vol.~40, no.~7, pp. 1711--1725, 2017.

\bibitem[Han et~al.(2017)Han, Rezende, Ham, Wong, Cho, Schmid, and
  Ponce]{han2017scnet}
K.~Han, R.~S. Rezende, B.~Ham, K.-Y.~K. Wong, M.~Cho, C.~Schmid, and J.~Ponce,
  ``Scnet: Learning semantic correspondence,'' in \emph{Proceedings of the IEEE
  International Conference on Computer Vision}, 2017, pp. 1831--1840.

\bibitem[He et~al.(2018)He, Chen, Liao, Sander, and Yuan]{he2018deep}
M.~He, D.~Chen, J.~Liao, P.~V. Sander, and L.~Yuan, ``Deep exemplar-based
  colorization,'' \emph{ACM Transactions on Graphics (TOG)}, vol.~37, no.~4,
  p.~47, 2018.

\bibitem[Hertzmann et~al.(2001)Hertzmann, Jacobs, Oliver, Curless, and
  Salesin]{hertzmann2001image}
A.~Hertzmann, C.~E. Jacobs, N.~Oliver, B.~Curless, and D.~H. Salesin, ``Image
  analogies,'' in \emph{Proceedings of the 28th annual conference on Computer
  graphics and interactive techniques}.\hskip 1em plus 0.5em minus 0.4em\relax
  ACM, 2001, pp. 327--340.

\bibitem[Heusel et~al.(2017)Heusel, Ramsauer, Unterthiner, Nessler, and
  Hochreiter]{heusel2017gans}
M.~Heusel, H.~Ramsauer, T.~Unterthiner, B.~Nessler, and S.~Hochreiter, ``Gans
  trained by a two time-scale update rule converge to a local nash
  equilibrium,'' in \emph{Advances in Neural Information Processing Systems},
  2017, pp. 6626--6637.

\bibitem[Huang and Belongie(2017)]{huang2017arbitrary}
X.~Huang and S.~Belongie, ``Arbitrary style transfer in real-time with adaptive
  instance normalization,'' in \emph{Proceedings of the IEEE International
  Conference on Computer Vision}, 2017, pp. 1501--1510.

\bibitem[Huang et~al.(2018)Huang, Liu, Belongie, and
  Kautz]{huang2018multimodal}
X.~Huang, M.-Y. Liu, S.~Belongie, and J.~Kautz, ``Multimodal unsupervised
  image-to-image translation,'' in \emph{Proceedings of the European Conference
  on Computer Vision (ECCV)}, 2018, pp. 172--189.

\bibitem[Isola et~al.(2017)Isola, Zhu, Zhou, and Efros]{isola2017image}
P.~Isola, J.-Y. Zhu, T.~Zhou, and A.~A. Efros, ``Image-to-image translation
  with conditional adversarial networks,'' in \emph{Proceedings of the IEEE
  conference on computer vision and pattern recognition}, 2017, pp. 1125--1134.

\bibitem[Johnson et~al.(2016)Johnson, Alahi, and
  Fei-Fei]{johnson2016perceptual}
J.~Johnson, A.~Alahi, and L.~Fei-Fei, ``Perceptual losses for real-time style
  transfer and super-resolution,'' in \emph{European conference on computer
  vision}.\hskip 1em plus 0.5em minus 0.4em\relax Springer, 2016, pp. 694--711.

\bibitem[Karras et~al.(2017)Karras, Aila, Laine, and
  Lehtinen]{karras2017progressive}
T.~Karras, T.~Aila, S.~Laine, and J.~Lehtinen, ``Progressive growing of {GAN}s
  for improved quality, stability, and variation,'' \emph{arXiv preprint
  arXiv:1710.10196}, 2017.

\bibitem[Kim et~al.(2017{\natexlab{b}})Kim, Min, Ham, Jeon, Lin, and
  Sohn]{kim2017fcss}
S.~Kim, D.~Min, B.~Ham, S.~Jeon, S.~Lin, and K.~Sohn, ``Fcss: Fully
  convolutional self-similarity for dense semantic correspondence,'' in
  \emph{Proceedings of the IEEE Conference on Computer Vision and Pattern
  Recognition}, 2017, pp. 6560--6569.

\bibitem[Kim et~al.(2017{\natexlab{a}})Kim, Cha, Kim, Lee, and
  Kim]{kim2017learning}
T.~Kim, M.~Cha, H.~Kim, J.~K. Lee, and J.~Kim, ``Learning to discover
  cross-domain relations with generative adversarial networks,'' in
  \emph{Proceedings of the 34th International Conference on Machine
  Learning-Volume 70}.\hskip 1em plus 0.5em minus 0.4em\relax JMLR. org, 2017,
  pp. 1857--1865.

\bibitem[Kingma and Ba(2014)]{kingma2014adam}
D.~P. Kingma and J.~Ba, ``Adam: A method for stochastic optimization,''
  \emph{arXiv preprint arXiv:1412.6980}, 2014.

\bibitem[Lee et~al.(2018)Lee, Tseng, Huang, Singh, and Yang]{lee2018diverse}
H.-Y. Lee, H.-Y. Tseng, J.-B. Huang, M.~Singh, and M.-H. Yang, ``Diverse
  image-to-image translation via disentangled representations,'' in
  \emph{Proceedings of the European Conference on Computer Vision (ECCV)},
  2018, pp. 35--51.

\bibitem[Lee et~al.(2019)Lee, Kim, Ponce, and Ham]{lee2019sfnet}
J.~Lee, D.~Kim, J.~Ponce, and B.~Ham, ``Sfnet: Learning object-aware semantic
  correspondence,'' in \emph{Proceedings of the IEEE Conference on Computer
  Vision and Pattern Recognition}, 2019, pp. 2278--2287.

\bibitem[{Li} et~al.(2019){Li}, {Wu}, {Weinberger}, and
  {Belongie}]{2019arXiv190704312L}
B.~{Li}, F.~{Wu}, K.~Q. {Weinberger}, and S.~{Belongie}, ``{Positional
  Normalization},'' \emph{arXiv e-prints}, p. arXiv:1907.04312, Jul. 2019.

\bibitem[Liao et~al.(2017)Liao, Yao, Yuan, Hua, and Kang]{liao2017visual}
J.~Liao, Y.~Yao, L.~Yuan, G.~Hua, and S.~B. Kang, ``Visual attribute transfer
  through deep image analogy,'' \emph{arXiv preprint arXiv:1705.01088}, 2017.

\bibitem[Lin et~al.(2017)Lin, Doll{\'a}r, Girshick, He, Hariharan, and
  Belongie]{lin2017feature}
T.-Y. Lin, P.~Doll{\'a}r, R.~Girshick, K.~He, B.~Hariharan, and S.~Belongie,
  ``Feature pyramid networks for object detection,'' in \emph{Proceedings of
  the IEEE conference on computer vision and pattern recognition}, 2017, pp.
  2117--2125.

\bibitem[Liu et~al.(2017)Liu, Breuel, and Kautz]{liu2017unsupervised}
M.-Y. Liu, T.~Breuel, and J.~Kautz, ``Unsupervised image-to-image translation
  networks,'' in \emph{Advances in neural information processing systems},
  2017, pp. 700--708.

\bibitem[Liu et~al.(2015)Liu, Luo, Wang, and Tang]{liu2015faceattributes}
Z.~Liu, P.~Luo, X.~Wang, and X.~Tang, ``Deep learning face attributes in the
  wild,'' in \emph{Proceedings of International Conference on Computer Vision
  (ICCV)}, Dec. 2015.

\bibitem[Liu et~al.(2016)Liu, Luo, Qiu, Wang, and
  Tang]{liuLQWTcvpr16DeepFashion}
Z.~Liu, P.~Luo, S.~Qiu, X.~Wang, and X.~Tang, ``Deepfashion: Powering robust
  clothes recognition and retrieval with rich annotations,'' in
  \emph{Proceedings of IEEE Conference on Computer Vision and Pattern
  Recognition (CVPR)}, Jun. 2016.

\bibitem[Long et~al.(2014)Long, Zhang, and Darrell]{long2014convnets}
J.~L. Long, N.~Zhang, and T.~Darrell, ``Do convnets learn correspondence?'' in
  \emph{Advances in Neural Information Processing Systems}, 2014, pp.
  1601--1609.

\bibitem[Lowe(2004)]{lowe2004distinctive}
D.~G. Lowe, ``Distinctive image features from scale-invariant keypoints,''
  \emph{International journal of computer vision}, vol.~60, no.~2, pp. 91--110,
  2004.

\bibitem[Ma et~al.(2019)Ma, Jia, Georgoulis, Tuytelaars, and
  Van~Gool]{ma2018exemplar}
L.~Ma, X.~Jia, S.~Georgoulis, T.~Tuytelaars, and L.~Van~Gool, ``Exemplar guided
  unsupervised image-to-image translation with semantic consistency,''
  \emph{ICLR}, 2019.

\bibitem[Mechrez et~al.(2018)Mechrez, Talmi, and
  Zelnik-Manor]{mechrez2018contextual}
R.~Mechrez, I.~Talmi, and L.~Zelnik-Manor, ``The contextual loss for image
  transformation with non-aligned data,'' in \emph{Proceedings of the European
  Conference on Computer Vision (ECCV)}, 2018, pp. 768--783.

\bibitem[Mirza and Osindero(2014)]{mirza2014conditional}
M.~Mirza and S.~Osindero, ``Conditional generative adversarial nets,''
  \emph{arXiv preprint arXiv:1411.1784}, 2014.

\bibitem[Miyato et~al.(2018)Miyato, Kataoka, Koyama, and
  Yoshida]{miyato2018spectral}
T.~Miyato, T.~Kataoka, M.~Koyama, and Y.~Yoshida, ``Spectral normalization for
  generative adversarial networks,'' \emph{arXiv preprint arXiv:1802.05957},
  2018.

\bibitem[Park et~al.(2019)Park, Liu, Wang, and Zhu]{park2019semantic}
T.~Park, M.-Y. Liu, T.-C. Wang, and J.-Y. Zhu, ``Semantic image synthesis with
  spatially-adaptive normalization,'' in \emph{Proceedings of the IEEE
  Conference on Computer Vision and Pattern Recognition}, 2019, pp. 2337--2346.

\bibitem[Qi et~al.(2018)Qi, Chen, Jia, and Koltun]{qi2018semi}
X.~Qi, Q.~Chen, J.~Jia, and V.~Koltun, ``Semi-parametric image synthesis,'' in
  \emph{Proceedings of the IEEE Conference on Computer Vision and Pattern
  Recognition}, 2018, pp. 8808--8816.

\bibitem[Riviere et~al.(2019)Riviere, Teytaud, Rapin, LeCun, and
  Couprie]{riviere2019inspirational}
M.~Riviere, O.~Teytaud, J.~Rapin, Y.~LeCun, and C.~Couprie, ``Inspirational
  adversarial image generation,'' \emph{arXiv preprint arXiv:1906.11661}, 2019.

\bibitem[Ronneberger et~al.(2015)Ronneberger, Fischer, and
  Brox]{ronneberger2015u}
O.~Ronneberger, P.~Fischer, and T.~Brox, ``U-net: Convolutional networks for
  biomedical image segmentation,'' in \emph{International Conference on Medical
  image computing and computer-assisted intervention}.\hskip 1em plus 0.5em
  minus 0.4em\relax Springer, 2015, pp. 234--241.

\bibitem[Royer et~al.(2017)Royer, Bousmalis, Gouws, Bertsch, Mosseri, Cole, and
  Murphy]{royer2017xgan}
A.~Royer, K.~Bousmalis, S.~Gouws, F.~Bertsch, I.~Mosseri, F.~Cole, and
  K.~Murphy, ``Xgan: Unsupervised image-to-image translation for many-to-many
  mappings,'' \emph{arXiv preprint arXiv:1711.05139}, 2017.

\bibitem[Tola et~al.(2009)Tola, Lepetit, and Fua]{tola2009daisy}
E.~Tola, V.~Lepetit, and P.~Fua, ``Daisy: An efficient dense descriptor applied
  to wide-baseline stereo,'' \emph{IEEE transactions on pattern analysis and
  machine intelligence}, vol.~32, no.~5, pp. 815--830, 2009.

\bibitem[Wang et~al.(2019)Wang, Yang, Li, Liang, Zhang, Hall, Hu,
  et~al.]{wang2019example}
M.~Wang, G.-Y. Yang, R.~Li, R.-Z. Liang, S.-H. Zhang, P.~Hall, S.-M. Hu
  \emph{et~al.}, ``Example-guided style consistent image synthesis from
  semantic labeling,'' \emph{arXiv preprint arXiv:1906.01314}, 2019.

\bibitem[Wang et~al.(2018)Wang, Liu, Zhu, Tao, Kautz, and
  Catanzaro]{wang2018high}
T.-C. Wang, M.-Y. Liu, J.-Y. Zhu, A.~Tao, J.~Kautz, and B.~Catanzaro,
  ``High-resolution image synthesis and semantic manipulation with conditional
  {GAN}s,'' in \emph{Proceedings of the IEEE conference on computer vision and
  pattern recognition}, 2018, pp. 8798--8807.

\bibitem[Xiao et~al.(2018)Xiao, Liu, Zhou, Jiang, and Sun]{xiao2018unified}
T.~Xiao, Y.~Liu, B.~Zhou, Y.~Jiang, and J.~Sun, ``Unified perceptual parsing
  for scene understanding,'' in \emph{Proceedings of the European Conference on
  Computer Vision (ECCV)}, 2018, pp. 418--434.

\bibitem[Yi et~al.(2019)Yi, Liu, Lai, and Rosin]{yi2019apdrawinggan}
R.~Yi, Y.-J. Liu, Y.-K. Lai, and P.~L. Rosin, ``Apdrawinggan: Generating
  artistic portrait drawings from face photos with hierarchical gans,'' in
  \emph{Proceedings of the IEEE Conference on Computer Vision and Pattern
  Recognition}, 2019, pp. 10\,743--10\,752.

\bibitem[Yi et~al.(2017)Yi, Zhang, Tan, and Gong]{yi2017dualgan}
Z.~Yi, H.~Zhang, P.~Tan, and M.~Gong, ``Dualgan: Unsupervised dual learning for
  image-to-image translation,'' in \emph{Proceedings of the IEEE international
  conference on computer vision}, 2017, pp. 2849--2857.

\bibitem[Zhang et~al.(2019)Zhang, He, Liao, Sander, Yuan, Bermak, and
  Chen]{zhang2019deep}
B.~Zhang, M.~He, J.~Liao, P.~V. Sander, L.~Yuan, A.~Bermak, and D.~Chen, ``Deep
  exemplar-based video colorization,'' in \emph{Proceedings of the IEEE
  Conference on Computer Vision and Pattern Recognition}, 2019, pp. 8052--8061.

\bibitem[Zhang et~al.(2018)Zhang, Goodfellow, Metaxas, and
  Odena]{zhang2018self}
H.~Zhang, I.~Goodfellow, D.~Metaxas, and A.~Odena, ``Self-attention generative
  adversarial networks,'' \emph{arXiv preprint arXiv:1805.08318}, 2018.

\bibitem[Zhou et~al.(2017)Zhou, Zhao, Puig, Fidler, Barriuso, and
  Torralba]{zhou2017scene}
B.~Zhou, H.~Zhao, X.~Puig, S.~Fidler, A.~Barriuso, and A.~Torralba, ``Scene
  parsing through ade20k dataset,'' in \emph{Proceedings of the IEEE conference
  on computer vision and pattern recognition}, 2017, pp. 633--641.

\bibitem[Zhou et~al.(2016)Zhou, Krahenbuhl, Aubry, Huang, and
  Efros]{zhou2016learning}
T.~Zhou, P.~Krahenbuhl, M.~Aubry, Q.~Huang, and A.~A. Efros, ``Learning dense
  correspondence via 3d-guided cycle consistency,'' in \emph{Proceedings of the
  IEEE Conference on Computer Vision and Pattern Recognition}, 2016, pp.
  117--126.

\bibitem[Zhu et~al.(2017{\natexlab{a}})Zhu, Park, Isola, and
  Efros]{zhu2017unpaired}
J.-Y. Zhu, T.~Park, P.~Isola, and A.~A. Efros, ``Unpaired image-to-image
  translation using cycle-consistent adversarial networks,'' in
  \emph{Proceedings of the IEEE international conference on computer vision},
  2017, pp. 2223--2232.

\bibitem[Zhu et~al.(2017{\natexlab{b}})Zhu, Zhang, Pathak, Darrell, Efros,
  Wang, and Shechtman]{zhu2017toward}
J.-Y. Zhu, R.~Zhang, D.~Pathak, T.~Darrell, A.~A. Efros, O.~Wang, and
  E.~Shechtman, ``Toward multimodal image-to-image translation,'' in
  \emph{Advances in Neural Information Processing Systems}, 2017, pp. 465--476.

\end{thebibliography}
}

\cleardoublepage
\onecolumn
\label{sec:supplemental}

\renewcommand\thesubsection{Appendix \Alph{subsection}}

\vspace{-4em}
\subsection{Additional Qualitative Results}

\begin{figure*}[h!]
\center
\small
\setlength\tabcolsep{0pt}
{
\renewcommand{\arraystretch}{0.0}
\begin{tabular}{@{}rrccccccc@{}}
    &
    \raisebox{0.25\height}{\rotatebox{90}{Exemplars}}&
    \includegraphics[width=0.125\columnwidth]{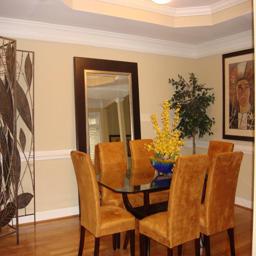}&
    \includegraphics[width=0.125\columnwidth]{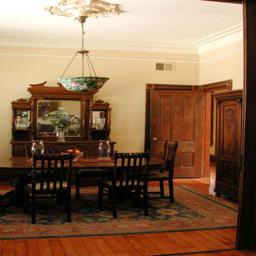}&
    \includegraphics[width=0.125\columnwidth]{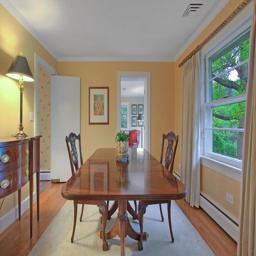}&
    \includegraphics[width=0.125\columnwidth]{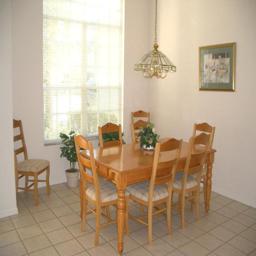}&
    \includegraphics[width=0.125\columnwidth]{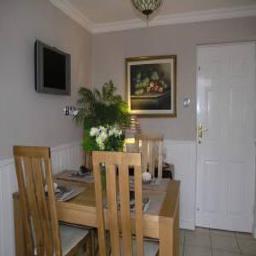}&
    \includegraphics[width=0.125\columnwidth]{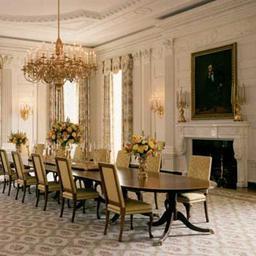}&
    \includegraphics[width=0.125\columnwidth]{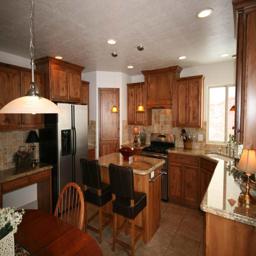}\\

    \raisebox{0.95\height}{\rotatebox{90}{Mask}}&
    \includegraphics[width=0.125\columnwidth]{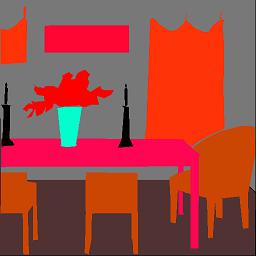}&
    \includegraphics[width=0.125\columnwidth]{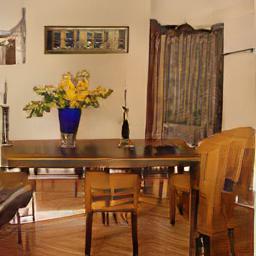}&
    \includegraphics[width=0.125\columnwidth]{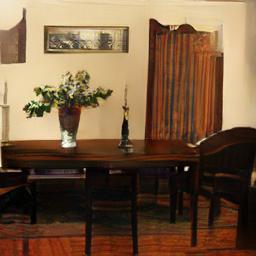}&
    \includegraphics[width=0.125\columnwidth]{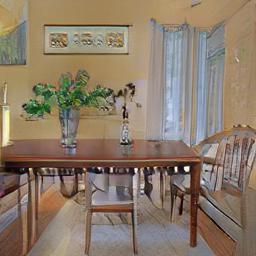}&
    \includegraphics[width=0.125\columnwidth]{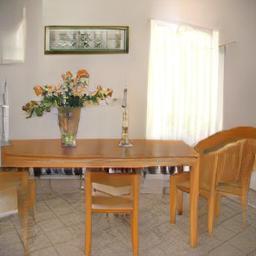}&
    \includegraphics[width=0.125\columnwidth]{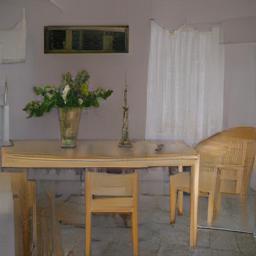}&
    \includegraphics[width=0.125\columnwidth]{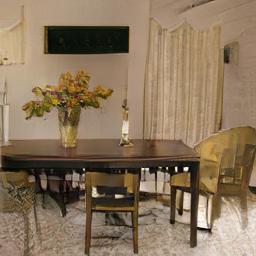}&
    \includegraphics[width=0.125\columnwidth]{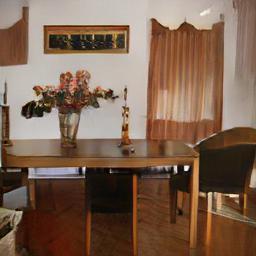}\\[0.5em]
    
    &
    \raisebox{0.25\height}{\rotatebox{90}{Exemplars}}&
    \includegraphics[width=0.125\columnwidth]{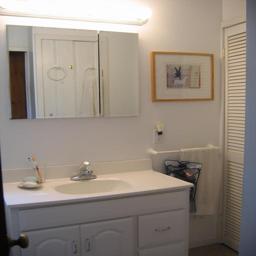}&
    \includegraphics[width=0.125\columnwidth]{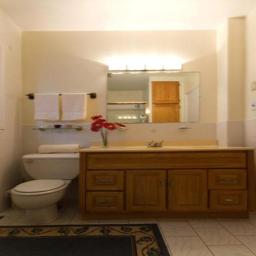}&
    \includegraphics[width=0.125\columnwidth]{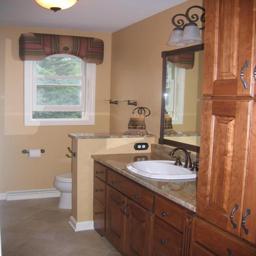}&
    \includegraphics[width=0.125\columnwidth]{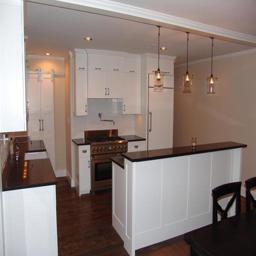}&
    \includegraphics[width=0.125\columnwidth]{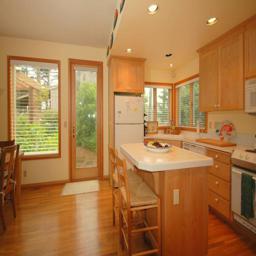}&
    \includegraphics[width=0.125\columnwidth]{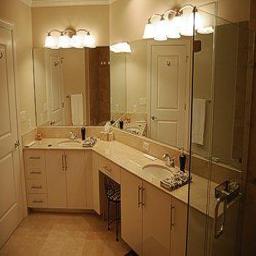}&
    \includegraphics[width=0.125\columnwidth]{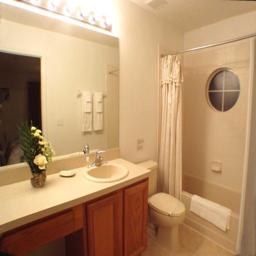}\\

    \raisebox{0.95\height}{\rotatebox{90}{Mask}}&
    \includegraphics[width=0.125\columnwidth]{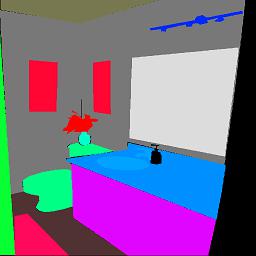}&
    \includegraphics[width=0.125\columnwidth]{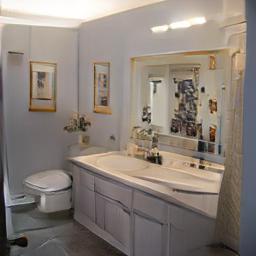}&
    \includegraphics[width=0.125\columnwidth]{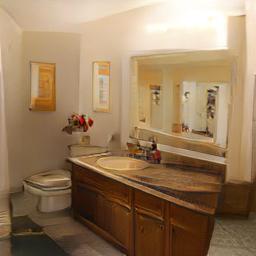}&
    \includegraphics[width=0.125\columnwidth]{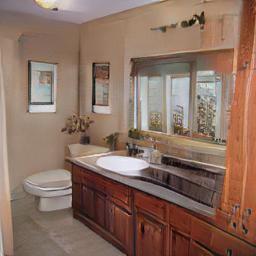}&
    \includegraphics[width=0.125\columnwidth]{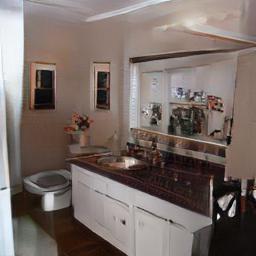}&
    \includegraphics[width=0.125\columnwidth]{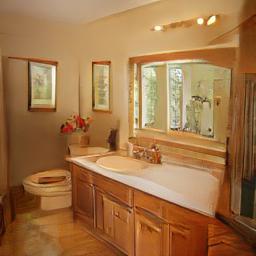}&
    \includegraphics[width=0.125\columnwidth]{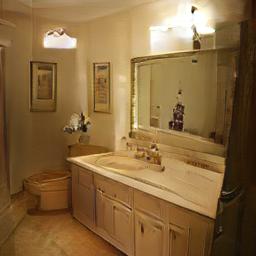}&
    \includegraphics[width=0.125\columnwidth]{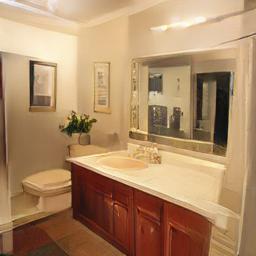}\\[0.5em]
    
        &
    \raisebox{0.25\height}{\rotatebox{90}{Exemplars}}&
    \includegraphics[width=0.125\columnwidth]{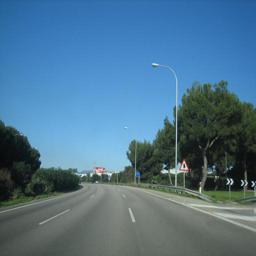}&
    \includegraphics[width=0.125\columnwidth]{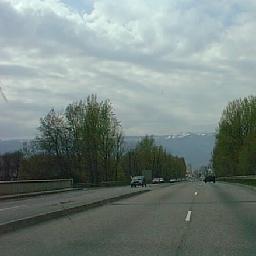}&
    \includegraphics[width=0.125\columnwidth]{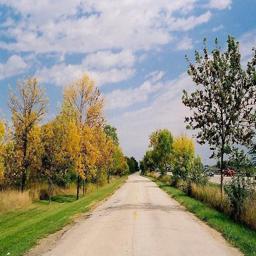}&
    \includegraphics[width=0.125\columnwidth]{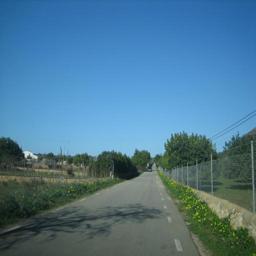}&
    \includegraphics[width=0.125\columnwidth]{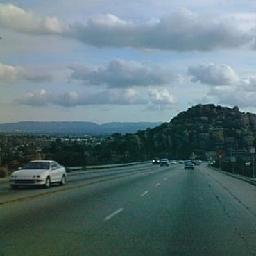}&
    \includegraphics[width=0.125\columnwidth]{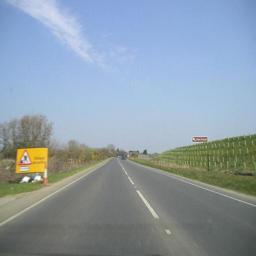}&
    \includegraphics[width=0.125\columnwidth]{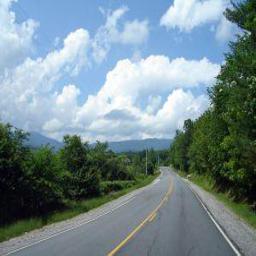}\\

    \raisebox{0.95\height}{\rotatebox{90}{Mask}}&
    \includegraphics[width=0.125\columnwidth]{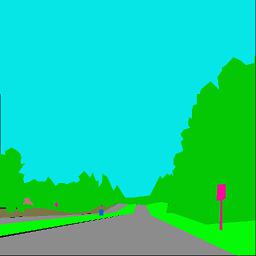}&
    \includegraphics[width=0.125\columnwidth]{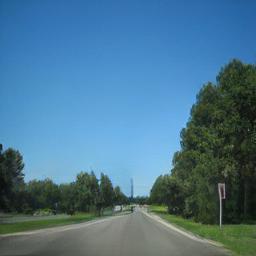}&
    \includegraphics[width=0.125\columnwidth]{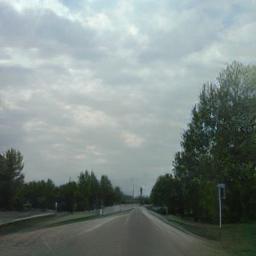}&
    \includegraphics[width=0.125\columnwidth]{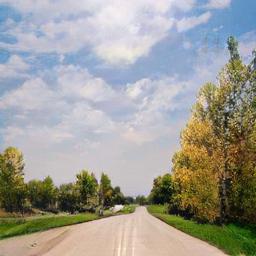}&
    \includegraphics[width=0.125\columnwidth]{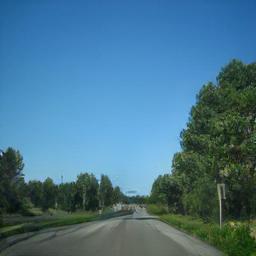}&
    \includegraphics[width=0.125\columnwidth]{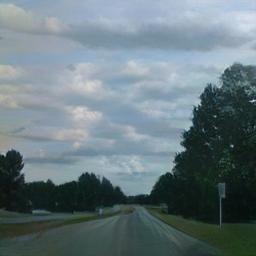}&
    \includegraphics[width=0.125\columnwidth]{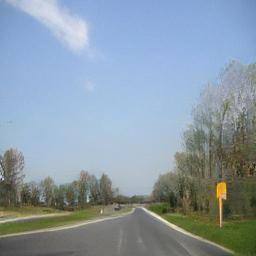}&
    \includegraphics[width=0.125\columnwidth]{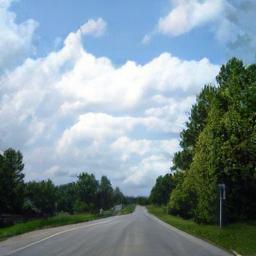}\\
\end{tabular}
}
\caption{\textbf{Our results of mask-to-image synthesis (ADE20k dataset).} In each group, the first row shows exemplars, and the second row shows the segmentation masks along with our results.}
\label{figure:ADE20k_1_result}
\end{figure*}

\begin{figure*}[h!]
\center
\small
\setlength\tabcolsep{0pt}
{
\renewcommand{\arraystretch}{0.0}
\begin{tabular}{@{}rrccccccc@{}}
    &
    \raisebox{0.25\height}{\rotatebox{90}{Exemplars}}&
    \includegraphics[width=0.125\columnwidth]{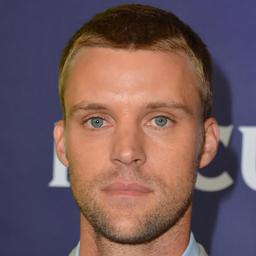}&
    \includegraphics[width=0.125\columnwidth]{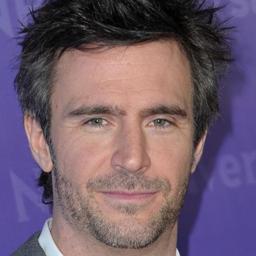}&
    \includegraphics[width=0.125\columnwidth]{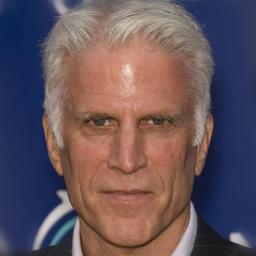}&
    \includegraphics[width=0.125\columnwidth]{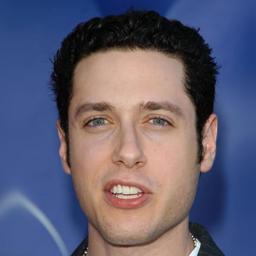}&
    \includegraphics[width=0.125\columnwidth]{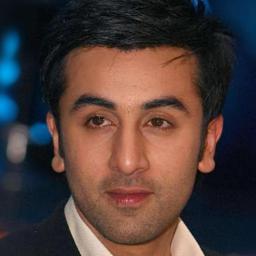}&
    \includegraphics[width=0.125\columnwidth]{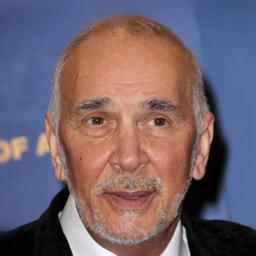}&
    \includegraphics[width=0.125\columnwidth]{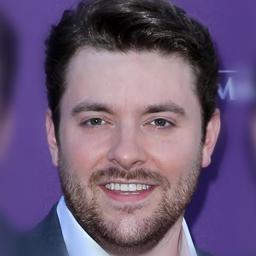}\\

    \raisebox{0.95\height}{\rotatebox{90}{Mask}}&
    \includegraphics[width=0.125\columnwidth]{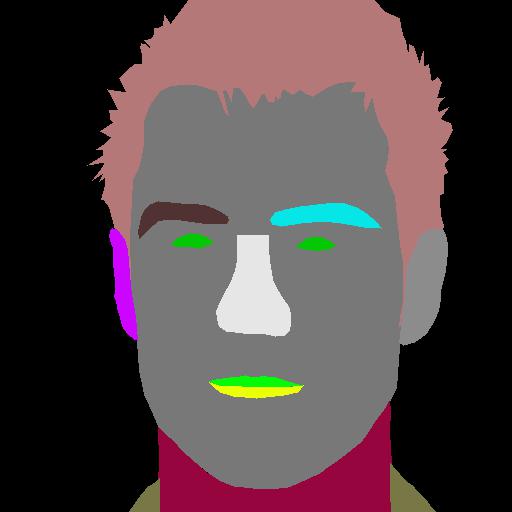}&
    \includegraphics[width=0.125\columnwidth]{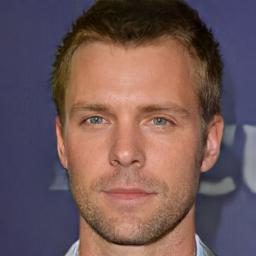}&
    \includegraphics[width=0.125\columnwidth]{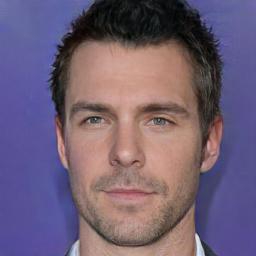}&
    \includegraphics[width=0.125\columnwidth]{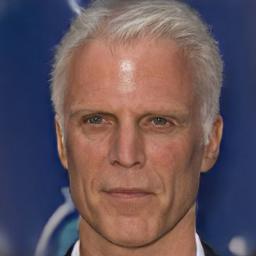}&
    \includegraphics[width=0.125\columnwidth]{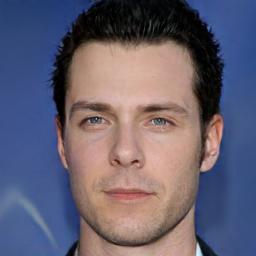}&
    \includegraphics[width=0.125\columnwidth]{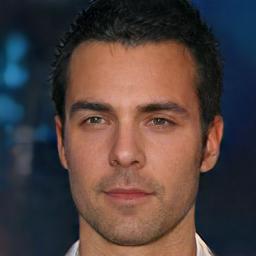}&
    \includegraphics[width=0.125\columnwidth]{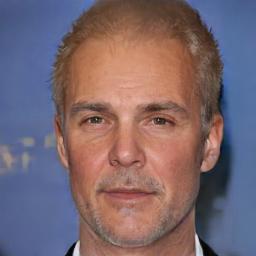}&
    \includegraphics[width=0.125\columnwidth]{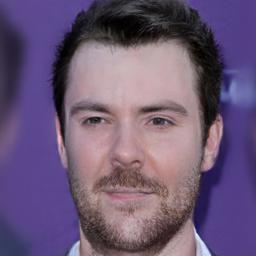}\\ [0.5em]
    
    &
    \raisebox{0.25\height}{\rotatebox{90}{Exemplars}}&
    \includegraphics[width=0.125\columnwidth]{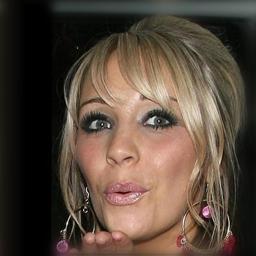}&
    \includegraphics[width=0.125\columnwidth]{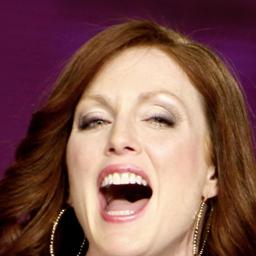}&
    \includegraphics[width=0.125\columnwidth]{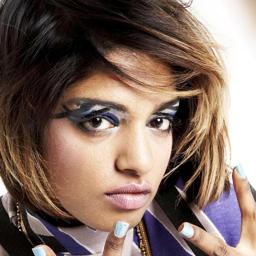}&
    \includegraphics[width=0.125\columnwidth]{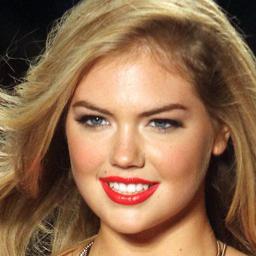}&
    \includegraphics[width=0.125\columnwidth]{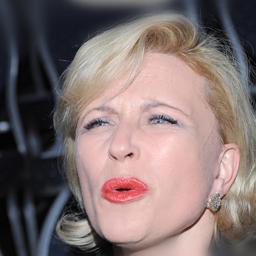}&
    \includegraphics[width=0.125\columnwidth]{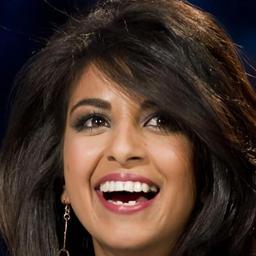}&
    \includegraphics[width=0.125\columnwidth]{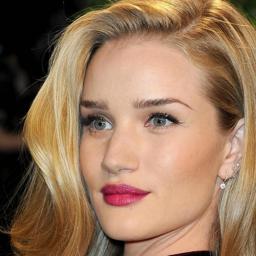}\\

    \raisebox{0.95\height}{\rotatebox{90}{Mask}}&
    \includegraphics[width=0.125\columnwidth]{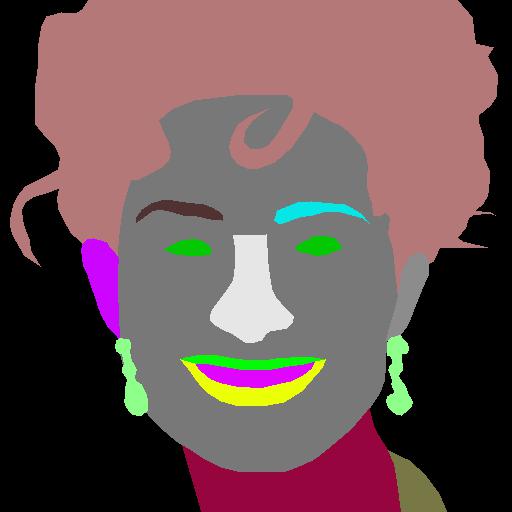}&
    \includegraphics[width=0.125\columnwidth]{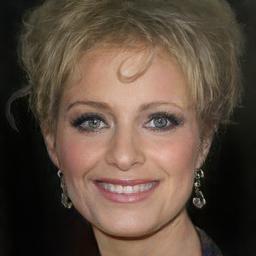}&
    \includegraphics[width=0.125\columnwidth]{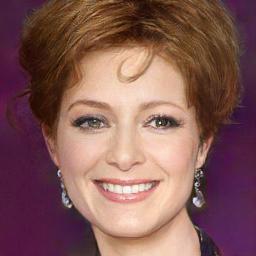}&
    \includegraphics[width=0.125\columnwidth]{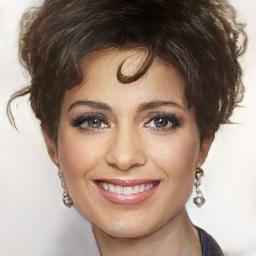}&
    \includegraphics[width=0.125\columnwidth]{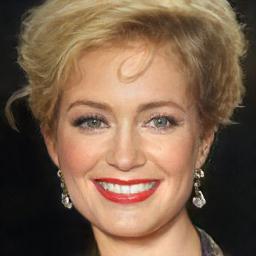}&
    \includegraphics[width=0.125\columnwidth]{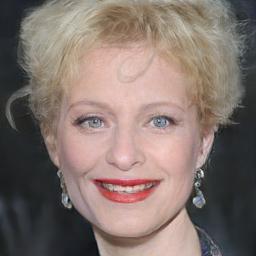}&
    \includegraphics[width=0.125\columnwidth]{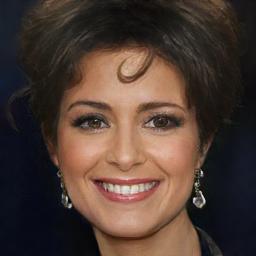}&
    \includegraphics[width=0.125\columnwidth]{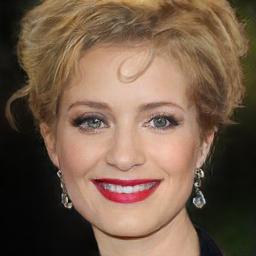}\\ [0.5em]

    &
    \raisebox{0.25\height}{\rotatebox{90}{Exemplars}}&
    \includegraphics[width=0.125\columnwidth]{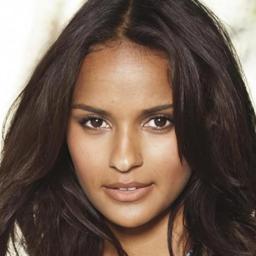}&
    \includegraphics[width=0.125\columnwidth]{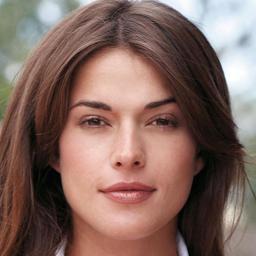}&
    \includegraphics[width=0.125\columnwidth]{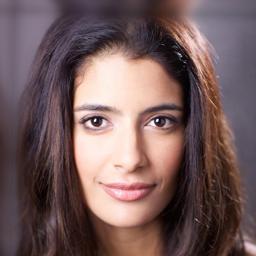}&
    \includegraphics[width=0.125\columnwidth]{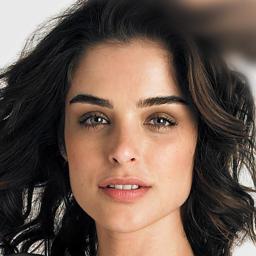}&
    \includegraphics[width=0.125\columnwidth]{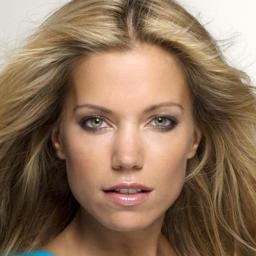}&
    \includegraphics[width=0.125\columnwidth]{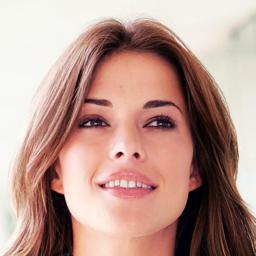}&
    \includegraphics[width=0.125\columnwidth]{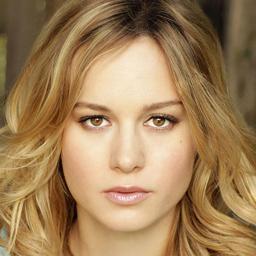}\\

    \raisebox{0.95\height}{\rotatebox{90}{Mask}}&
    \includegraphics[width=0.125\columnwidth]{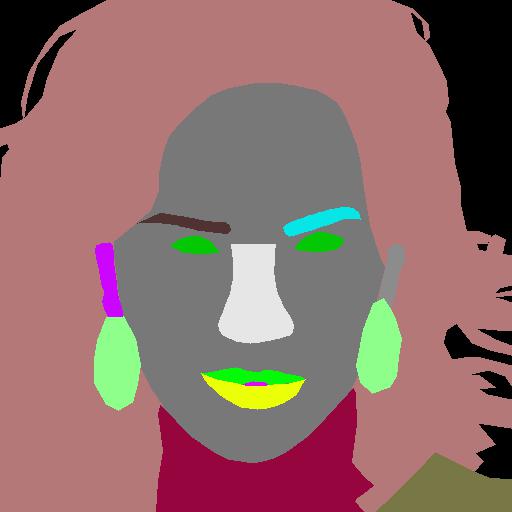}&
    \includegraphics[width=0.125\columnwidth]{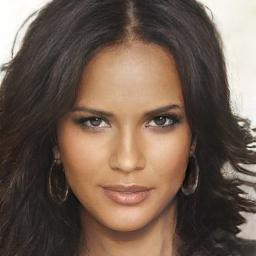}&
    \includegraphics[width=0.125\columnwidth]{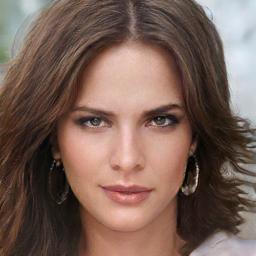}&
    \includegraphics[width=0.125\columnwidth]{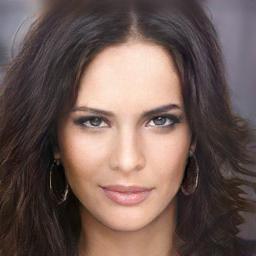}&
    \includegraphics[width=0.125\columnwidth]{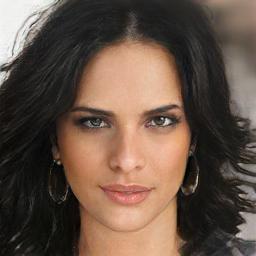}&
    \includegraphics[width=0.125\columnwidth]{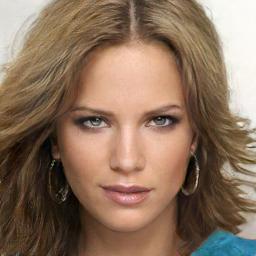}&
    \includegraphics[width=0.125\columnwidth]{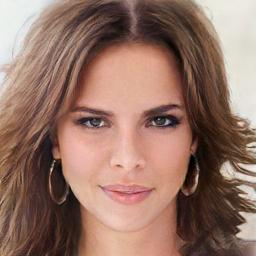}&
    \includegraphics[width=0.125\columnwidth]{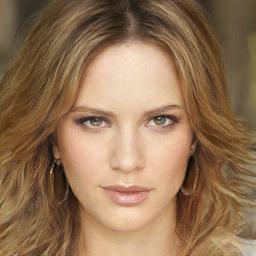}\\[0.5em]
    
    &
    \raisebox{0.25\height}{\rotatebox{90}{Exemplars}}&
    \includegraphics[width=0.125\columnwidth]{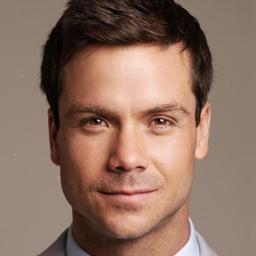}&
    \includegraphics[width=0.125\columnwidth]{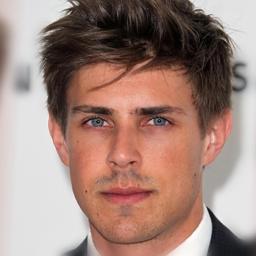}&
    \includegraphics[width=0.125\columnwidth]{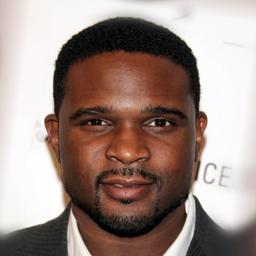}&
    \includegraphics[width=0.125\columnwidth]{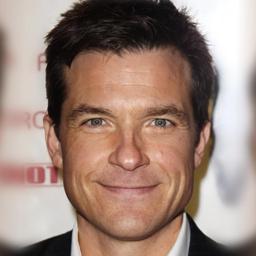}&
    \includegraphics[width=0.125\columnwidth]{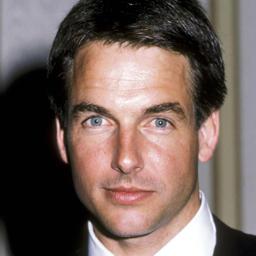}&
    \includegraphics[width=0.125\columnwidth]{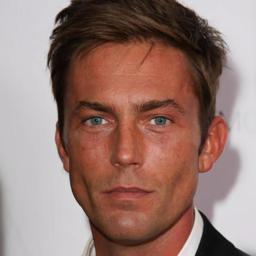}&
    \includegraphics[width=0.125\columnwidth]{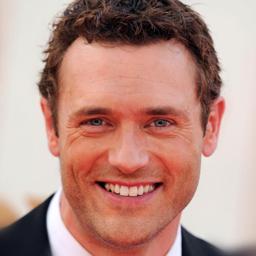}\\
    
    \raisebox{0.95\height}{\rotatebox{90}{Mask}}&
    \includegraphics[width=0.125\columnwidth]{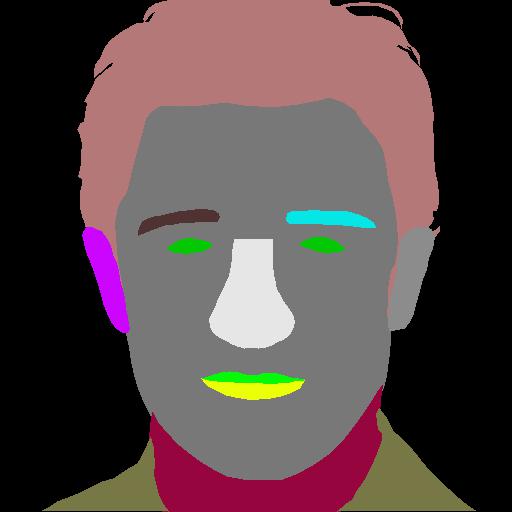}&
    \includegraphics[width=0.125\columnwidth]{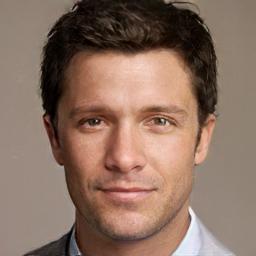}&
    \includegraphics[width=0.125\columnwidth]{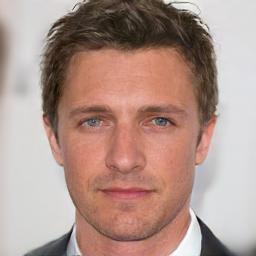}&
    \includegraphics[width=0.125\columnwidth]{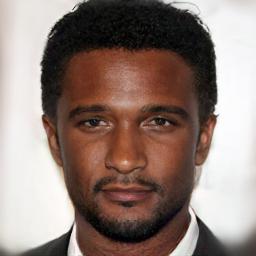}&
    \includegraphics[width=0.125\columnwidth]{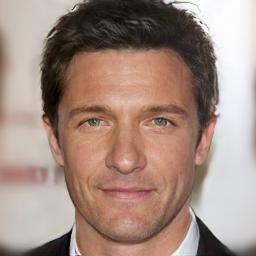}&
    \includegraphics[width=0.125\columnwidth]{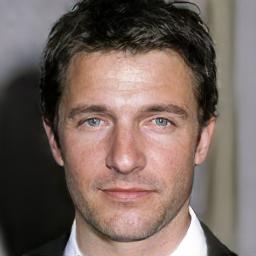}&
    \includegraphics[width=0.125\columnwidth]{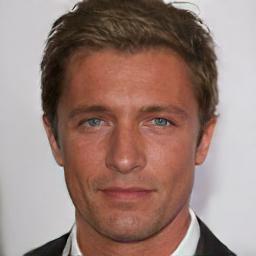}&
    \includegraphics[width=0.125\columnwidth]{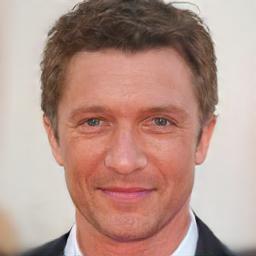}\\
\end{tabular}
}
\caption{\textbf{Our results of mask-to-image synthesis (CelebAHQ dataset).} In each group, the first row shows exemplars, and the second row shows the segmentation masks along with our results.}
\label{figure:CelebAHQM_0_result}
\end{figure*}

\begin{figure*}[h!]
\center
\small
\setlength\tabcolsep{0pt}
{
\renewcommand{\arraystretch}{0.0}
\begin{tabular}{@{}rrccccccc@{}}
    &
    \raisebox{0.25\height}{\rotatebox{90}{Exemplars}}&
    \includegraphics[width=0.125\columnwidth]{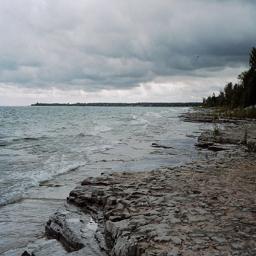}&
    \includegraphics[width=0.125\columnwidth]{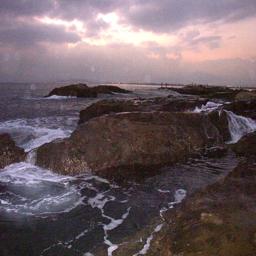}&
    \includegraphics[width=0.125\columnwidth]{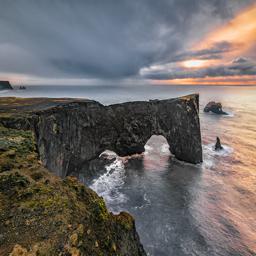}&
    \includegraphics[width=0.125\columnwidth]{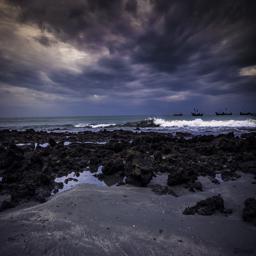}&
    \includegraphics[width=0.125\columnwidth]{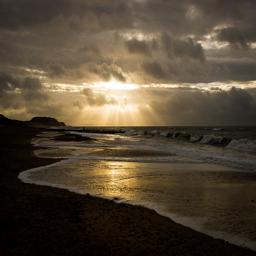}&
    \includegraphics[width=0.125\columnwidth]{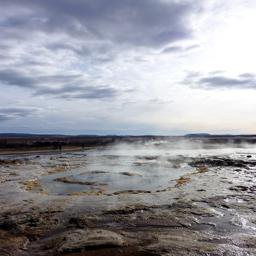}&
    \includegraphics[width=0.125\columnwidth]{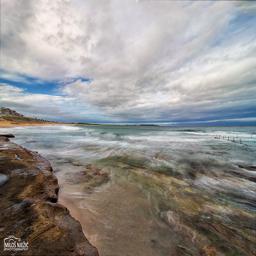}\\

    \raisebox{0.95\height}{\rotatebox{90}{Mask}}&
    \includegraphics[width=0.125\columnwidth]{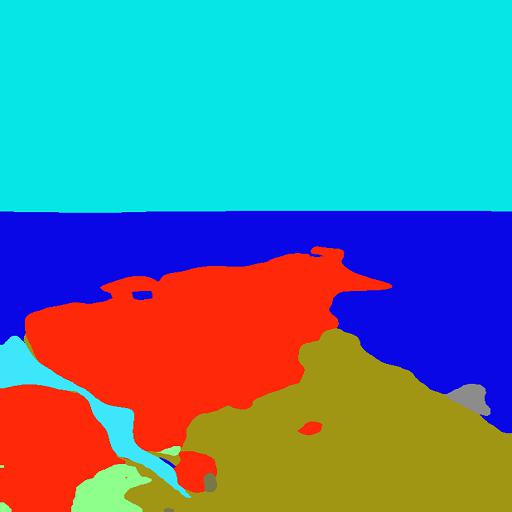}&
    \includegraphics[width=0.125\columnwidth]{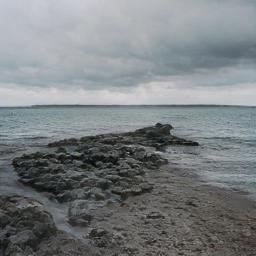}&
    \includegraphics[width=0.125\columnwidth]{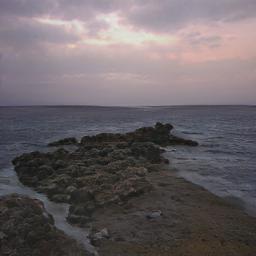}&
    \includegraphics[width=0.125\columnwidth]{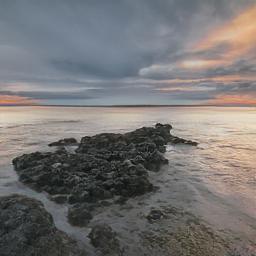}&
    \includegraphics[width=0.125\columnwidth]{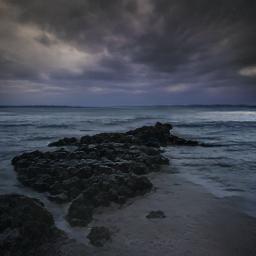}&
    \includegraphics[width=0.125\columnwidth]{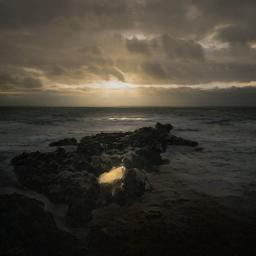}&
    \includegraphics[width=0.125\columnwidth]{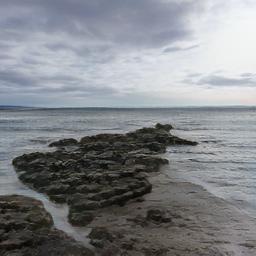}&
    \includegraphics[width=0.125\columnwidth]{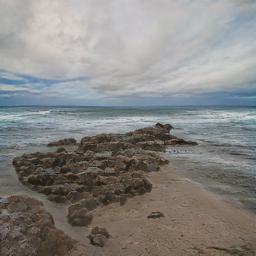}\\ [0.5em]


    &
    \raisebox{0.25\height}{\rotatebox{90}{Exemplars}}&
    \includegraphics[width=0.125\columnwidth]{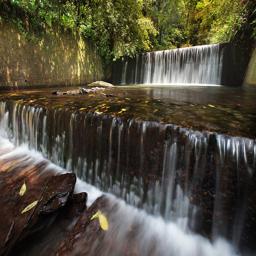}&
    \includegraphics[width=0.125\columnwidth]{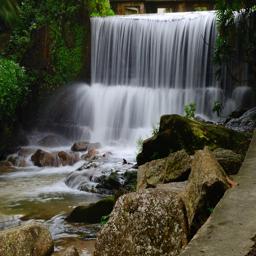}&
    \includegraphics[width=0.125\columnwidth]{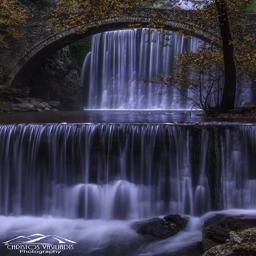}&
    \includegraphics[width=0.125\columnwidth]{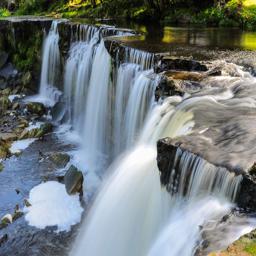}&
    \includegraphics[width=0.125\columnwidth]{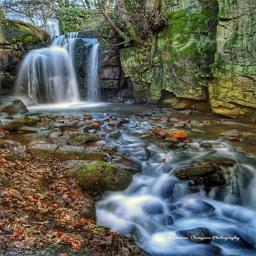}&
    \includegraphics[width=0.125\columnwidth]{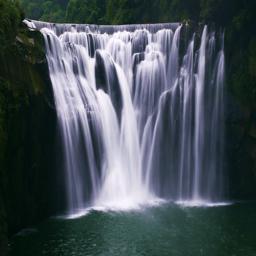}&
    \includegraphics[width=0.125\columnwidth]{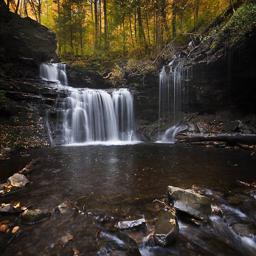}\\

    \raisebox{0.95\height}{\rotatebox{90}{Mask}}&
    \includegraphics[width=0.125\columnwidth]{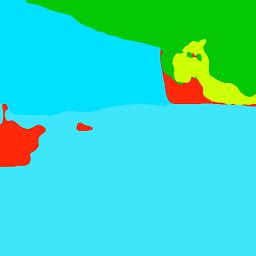}&
    \includegraphics[width=0.125\columnwidth]{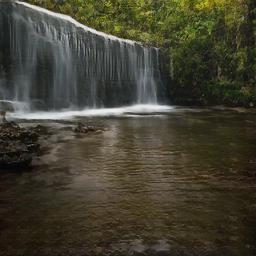}&
    \includegraphics[width=0.125\columnwidth]{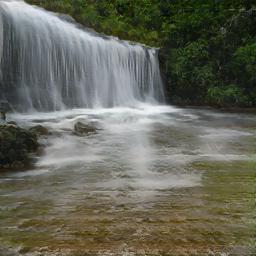}&
    \includegraphics[width=0.125\columnwidth]{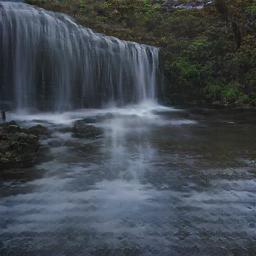}&
    \includegraphics[width=0.125\columnwidth]{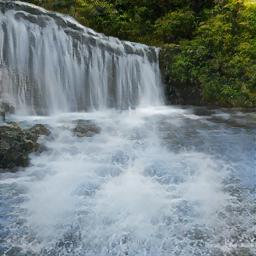}&
    \includegraphics[width=0.125\columnwidth]{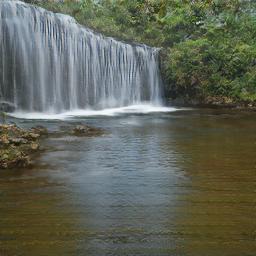}&
    \includegraphics[width=0.125\columnwidth]{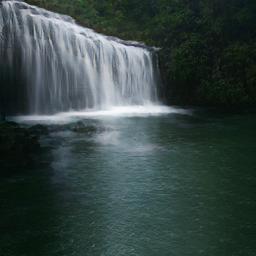}&
    \includegraphics[width=0.125\columnwidth]{Figures/Supplementary/Flickr/1915_1.jpg}\\ [0.5em]

    &
    \raisebox{0.25\height}{\rotatebox{90}{Exemplars}}&
    \includegraphics[width=0.125\columnwidth]{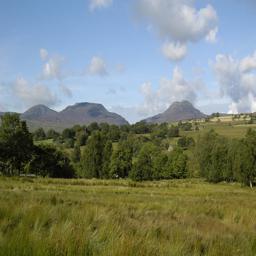}&
    \includegraphics[width=0.125\columnwidth]{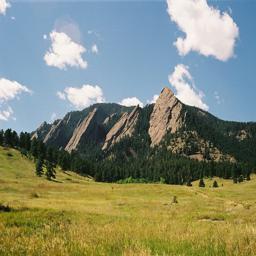}&
    \includegraphics[width=0.125\columnwidth]{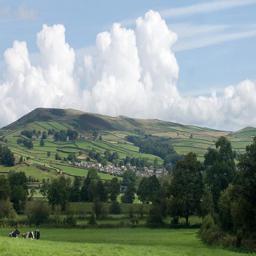}&
    \includegraphics[width=0.125\columnwidth]{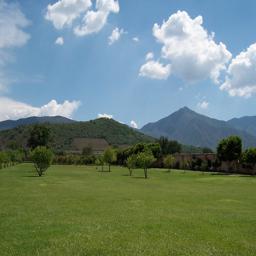}&
    \includegraphics[width=0.125\columnwidth]{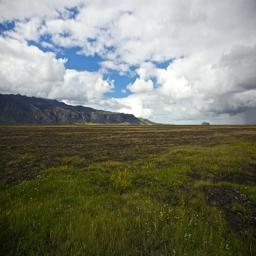}&
    \includegraphics[width=0.125\columnwidth]{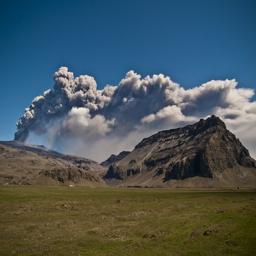}&
    \includegraphics[width=0.125\columnwidth]{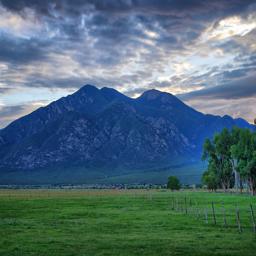}\\

    \raisebox{0.95\height}{\rotatebox{90}{Mask}}&
    \includegraphics[width=0.125\columnwidth]{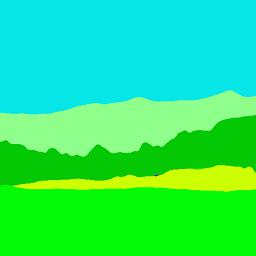}&
    \includegraphics[width=0.125\columnwidth]{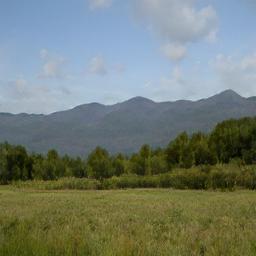}&
    \includegraphics[width=0.125\columnwidth]{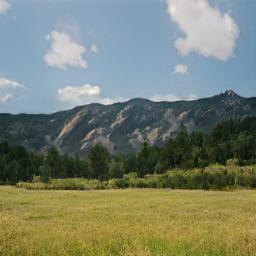}&
    \includegraphics[width=0.125\columnwidth]{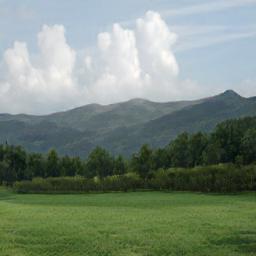}&
    \includegraphics[width=0.125\columnwidth]{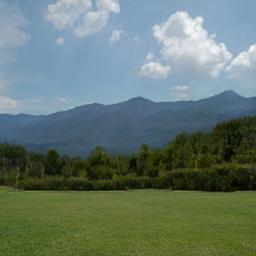}&
    \includegraphics[width=0.125\columnwidth]{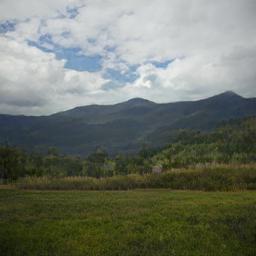}&
    \includegraphics[width=0.125\columnwidth]{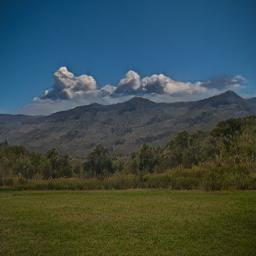}&
    \includegraphics[width=0.125\columnwidth]{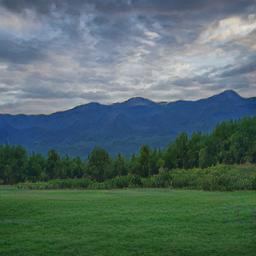}\\ [0.5em]

    &
    \raisebox{0.25\height}{\rotatebox{90}{Exemplars}}&
    \includegraphics[width=0.125\columnwidth]{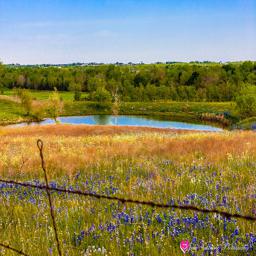}&
    \includegraphics[width=0.125\columnwidth]{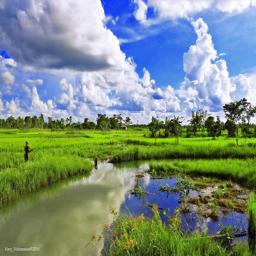}&
    \includegraphics[width=0.125\columnwidth]{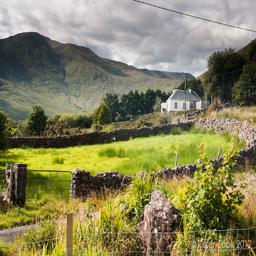}&
    \includegraphics[width=0.125\columnwidth]{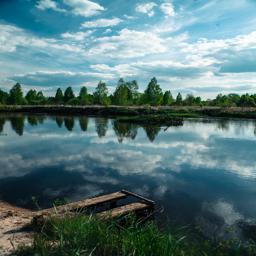}&
    \includegraphics[width=0.125\columnwidth]{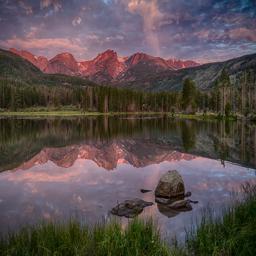}&
    \includegraphics[width=0.125\columnwidth]{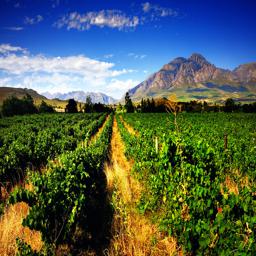}&
    \includegraphics[width=0.125\columnwidth]{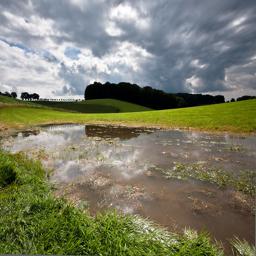}\\

    \raisebox{0.95\height}{\rotatebox{90}{Mask}}&
    \includegraphics[width=0.125\columnwidth]{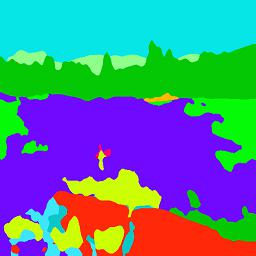}&
    \includegraphics[width=0.125\columnwidth]{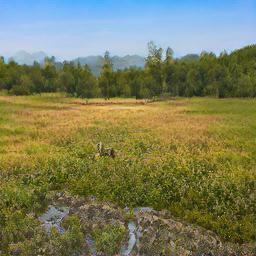}&
    \includegraphics[width=0.125\columnwidth]{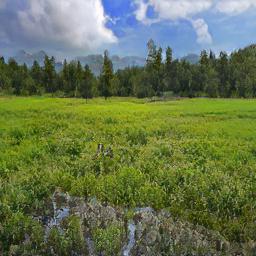}&
    \includegraphics[width=0.125\columnwidth]{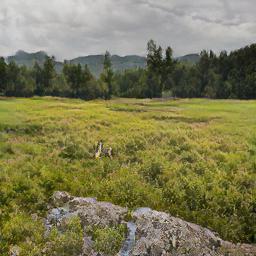}&
    \includegraphics[width=0.125\columnwidth]{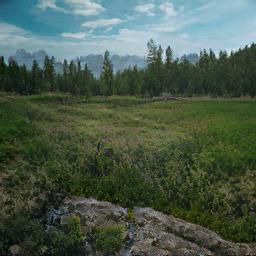}&
    \includegraphics[width=0.125\columnwidth]{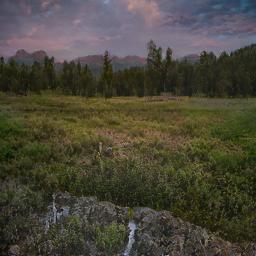}&
    \includegraphics[width=0.125\columnwidth]{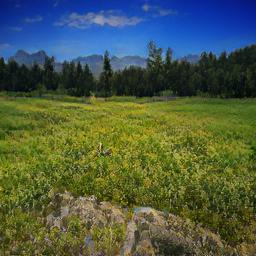}&
    \includegraphics[width=0.125\columnwidth]{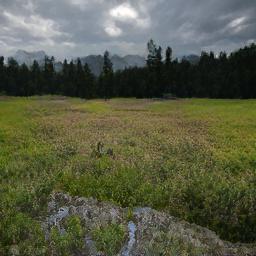}\\ 
\end{tabular}
}
\caption{\textbf{Our results of mask-to-image synthesis (Flickr dataset).} In each group, the first row shows exemplars, and the second row shows the segmentation masks along with our results.}
\label{figure:Flickr_0_result}
\end{figure*}

\noindent\textbf{Mask-to-image}~~
We perform mask-to-image synthesis on three datasets --- ADE20k, CelebA-HQ and Flickr dataset, and we show their results in Figure~\ref{figure:ADE20k_1_result}-\ref{figure:Flickr_0_result} respectively.

\begin{figure*}[h!]
\center
\small
\setlength\tabcolsep{0pt}
{
\renewcommand{\arraystretch}{0.0}
\begin{tabular}{@{}rrccccccc@{}}
    &
    \raisebox{0.25\height}{\rotatebox{90}{Exemplars}}&
    \includegraphics[width=0.125\columnwidth]{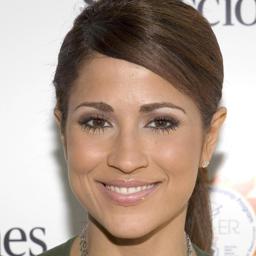}&
    \includegraphics[width=0.125\columnwidth]{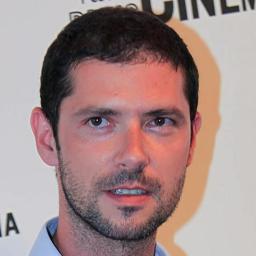}&
    \includegraphics[width=0.125\columnwidth]{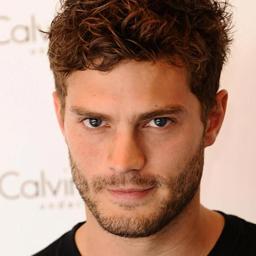}&
    \includegraphics[width=0.125\columnwidth]{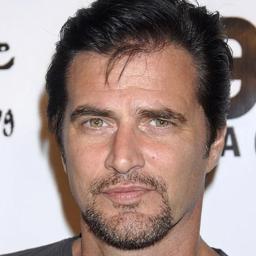}&
    \includegraphics[width=0.125\columnwidth]{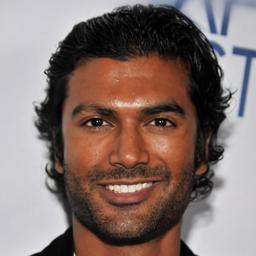}&
    \includegraphics[width=0.125\columnwidth]{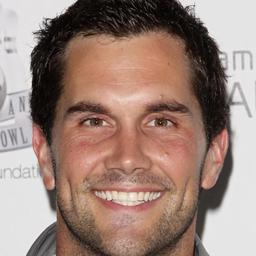}&
    \includegraphics[width=0.125\columnwidth]{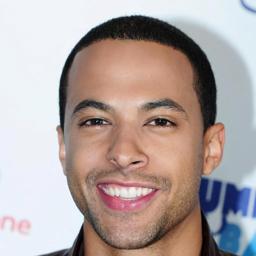}\\

    \raisebox{0.95\height}{\rotatebox{90}{Edge}}&
    \includegraphics[width=0.125\columnwidth]{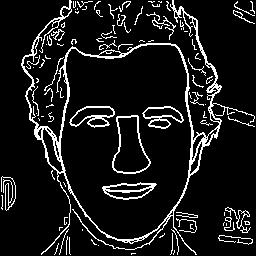}&
    \includegraphics[width=0.125\columnwidth]{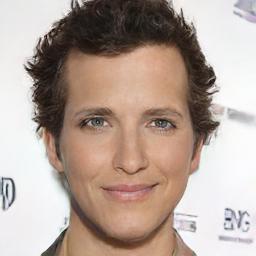}&
    \includegraphics[width=0.125\columnwidth]{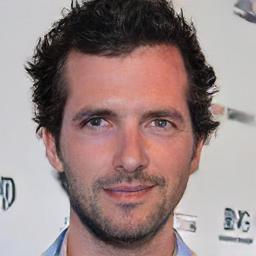}&
    \includegraphics[width=0.125\columnwidth]{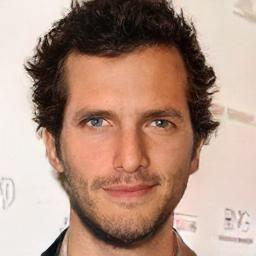}&
    \includegraphics[width=0.125\columnwidth]{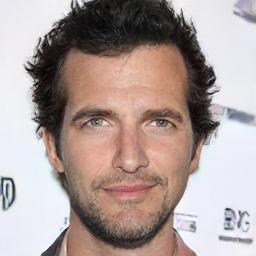}&
    \includegraphics[width=0.125\columnwidth]{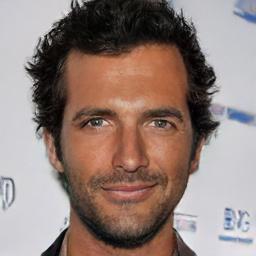}&
    \includegraphics[width=0.125\columnwidth]{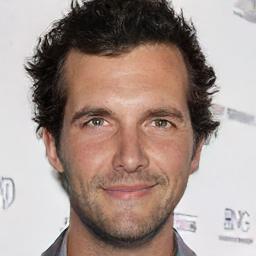}&
    \includegraphics[width=0.125\columnwidth]{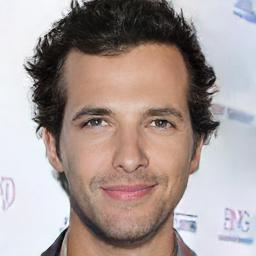}\\[0.5em]

    &
    \raisebox{0.25\height}{\rotatebox{90}{Exemplars}}&
    \includegraphics[width=0.125\columnwidth]{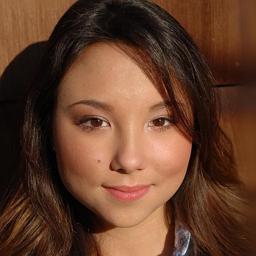}&
    \includegraphics[width=0.125\columnwidth]{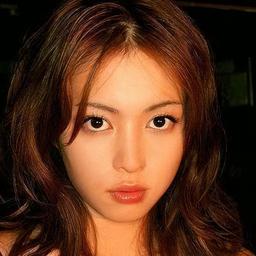}&
    \includegraphics[width=0.125\columnwidth]{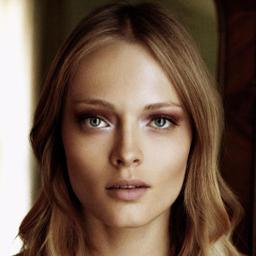}&
    \includegraphics[width=0.125\columnwidth]{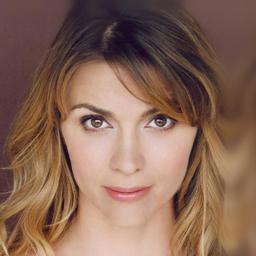}&
    \includegraphics[width=0.125\columnwidth]{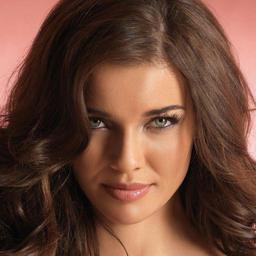}&
    \includegraphics[width=0.125\columnwidth]{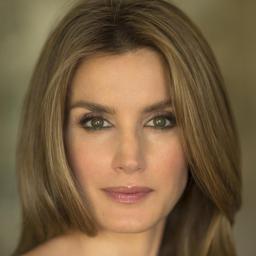}&
    \includegraphics[width=0.125\columnwidth]{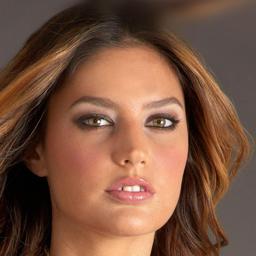}\\

    \raisebox{0.95\height}{\rotatebox{90}{Edge}}&
    \includegraphics[width=0.125\columnwidth]{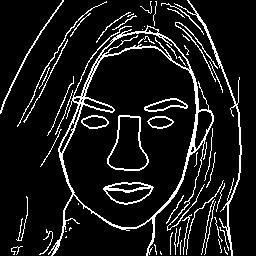}&
    \includegraphics[width=0.125\columnwidth]{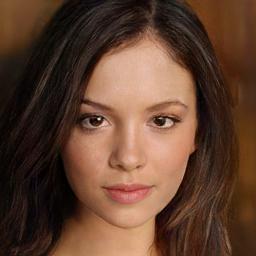}&
    \includegraphics[width=0.125\columnwidth]{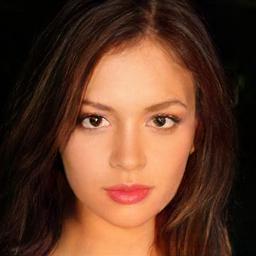}&
    \includegraphics[width=0.125\columnwidth]{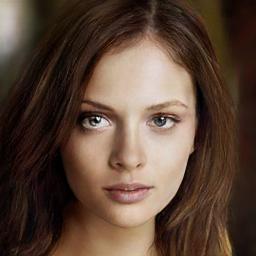}&
    \includegraphics[width=0.125\columnwidth]{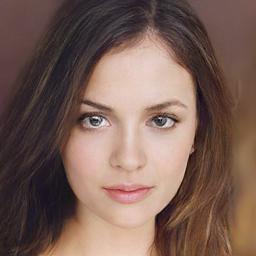}&
    \includegraphics[width=0.125\columnwidth]{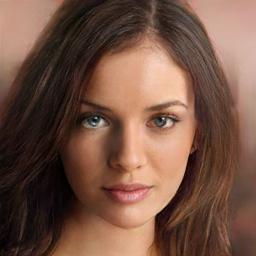}&
    \includegraphics[width=0.125\columnwidth]{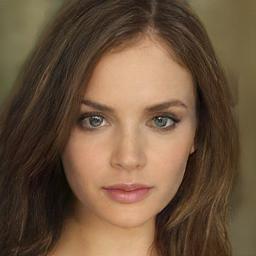}&
    \includegraphics[width=0.125\columnwidth]{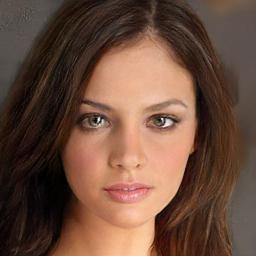}\\ [0.5em]
    
    &
    \raisebox{0.25\height}{\rotatebox{90}{Exemplars}}&
    \includegraphics[width=0.125\columnwidth]{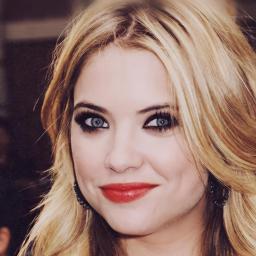}&
    \includegraphics[width=0.125\columnwidth]{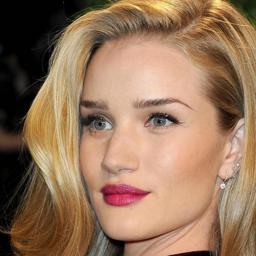}&
    \includegraphics[width=0.125\columnwidth]{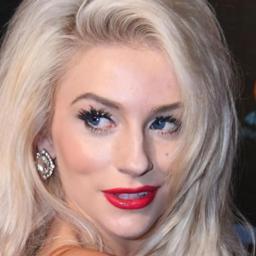}&
    \includegraphics[width=0.125\columnwidth]{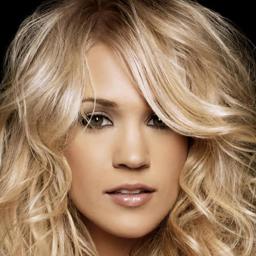}&
    \includegraphics[width=0.125\columnwidth]{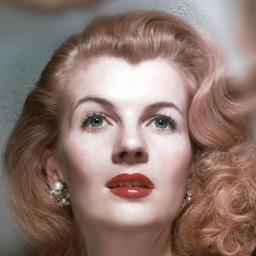}&
    \includegraphics[width=0.125\columnwidth]{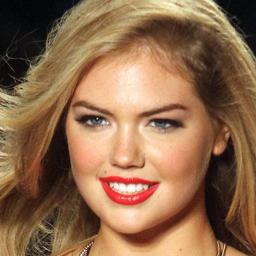}&
    \includegraphics[width=0.125\columnwidth]{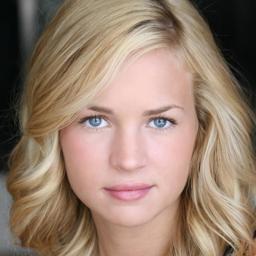}\\

    \raisebox{0.95\height}{\rotatebox{90}{Edge}}&
    \includegraphics[width=0.125\columnwidth]{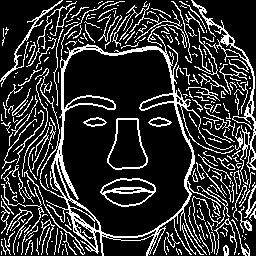}&
    \includegraphics[width=0.125\columnwidth]{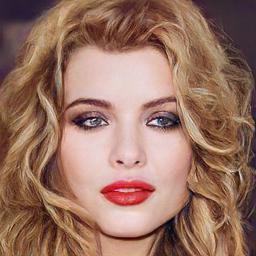}&
    \includegraphics[width=0.125\columnwidth]{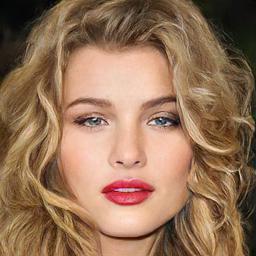}&
    \includegraphics[width=0.125\columnwidth]{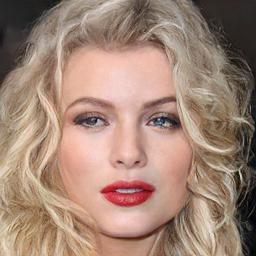}&
    \includegraphics[width=0.125\columnwidth]{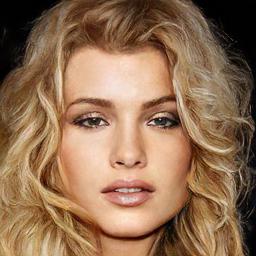}&
    \includegraphics[width=0.125\columnwidth]{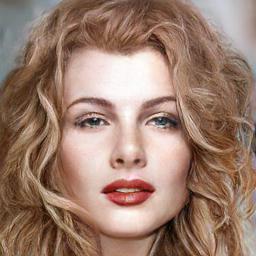}&
    \includegraphics[width=0.125\columnwidth]{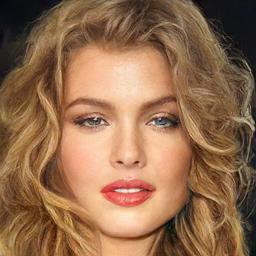}&
    \includegraphics[width=0.125\columnwidth]{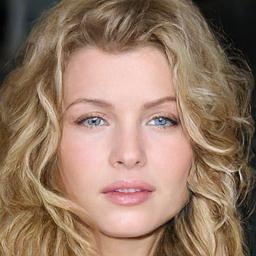}\\ [0.5em]
    
    &
    \raisebox{0.25\height}{\rotatebox{90}{Exemplars}}&
    \includegraphics[width=0.125\columnwidth]{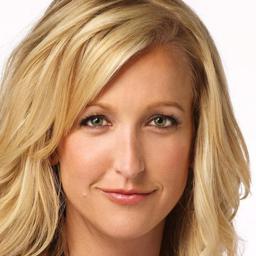}&
    \includegraphics[width=0.125\columnwidth]{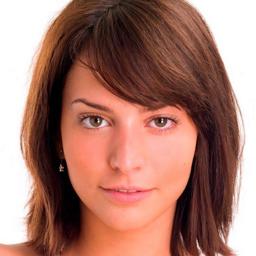}&
    \includegraphics[width=0.125\columnwidth]{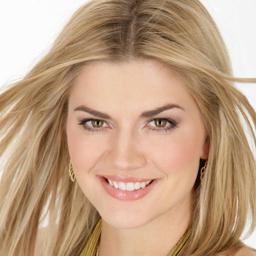}&
    \includegraphics[width=0.125\columnwidth]{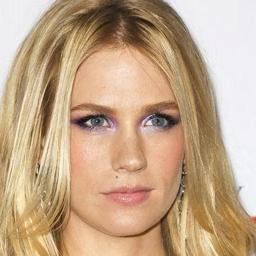}&
    \includegraphics[width=0.125\columnwidth]{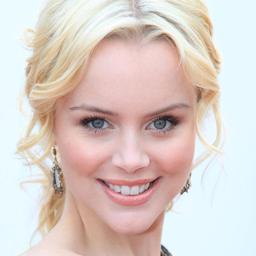}&
    \includegraphics[width=0.125\columnwidth]{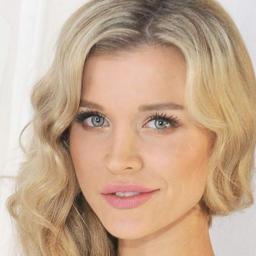}&
    \includegraphics[width=0.125\columnwidth]{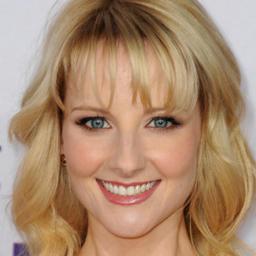}\\

    \raisebox{0.95\height}{\rotatebox{90}{Edge}}&
    \includegraphics[width=0.125\columnwidth]{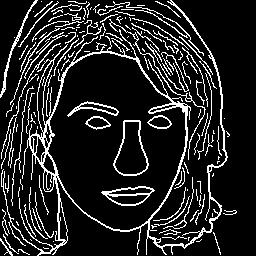}&
    \includegraphics[width=0.125\columnwidth]{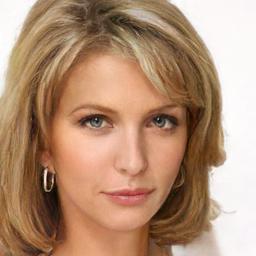}&
    \includegraphics[width=0.125\columnwidth]{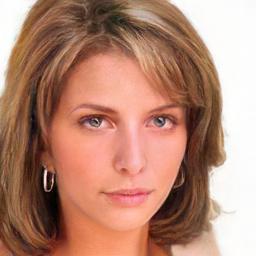}&
    \includegraphics[width=0.125\columnwidth]{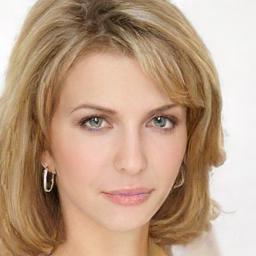}&
    \includegraphics[width=0.125\columnwidth]{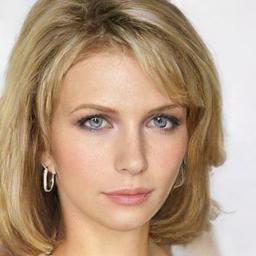}&
    \includegraphics[width=0.125\columnwidth]{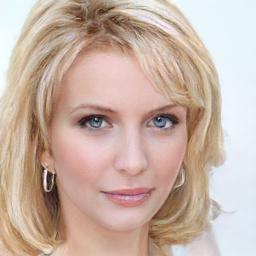}&
    \includegraphics[width=0.125\columnwidth]{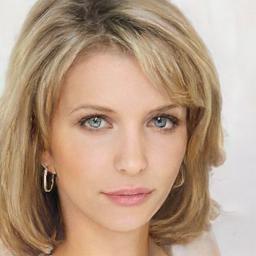}&
    \includegraphics[width=0.125\columnwidth]{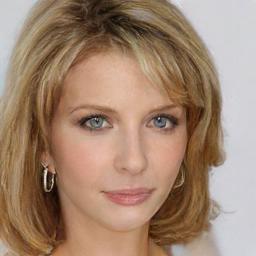}\\
\end{tabular}
}
\caption{\textbf{Our results of edge-to-face synthesis (CelebA-HQ dataset).} In each group, the first row shows exemplars, and the second row shows the edge maps along with our results.}
\label{figure:CelebAHQE_0_result}
\end{figure*}

\clearpage
\FloatBarrier
\noindent\textbf{Edge-to-face}~~  Figure~\ref{figure:CelebAHQE_0_result} shows additional results of edge-to-face synthesis on CelebA-HQ dataset.

\begin{figure*}[h!]
\center
\small
\setlength\tabcolsep{0pt}
{
\renewcommand{\arraystretch}{0.0}
\begin{tabular}{@{}rrccccccccccc@{}}
    &
    \raisebox{0.25\height}{\rotatebox{90}{Exemplars}}&
    \includegraphics[width=0.08\columnwidth]{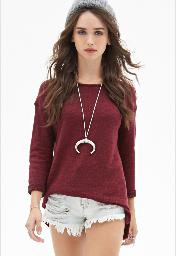}&
    \includegraphics[width=0.08\columnwidth]{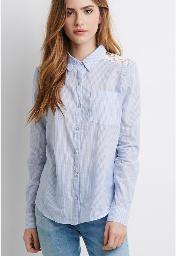}&
    \includegraphics[width=0.08\columnwidth]{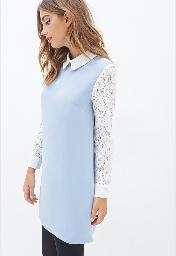}&
    \includegraphics[width=0.08\columnwidth]{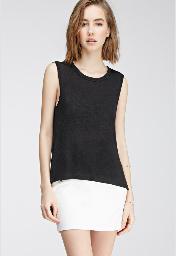}&
    \includegraphics[width=0.08\columnwidth]{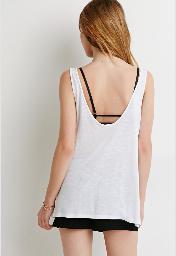}&
    \includegraphics[width=0.08\columnwidth]{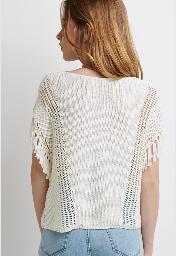}&
    \includegraphics[width=0.08\columnwidth]{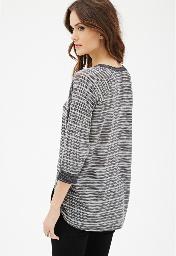}&
    \includegraphics[width=0.08\columnwidth]{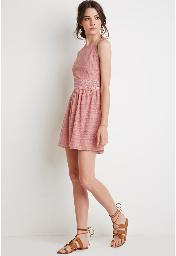}&
    \includegraphics[width=0.08\columnwidth]{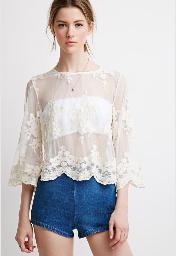}&
    \includegraphics[width=0.08\columnwidth]{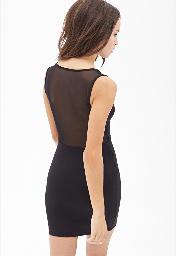}&\includegraphics[width=0.08\columnwidth]{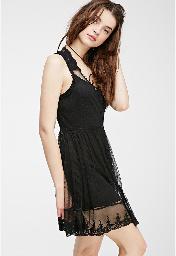}\\

    \raisebox{1.15\height}{\rotatebox{90}{Pose}}&
    \includegraphics[width=0.08\columnwidth]{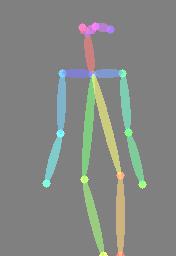}&
    \includegraphics[width=0.08\columnwidth]{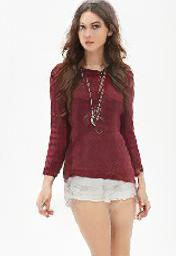}&
    \includegraphics[width=0.08\columnwidth]{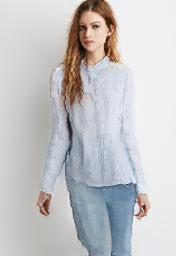}&
    \includegraphics[width=0.08\columnwidth]{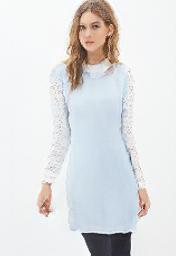}&
    \includegraphics[width=0.08\columnwidth]{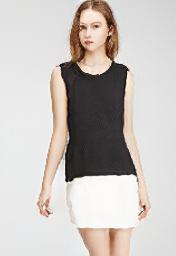}&
    \includegraphics[width=0.08\columnwidth]{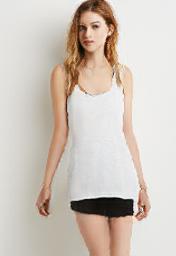}&
    \includegraphics[width=0.08\columnwidth]{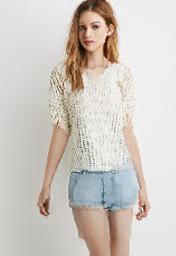}&
    \includegraphics[width=0.08\columnwidth]{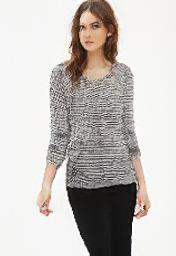}&
    \includegraphics[width=0.08\columnwidth]{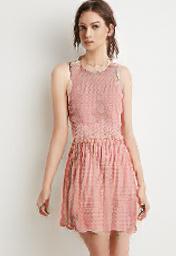}&
    \includegraphics[width=0.08\columnwidth]{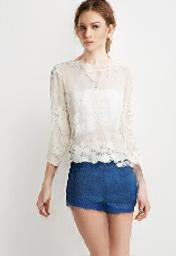}&
    \includegraphics[width=0.08\columnwidth]{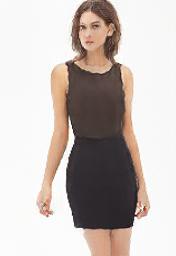}&
    \includegraphics[width=0.08\columnwidth]{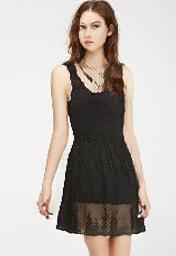}\\ [0.5em]

    &
    \raisebox{0.25\height}{\rotatebox{90}{Exemplars}}&
    \includegraphics[width=0.08\columnwidth]{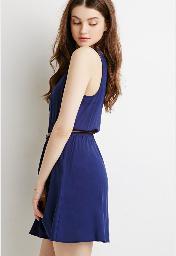}&
    \includegraphics[width=0.08\columnwidth]{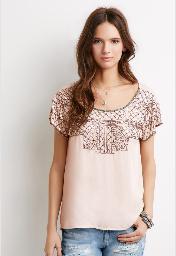}&
    \includegraphics[width=0.08\columnwidth]{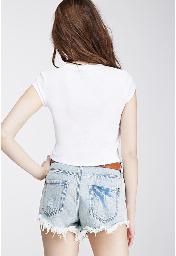}&
    \includegraphics[width=0.08\columnwidth]{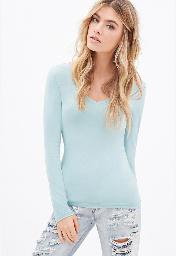}&
    \includegraphics[width=0.08\columnwidth]{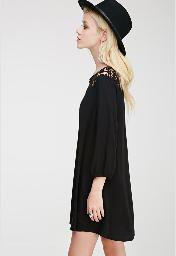}&
    \includegraphics[width=0.08\columnwidth]{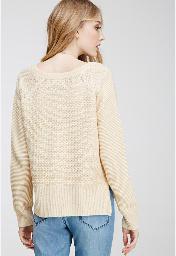}&
    \includegraphics[width=0.08\columnwidth]{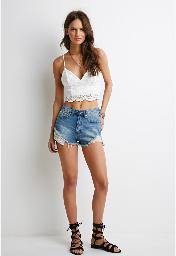}&
    \includegraphics[width=0.08\columnwidth]{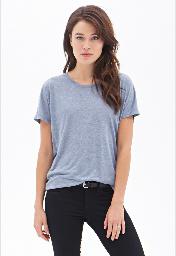}&
    \includegraphics[width=0.08\columnwidth]{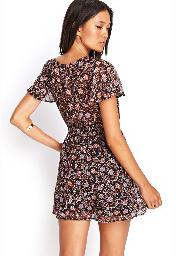}&
    \includegraphics[width=0.08\columnwidth]{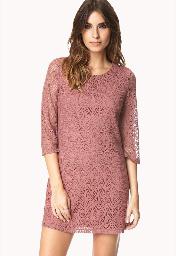}&\includegraphics[width=0.08\columnwidth]{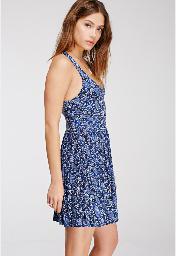}\\

    \raisebox{1.15\height}{\rotatebox{90}{Pose}}&
    \includegraphics[width=0.08\columnwidth]{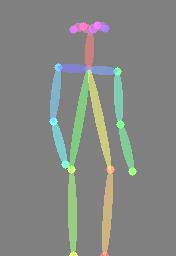}&
    \includegraphics[width=0.08\columnwidth]{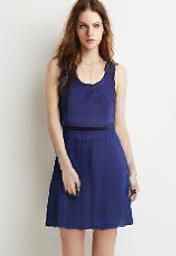}&
    \includegraphics[width=0.08\columnwidth]{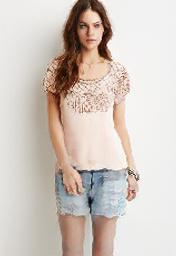}&
    \includegraphics[width=0.08\columnwidth]{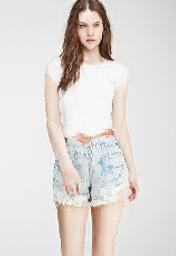}&
    \includegraphics[width=0.08\columnwidth]{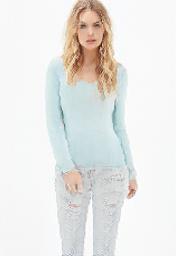}&
    \includegraphics[width=0.08\columnwidth]{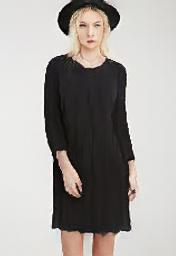}&
    \includegraphics[width=0.08\columnwidth]{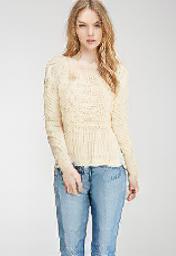}&
    \includegraphics[width=0.08\columnwidth]{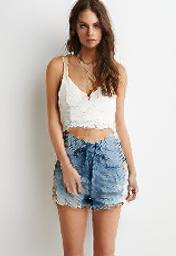}&
    \includegraphics[width=0.08\columnwidth]{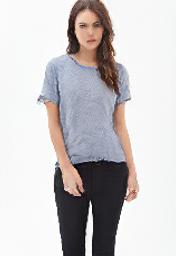}&
    \includegraphics[width=0.08\columnwidth]{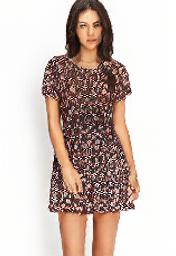}&
    \includegraphics[width=0.08\columnwidth]{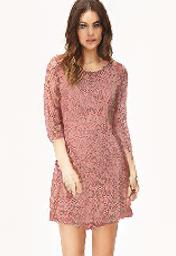}&
    \includegraphics[width=0.08\columnwidth]{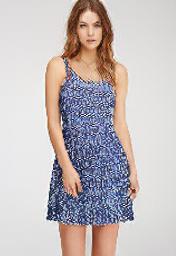}\\ [0.5em]
    
    &
    \raisebox{0.25\height}{\rotatebox{90}{Exemplars}}&
    \includegraphics[width=0.08\columnwidth]{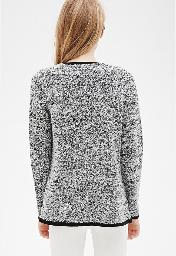}&
    \includegraphics[width=0.08\columnwidth]{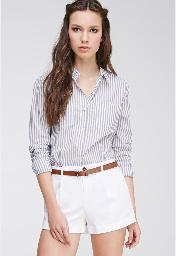}&
    \includegraphics[width=0.08\columnwidth]{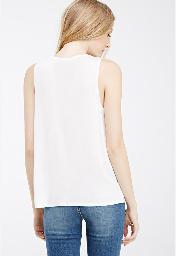}&
    \includegraphics[width=0.08\columnwidth]{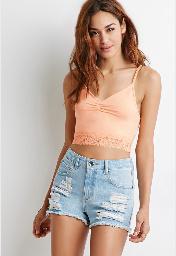}&
    \includegraphics[width=0.08\columnwidth]{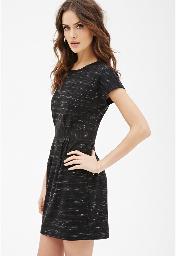}&
    \includegraphics[width=0.08\columnwidth]{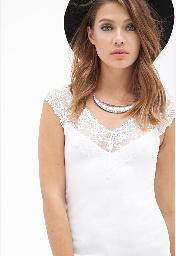}&
    \includegraphics[width=0.08\columnwidth]{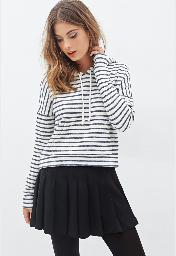}&
    \includegraphics[width=0.08\columnwidth]{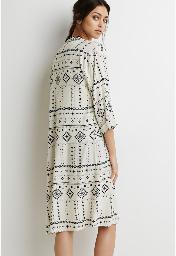}&
    \includegraphics[width=0.08\columnwidth]{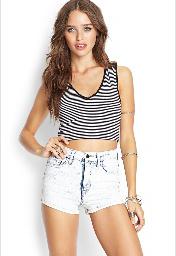}&
    \includegraphics[width=0.08\columnwidth]{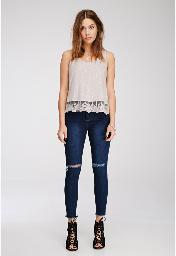}&\includegraphics[width=0.08\columnwidth]{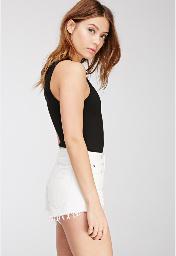}\\

    \raisebox{1.15\height}{\rotatebox{90}{Pose}}&
    \includegraphics[width=0.08\columnwidth]{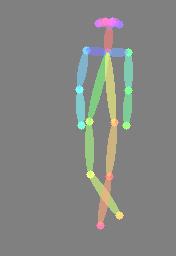}&
    \includegraphics[width=0.08\columnwidth]{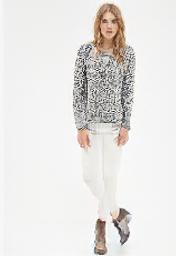}&
    \includegraphics[width=0.08\columnwidth]{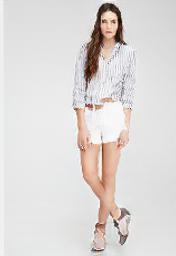}&
    \includegraphics[width=0.08\columnwidth]{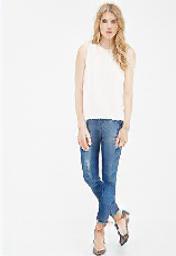}&
    \includegraphics[width=0.08\columnwidth]{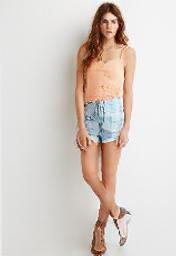}&
    \includegraphics[width=0.08\columnwidth]{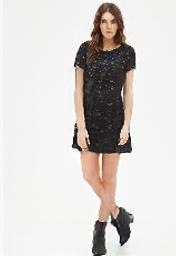}&
    \includegraphics[width=0.08\columnwidth]{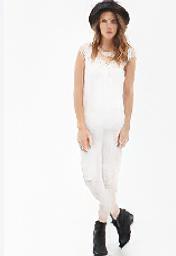}&
    \includegraphics[width=0.08\columnwidth]{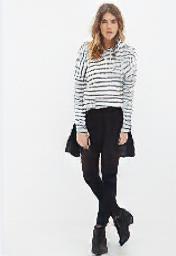}&
    \includegraphics[width=0.08\columnwidth]{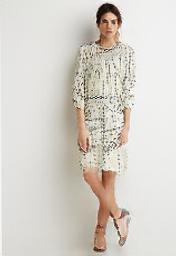}&
    \includegraphics[width=0.08\columnwidth]{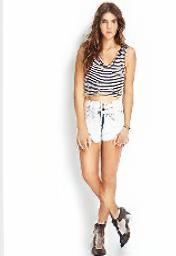}&
    \includegraphics[width=0.08\columnwidth]{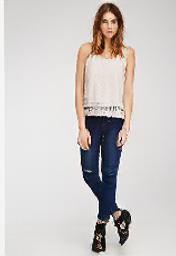}&
    \includegraphics[width=0.08\columnwidth]{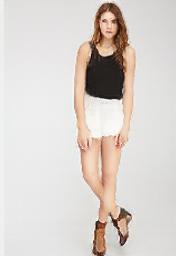}\\ [0.5em]
    
    &
    \raisebox{0.25\height}{\rotatebox{90}{Exemplars}}&
    \includegraphics[width=0.08\columnwidth]{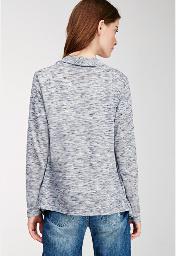}&
    \includegraphics[width=0.08\columnwidth]{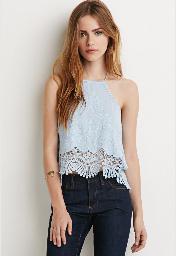}&
    \includegraphics[width=0.08\columnwidth]{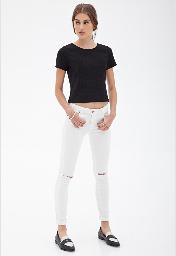}&
    \includegraphics[width=0.08\columnwidth]{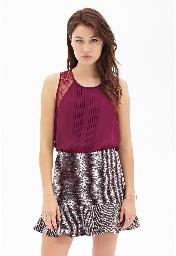}&
    \includegraphics[width=0.08\columnwidth]{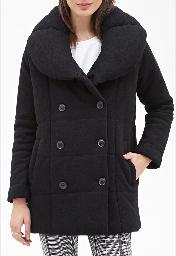}&
    \includegraphics[width=0.08\columnwidth]{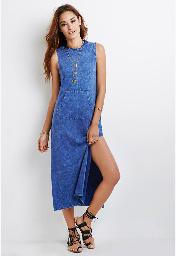}&
    \includegraphics[width=0.08\columnwidth]{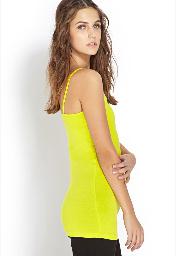}&
    \includegraphics[width=0.08\columnwidth]{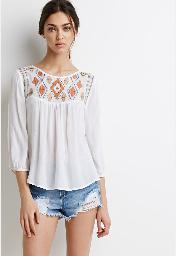}&
    \includegraphics[width=0.08\columnwidth]{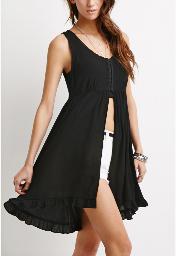}&
    \includegraphics[width=0.08\columnwidth]{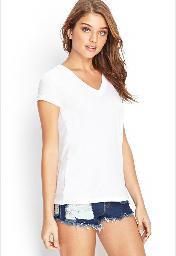}&
    \includegraphics[width=0.08\columnwidth]{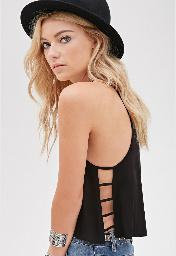}\\

    \raisebox{1.15\height}{\rotatebox{90}{Pose}}&
    \includegraphics[width=0.08\columnwidth]{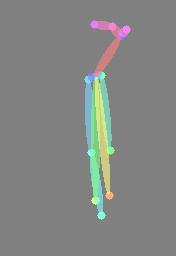}&
    \includegraphics[width=0.08\columnwidth]{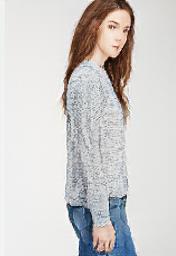}&
    \includegraphics[width=0.08\columnwidth]{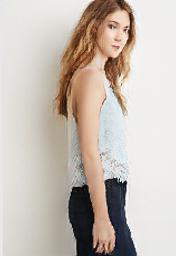}&
    \includegraphics[width=0.08\columnwidth]{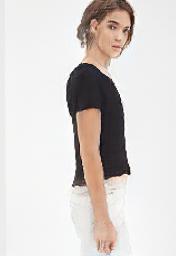}&
    \includegraphics[width=0.08\columnwidth]{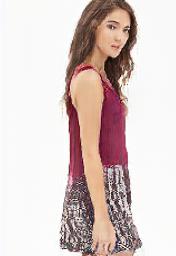}&
    \includegraphics[width=0.08\columnwidth]{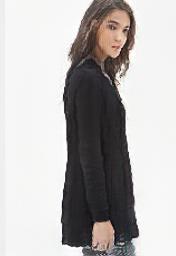}&
    \includegraphics[width=0.08\columnwidth]{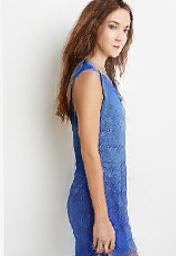}&
    \includegraphics[width=0.08\columnwidth]{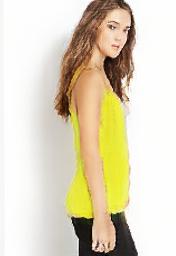}&
    \includegraphics[width=0.08\columnwidth]{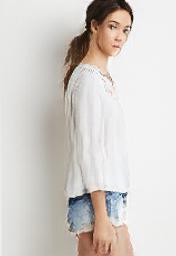}&
    \includegraphics[width=0.08\columnwidth]{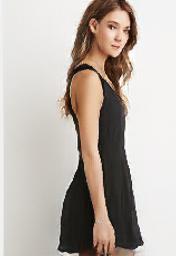}&
    \includegraphics[width=0.08\columnwidth]{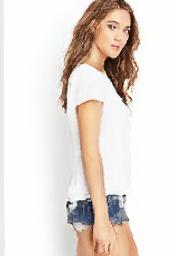}&
    \includegraphics[width=0.08\columnwidth]{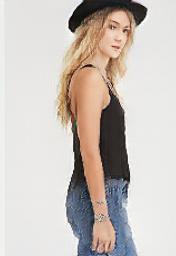}\\
\end{tabular}
}
\caption{\textbf{Our results of pose to image synthesis (DeepFashion dataset).} In each group, the first row shows exemplars, and the second row shows the pose images along with our results.}
\label{figure:deepfashion_0_result}
\end{figure*}

\clearpage
\FloatBarrier
\noindent\textbf{Pose-to-body}~~ Figure~\ref{figure:deepfashion_0_result} shows more pose synthesis results on DeepFashion dataset.

\clearpage
\subsection{Additional Results of Dense Correspondence}
\begin{figure*}[h!]
\center
\small
\setlength\tabcolsep{1pt}
{
\renewcommand{\arraystretch}{0.6}
\begin{tabular}{@{}cccccccc@{}}
    $y_B$& $x_A$& $r_{y\to x}$ & & $y_B$& $x_A$& $r_{y\to x}$\\
    \includegraphics[width=0.125\columnwidth]{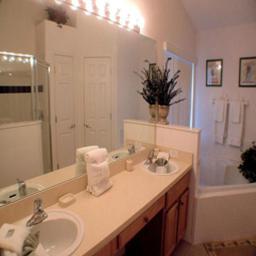}&
    \includegraphics[width=0.125\columnwidth]{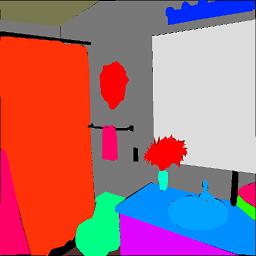}&
    \includegraphics[width=0.125\columnwidth]{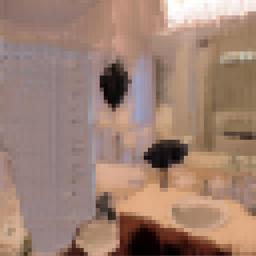}&~~&
    \includegraphics[width=0.125\columnwidth]{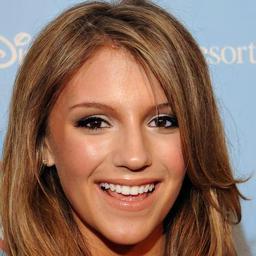}&
    \includegraphics[width=0.125\columnwidth]{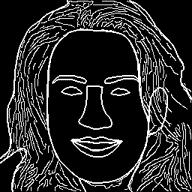}&
    \includegraphics[width=0.125\columnwidth]{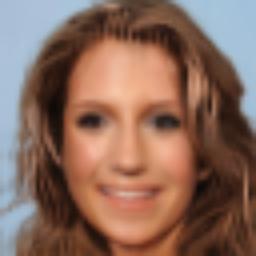}\\

    \includegraphics[width=0.125\columnwidth]{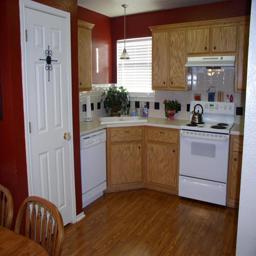}&
    \includegraphics[width=0.125\columnwidth]{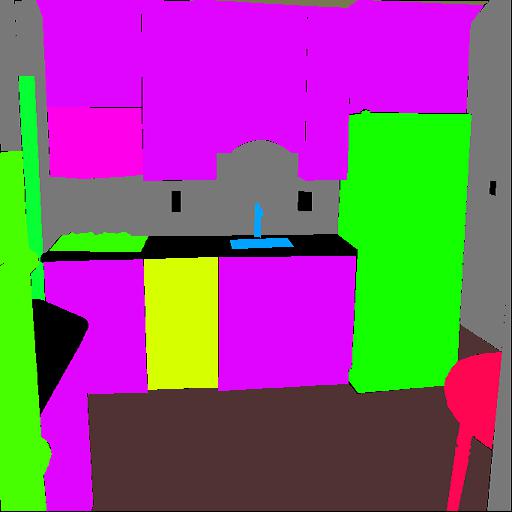}&
    \includegraphics[width=0.125\columnwidth]{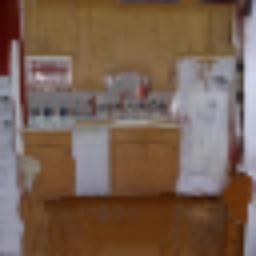}&~~&
    \includegraphics[width=0.125\columnwidth]{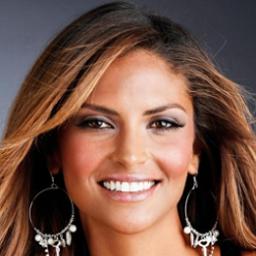}&
    \includegraphics[width=0.125\columnwidth]{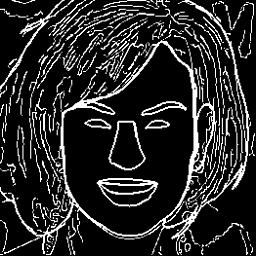}&
    \includegraphics[width=0.125\columnwidth]{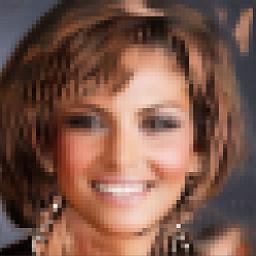}\\
    
    \includegraphics[width=0.125\columnwidth]{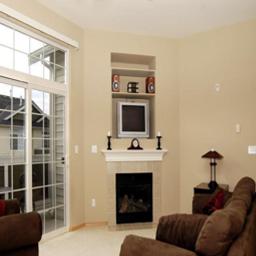}&
    \includegraphics[width=0.125\columnwidth]{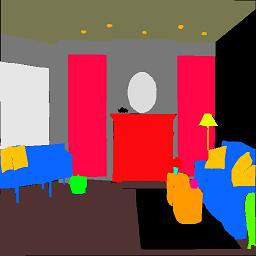}&
    \includegraphics[width=0.125\columnwidth]{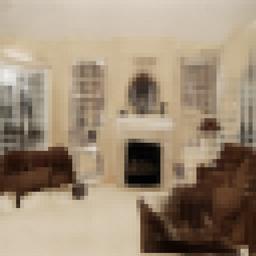}&~~&
    \includegraphics[width=0.125\columnwidth]{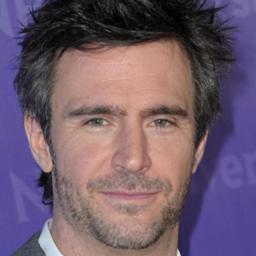}&
    \includegraphics[width=0.125\columnwidth]{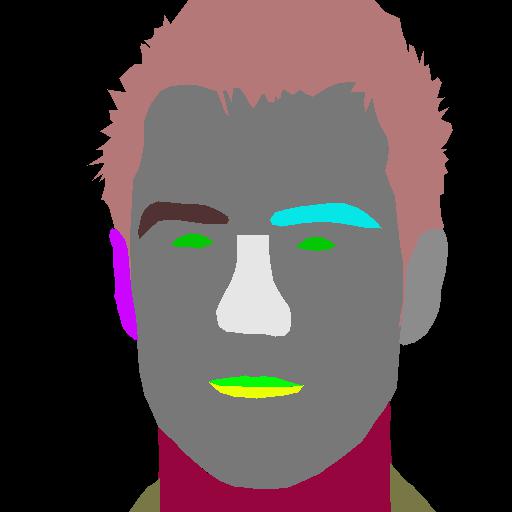}&
    \includegraphics[width=0.125\columnwidth]{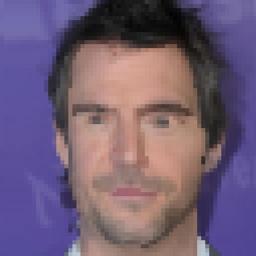}\\
    
    \includegraphics[width=0.125\columnwidth]{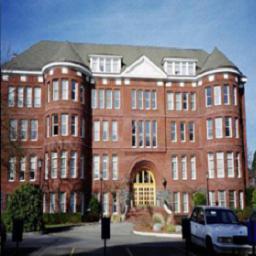}&
    \includegraphics[width=0.125\columnwidth]{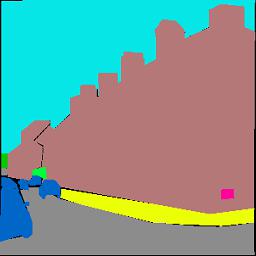}&
    \includegraphics[width=0.125\columnwidth]{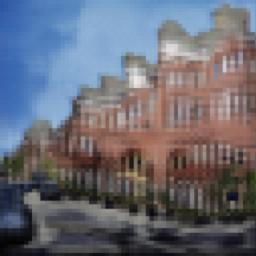}&~~&
    \includegraphics[width=0.125\columnwidth]{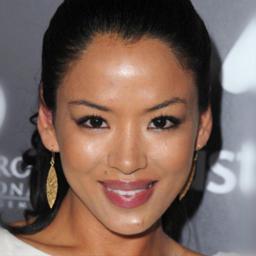}&
    \includegraphics[width=0.125\columnwidth]{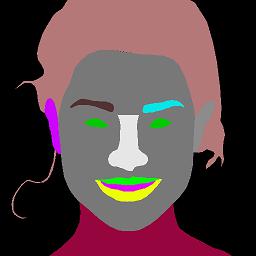}&
    \includegraphics[width=0.125\columnwidth]{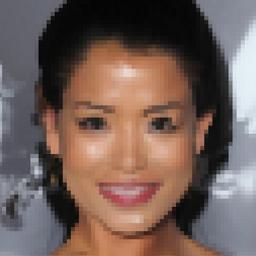}\\
    
    \includegraphics[width=0.125\columnwidth]{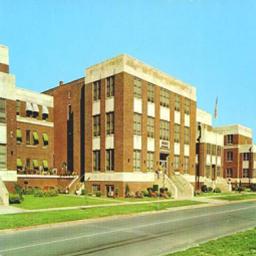}&
    \includegraphics[width=0.125\columnwidth]{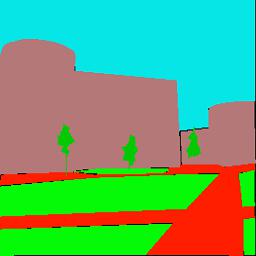}&
    \includegraphics[width=0.125\columnwidth]{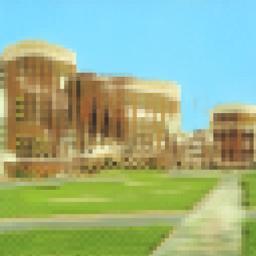}&~~&
    \includegraphics[width=0.125\columnwidth]{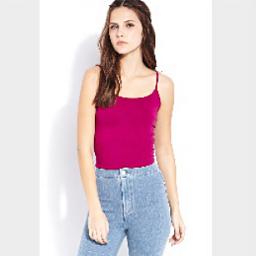}&
    \includegraphics[width=0.125\columnwidth]{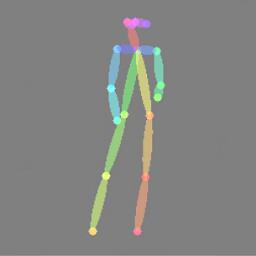}&
    \includegraphics[width=0.125\columnwidth]{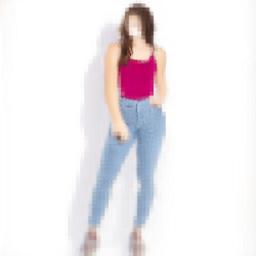}\\
    
    \includegraphics[width=0.125\columnwidth]{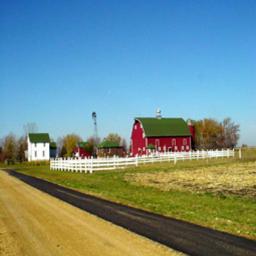}&
    \includegraphics[width=0.125\columnwidth]{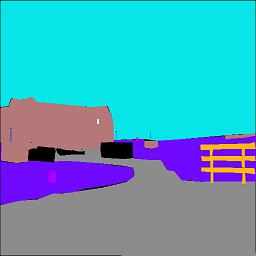}&
    \includegraphics[width=0.125\columnwidth]{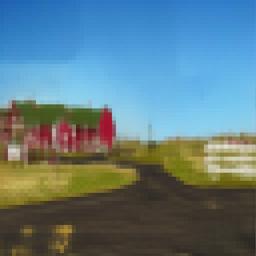}&~~&
    \includegraphics[width=0.125\columnwidth]{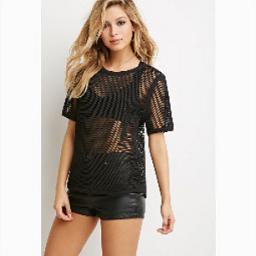}&
    \includegraphics[width=0.125\columnwidth]{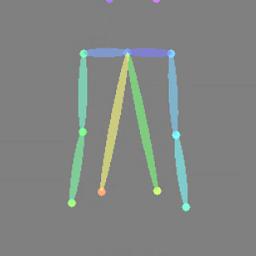}&
    \includegraphics[width=0.125\columnwidth]{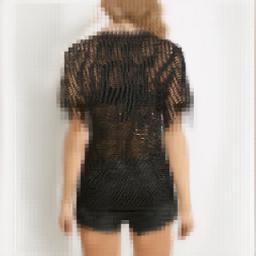}\\
\end{tabular}
}
\caption{\textbf{Warping according to the dense correspondence.} The warped image $r_{y\to x}$ is semantically aligned as the image in domain A.}
\label{figure:corr_result}
\end{figure*}

The proposed \emph{CoCosNet} is able to establish the dense correspondence between different domains. Figure~\ref{figure:corr_result} shows the dense warping results from domain B to domain A according to the correspondence ($r_{y\to x}$ in Equation 4).

\clearpage
\subsection{Additional Ablation Studies}
\begin{figure*}[h!]
\center
\small
\setlength\tabcolsep{1pt}
{
\renewcommand{\arraystretch}{0.6}
\begin{tabular}{@{}cccc@{}}
    Input & Exemplar& \makecell[c]{Batch\\normalization} & \makecell[c]{Positional\\normalization}\\
    \includegraphics[width=0.14\columnwidth]{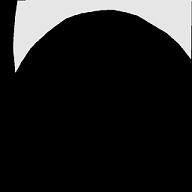}&
    \includegraphics[width=0.14\columnwidth]{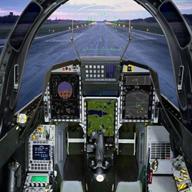} &
    \includegraphics[width=0.14\columnwidth]{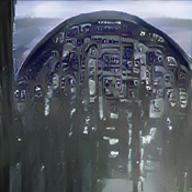}&
    \includegraphics[width=0.14\columnwidth]{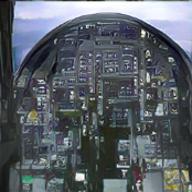}\\

    \includegraphics[width=0.14\columnwidth]{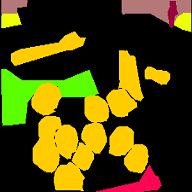}&
    \includegraphics[width=0.14\columnwidth]{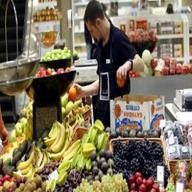} &
    \includegraphics[width=0.14\columnwidth]{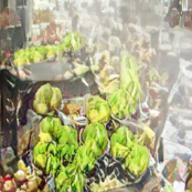}&
    \includegraphics[width=0.14\columnwidth]{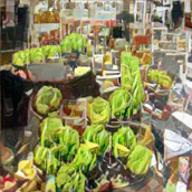}\\
\end{tabular}
}
\caption{\textbf{Positional Normalization vs. Batch Normalization.} The positional normalization significantly improves the \emph{lower bound} of the translation image quality.}
\label{figure:bn_pn}
\end{figure*}

\noindent\textbf{Positional Normalization} In the translation sub-network, we empirically find the normalization that computes the statistics at each spatial position better preserves the structure information synthesized in prior layers. Such positional normalization significantly improves the lower bound of our approach. We show the \emph{worst case} result of ADE20k dataset in  Figure~\ref{figure:bn_pn}, where the normalization helps produce vibrant image even when the correspondence is hard to be established in the complex scene.

\begin{figure*}[h!]
\center
\small
\setlength\tabcolsep{1pt}
{
\renewcommand{\arraystretch}{0.6}
\begin{tabular}{@{}cccc@{}}
    Input & Exemplar & \makecell[c]{Feature-wise \\ normalization} & \makecell[c]{Channel-wise \\ normalization} \\
    \includegraphics[width=0.14\columnwidth]{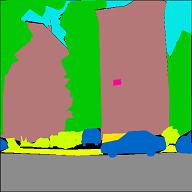}&
    \includegraphics[width=0.14\columnwidth]{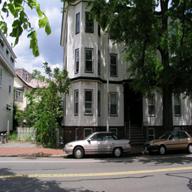}&
    \includegraphics[width=0.14\columnwidth]{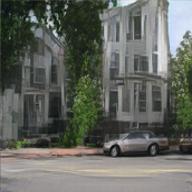}&
    \includegraphics[width=0.14\columnwidth]{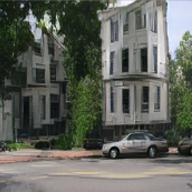}\\

     \includegraphics[width=0.14\columnwidth]{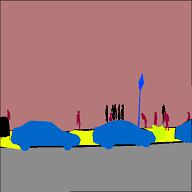}&
    \includegraphics[width=0.14\columnwidth]{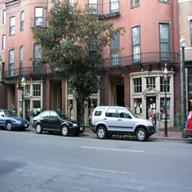}&
    \includegraphics[width=0.14\columnwidth]{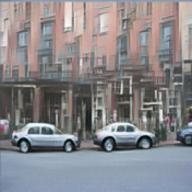}&
    \includegraphics[width=0.14\columnwidth]{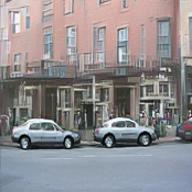}\\
\end{tabular}
}
\caption{\textbf{Channel-wise normalization during correspondence.} The channel-wise normalization helps transfer the window structures from the exemplar image.}
\label{figure:normalization_dimension}
\end{figure*}

\noindent\textbf{Feature normalization for correspondence}~~
Note that we normalize the features before computing the correlation matrix. Likewise, we propose to calculate the statistics along the channel dimension while keeping the spatial size as the feature maps. This helps transfer the fine structures in the exemplar. As shown in Figure~\ref{figure:normalization_dimension}, the channel-wise normalization betters maintains the window structures in the ultimate output.


\clearpage
\subsection{Additional Application Results}

\begin{figure*}[h!]
\center
\small
\setlength\tabcolsep{1pt}
{
\renewcommand{\arraystretch}{0.8}
\begin{tabular}{@{}ccccc@{}}
    Input&Edit 1&Edit 2&Edit 3&Edit 4\\
    \includegraphics[width=0.16\columnwidth]{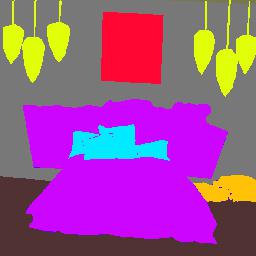}&
    \includegraphics[width=0.16\columnwidth]{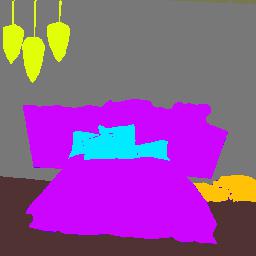}&
    \includegraphics[width=0.16\columnwidth]{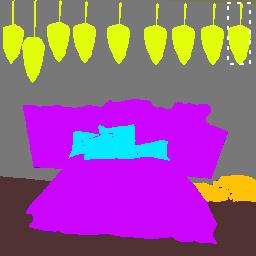}&
    \includegraphics[width=0.16\columnwidth]{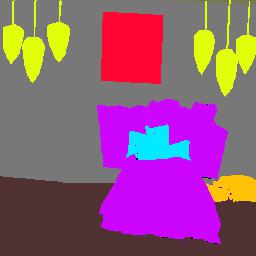}&
    \includegraphics[width=0.16\columnwidth]{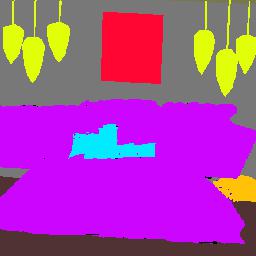}\\

    \includegraphics[width=0.16\columnwidth]{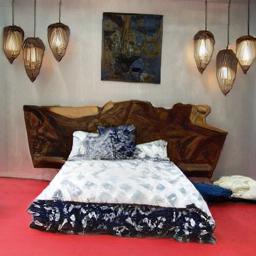}&
    \includegraphics[width=0.16\columnwidth]{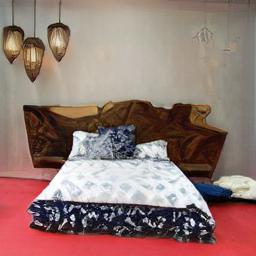}&
    \includegraphics[width=0.16\columnwidth]{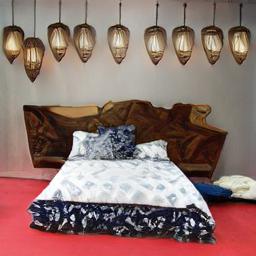}&
    \includegraphics[width=0.16\columnwidth]{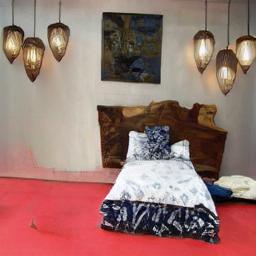}&
    \includegraphics[width=0.16\columnwidth]{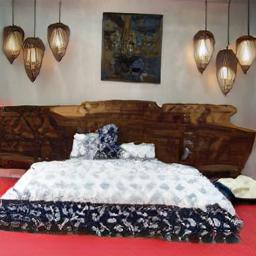}\\
\end{tabular}
}
\caption{\textbf{Image editing.} Giving the original input image along with segmentation mask (1st column), we manipulate the image by changing its semantic layout (2nd-5th columns).}
\label{figure:edit_result}
\end{figure*}

\noindent\textbf{Image editing}~~We show another example of the semantic image editing in Figure~\ref{figure:edit_result}, where we manipulate the instances in the image by modifying their segmentation masks.

\begin{figure*}[h!]
\center
\small
\setlength\tabcolsep{0pt}
{
\renewcommand{\arraystretch}{0.0}
\begin{tabular}{@{}cccccccc@{}}
    \includegraphics[width=0.125\columnwidth]{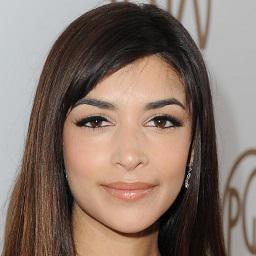}&
    \includegraphics[width=0.125\columnwidth]{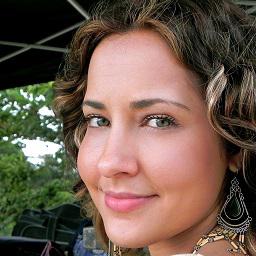}&
    \includegraphics[width=0.125\columnwidth]{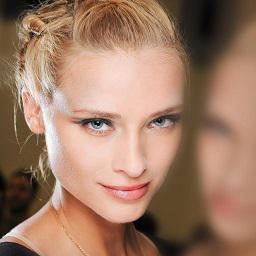}&
    \includegraphics[width=0.125\columnwidth]{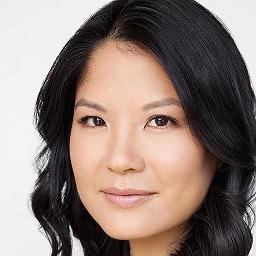}&
    \includegraphics[width=0.125\columnwidth]{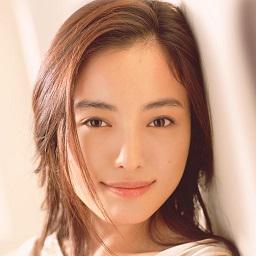}&
    \includegraphics[width=0.125\columnwidth]{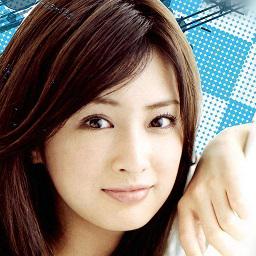}&
    \includegraphics[width=0.125\columnwidth]{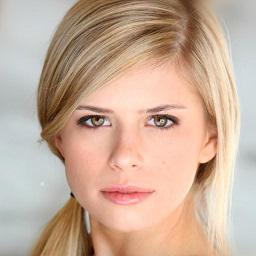}&
    \includegraphics[width=0.125\columnwidth]{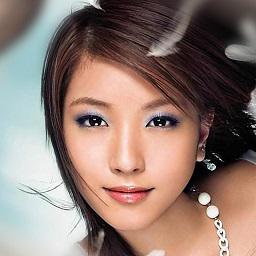}\\

    \includegraphics[width=0.125\columnwidth]{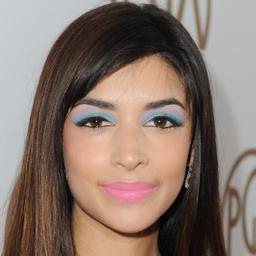}&
    \includegraphics[width=0.125\columnwidth]{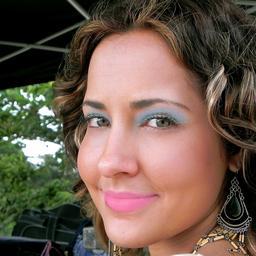}&
    \includegraphics[width=0.125\columnwidth]{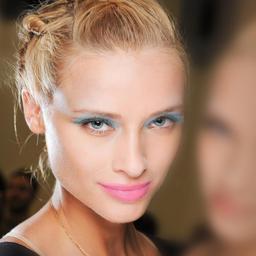}&
    \includegraphics[width=0.125\columnwidth]{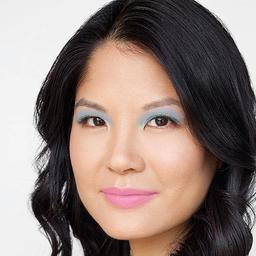}&
    \includegraphics[width=0.125\columnwidth]{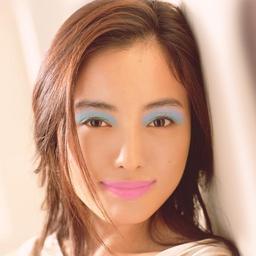}&
    \includegraphics[width=0.125\columnwidth]{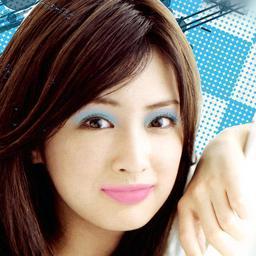}&
    \includegraphics[width=0.125\columnwidth]{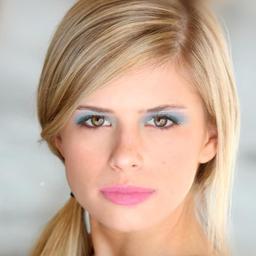}&
    \includegraphics[width=0.125\columnwidth]{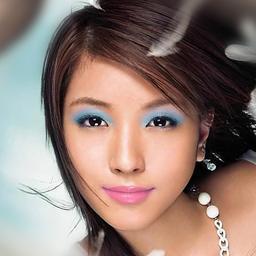}\\[0.5em]
    
    \includegraphics[width=0.125\columnwidth]{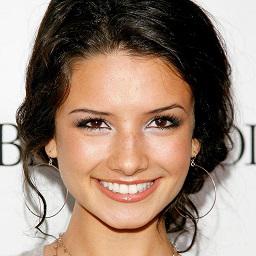}&
    \includegraphics[width=0.125\columnwidth]{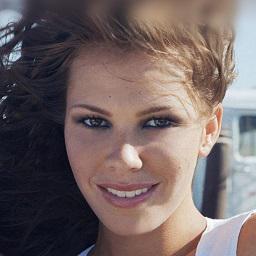}&
    \includegraphics[width=0.125\columnwidth]{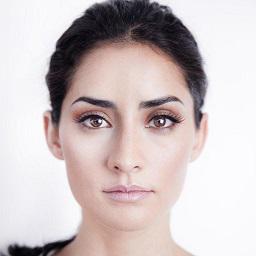}&
    \includegraphics[width=0.125\columnwidth]{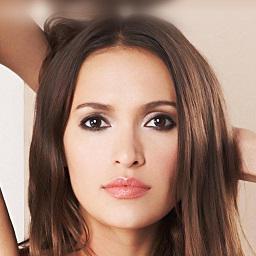}&
    \includegraphics[width=0.125\columnwidth]{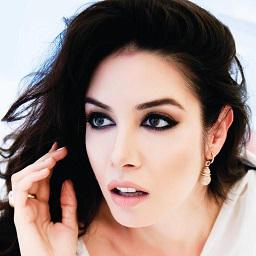}&
    \includegraphics[width=0.125\columnwidth]{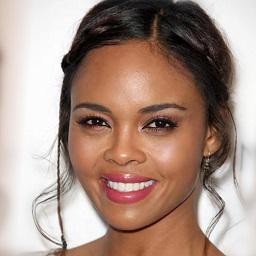}&
    \includegraphics[width=0.125\columnwidth]{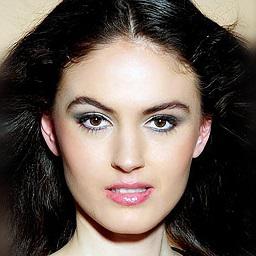}&
    \includegraphics[width=0.125\columnwidth]{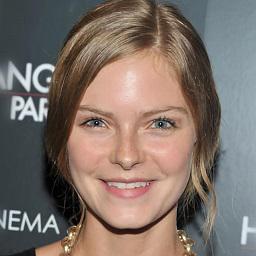}\\

    \includegraphics[width=0.125\columnwidth]{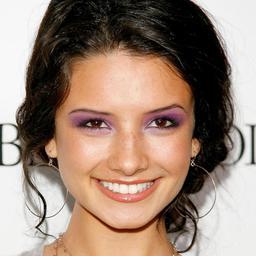}&
    \includegraphics[width=0.125\columnwidth]{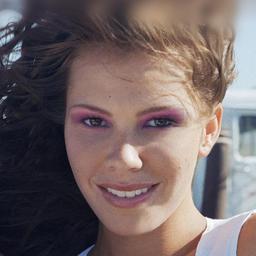}&
    \includegraphics[width=0.125\columnwidth]{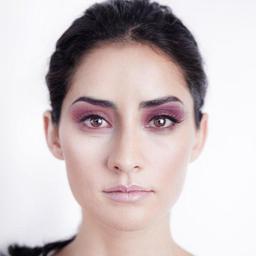}&
    \includegraphics[width=0.125\columnwidth]{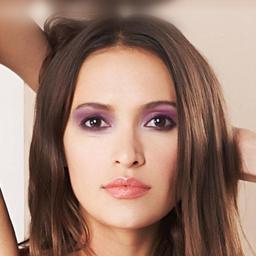}&
    \includegraphics[width=0.125\columnwidth]{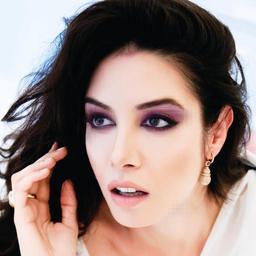}&
    \includegraphics[width=0.125\columnwidth]{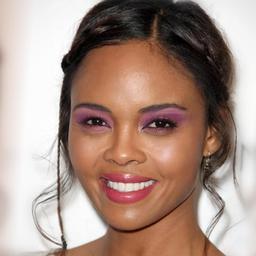}&
    \includegraphics[width=0.125\columnwidth]{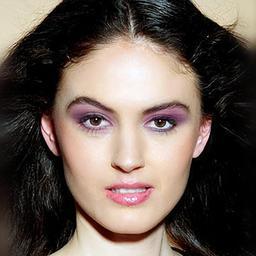}&
    \includegraphics[width=0.125\columnwidth]{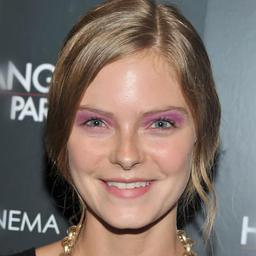}\\
\end{tabular}
}
\caption{\textbf{Makeup transfer.} Given a portrait along with makeup edits (1st column), we can transfer the makeup to other portraits by matching the semantic correspondence.}
\label{figure:makeup_result}
\end{figure*}

\noindent\textbf{Makeup transfer}~~Thanks to the dense semantic correspondence, we can transfer the makeup brushes to a batch of portraits. Figure~\ref{figure:makeup_result} gives more supplementary results.  

\clearpage
\subsection{Implementation Details}
The detailed architecture of \emph{CoCosNet} is shown in Table~\ref{tab:architecture of CoCosNet}, with the naming convention as the CycleGAN. 

\vspace{0.4em}
\noindent\textbf{Cross-domain correspondence network}~~
Two domain adaptors without weight sharing are used to adapt the input image and the exemplar to a shared domain $S$. The domain adaptors comprise several Conv-InstanceNorm-LeakReLU blocks and the spatial size of features in $S$ is 64$\times$64. Once the intermediate domain $S$ is found, a shared adaptive feature block further transforms the features from two branches to the representation suitable for correspondence. The correlation layer computes pairwise affinity values between 4096$\times$1 normalized features vectors. We downscale the exemplar image to 64$\times$64 to fit the size of correlation matrix, and thus obtain the warped image on this scale. We use synchronous batch normalization within this sub-network. 

\vspace{0.4em}
\noindent\textbf{Translation network}~~
The translation network generates the final output based on the style of the warped exemplar. We encode the exemplar style through two convolutional layers, which outputs $\alpha_i$ and $\beta_i$ to modulate the normalization layer in the generator network. We have seven such style encoder, each responsible for modulating an individual normalization layer. The generator consists of seven normalization layer, which progressively utilizes the style code to synthesize the final output. The generator also employs a nonlocal block so that a larger receptive field can be utilized to enhance the global structural consistency. We use positional normalization within this sub-network.

\noindent\textbf{Warm-up strategy}~~
For the most challenging ADE20k dataset, a mask warm strategy is used. At the beginning of the training, we explicitly provide the segmentation mask for the domain adaptors, and employ cross-entropy loss to encourage that the masks are correctly aligned after dense warping. Such warm-up helps speed up the convergence of the correspondence network and improve the correspondence accuracy. 
After training 80 epochs, we replace the segmentation masks with Gaussian noise. We just use the segmentation mask for warm up and there is no need to provide the masks during inference.

\begin{table}[!tbh]
  \small
  \begin{center}
    \caption{\textbf{The architecture of \emph{CoCosNet}.} k3s1 indicates the convolutional layer with kernel size 3 and stride 1. The $i$th style encoder outputs features with dimensions matching the $i$th Resblock in the generator.}
    \label{tab:architecture of CoCosNet}
    \begin{tabular}{@{}l|l|l|l@{}}
      \toprule
      {Sub-network} & {Module} & {Layers in the module} & {Output shape (H$\times$W$\times$C)}\\
      \midrule
      \multirow{7}{2.5cm}{Correspondence Network} 
      & \multirow{6}{3.0cm}{Domain adaptor$\times2$} & Conv2d / k3s1 & 256$\times$256$\times$64\\ 
      & & Conv2d / k4s2 & 128$\times$128$\times$128\\
      & & Conv2d / k3s1 & 128$\times$128$\times$256\\
      & & Conv2d / k4s2 & 64$\times$64$\times$512\\
      & & Conv2d / k3s1 & 64$\times$64$\times$512\\
      & & Resblock$\times3$ / k3s1 & 64$\times$64$\times$256\\
      \cmidrule{2-4}
      & \multirow{2}{3.0cm}{Adaptive feature block} & Resblock$\times4$ & 64$\times$64$\times$256\\
      & & Conv2d / k1s1 & 64$\times$64$\times$256\\
      \cmidrule{2-4}
      &Correspondence&Correlation\&warping & 64$\times$64$\times$3\\
      \midrule
      \multirow{7}{2.5cm}{Translation Network}
      & \multirow{3}{3.0cm}{Style encoder$\times7$} & Bilinear interpolation & $h^i \times w^i \times$3\\ 
      & & Conv2d / k3s1 & $h^i \times w^i \times$128\\
      & & Conv2d / k3s1 & $h^i \times w^i \times c^i$\\
      \cmidrule{2-4}
      & \multirow{5}{3.0cm}{Generator} & Conv2d / k3s1 & 8$\times$8$\times$1024\\ 
      &  & Resblock$\times5$ & 128$\times$128$\times$256\\
      &  & Nonlocal & 128$\times$128$\times$256\\
      &  & Resblock$\times2$ & 256$\times$256$\times$64\\
      &  & Conv2d / k3s1 & 256$\times$256$\times$3\\
      \bottomrule
    \end{tabular}
  \end{center}
\end{table}

\clearpage
\subsection{Detailed User Study Results}
\begin{figure*}[h!]
\center
\small
\setlength\tabcolsep{2pt}
{
\renewcommand{\arraystretch}{8.0}
\begin{tabular}{@{}cc@{}}
    \includegraphics[width=0.45\columnwidth]{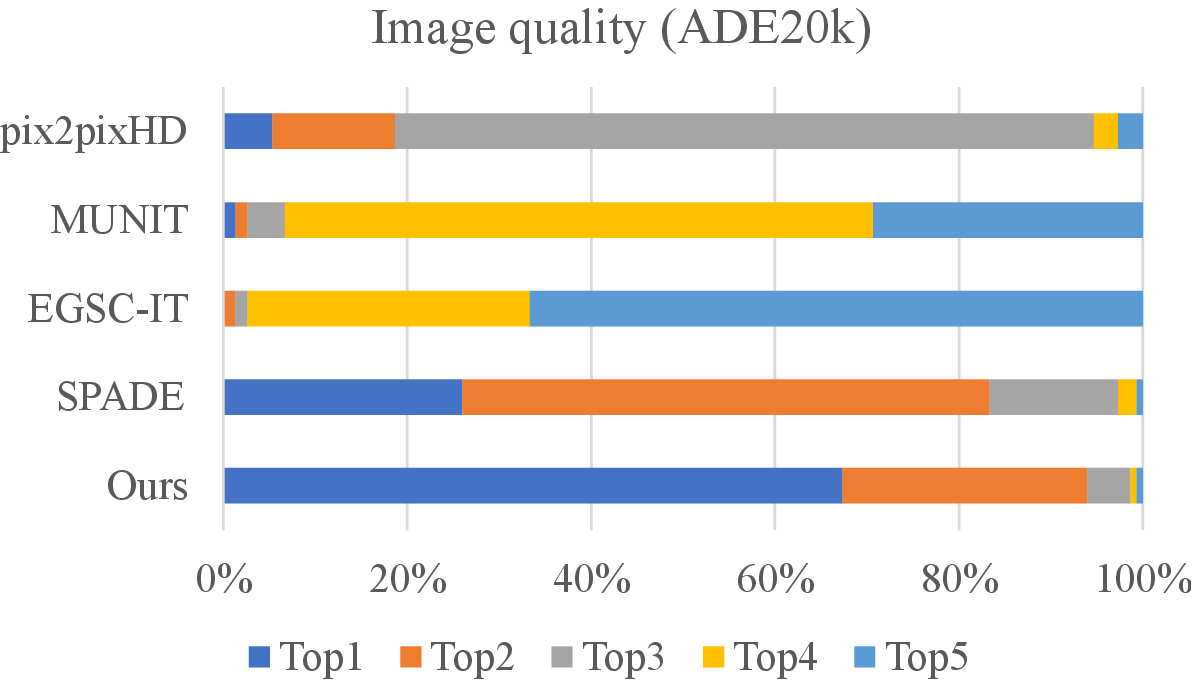}&
    \includegraphics[width=0.45\columnwidth]{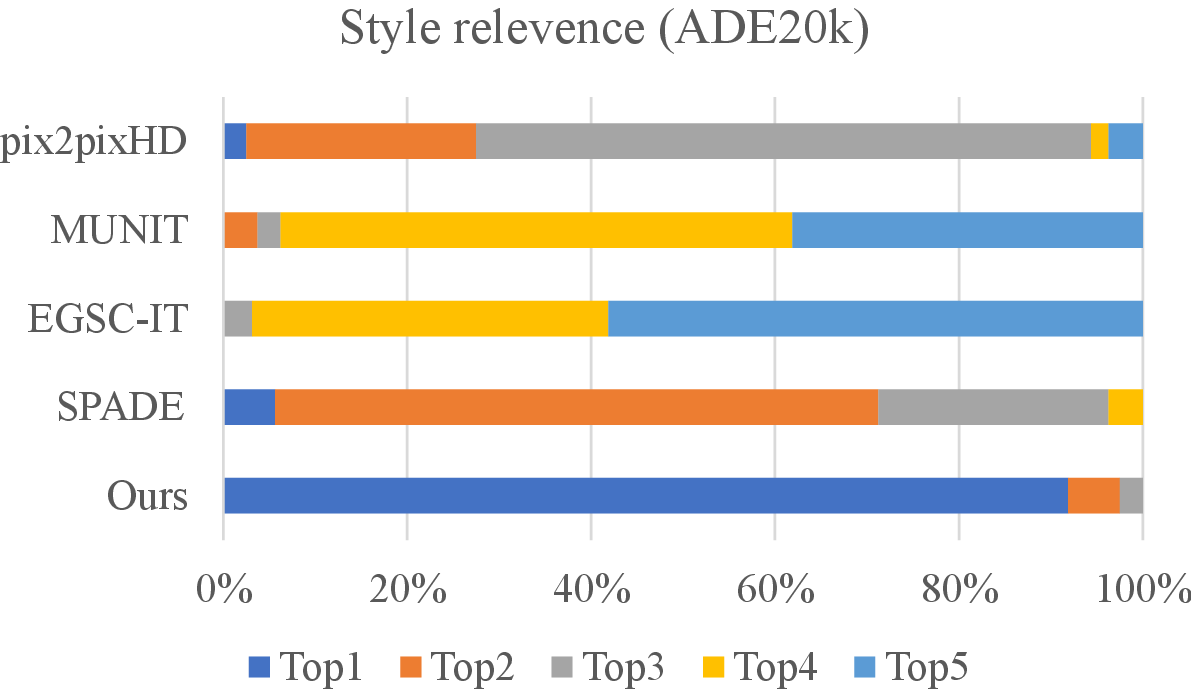}\\
    
    \includegraphics[width=0.45\columnwidth]{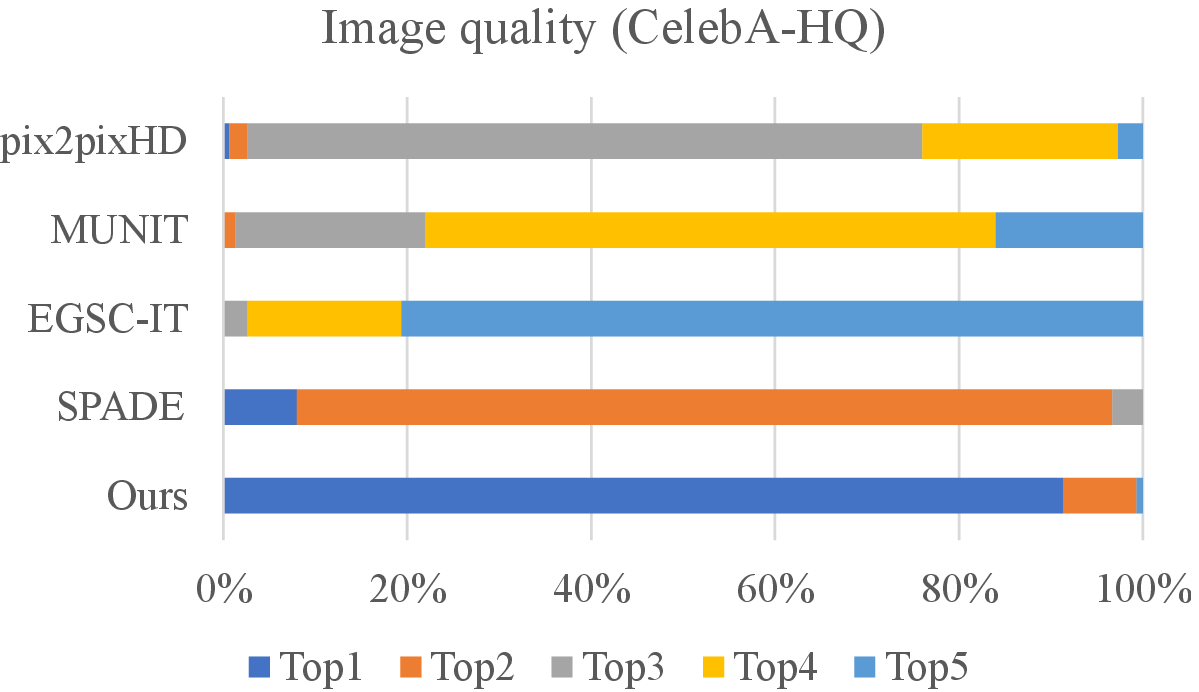}&
    \includegraphics[width=0.45\columnwidth]{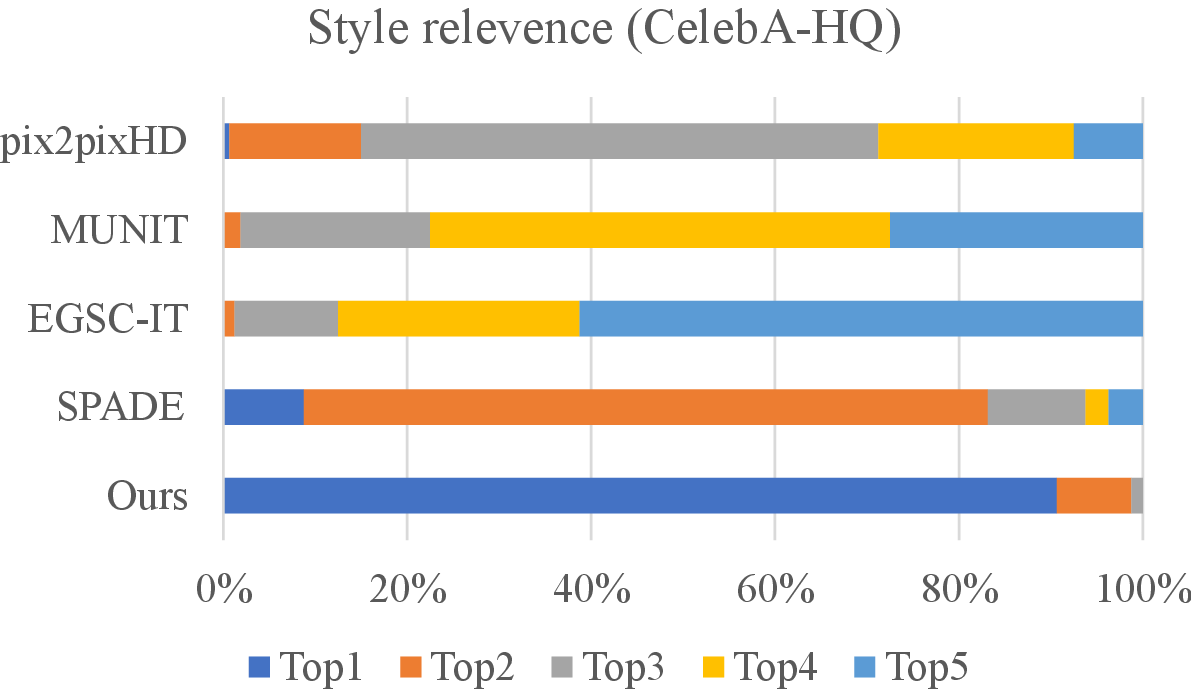}\\
    
    \includegraphics[width=0.45\columnwidth]{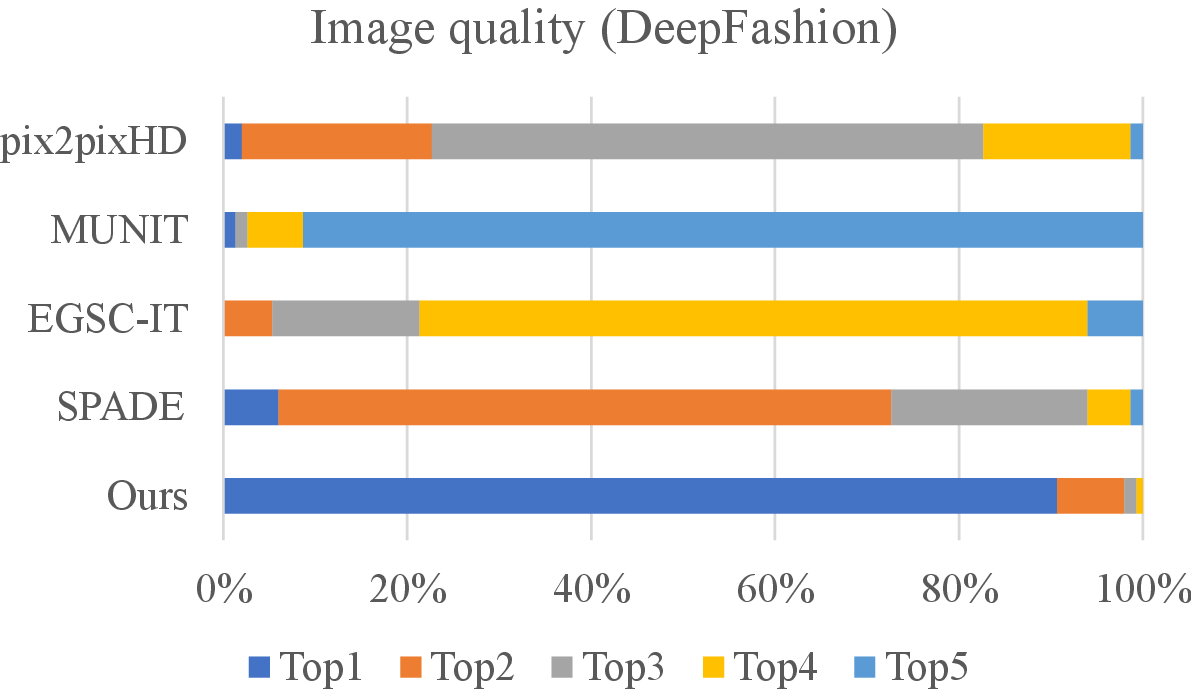}&
    \includegraphics[width=0.45\columnwidth]{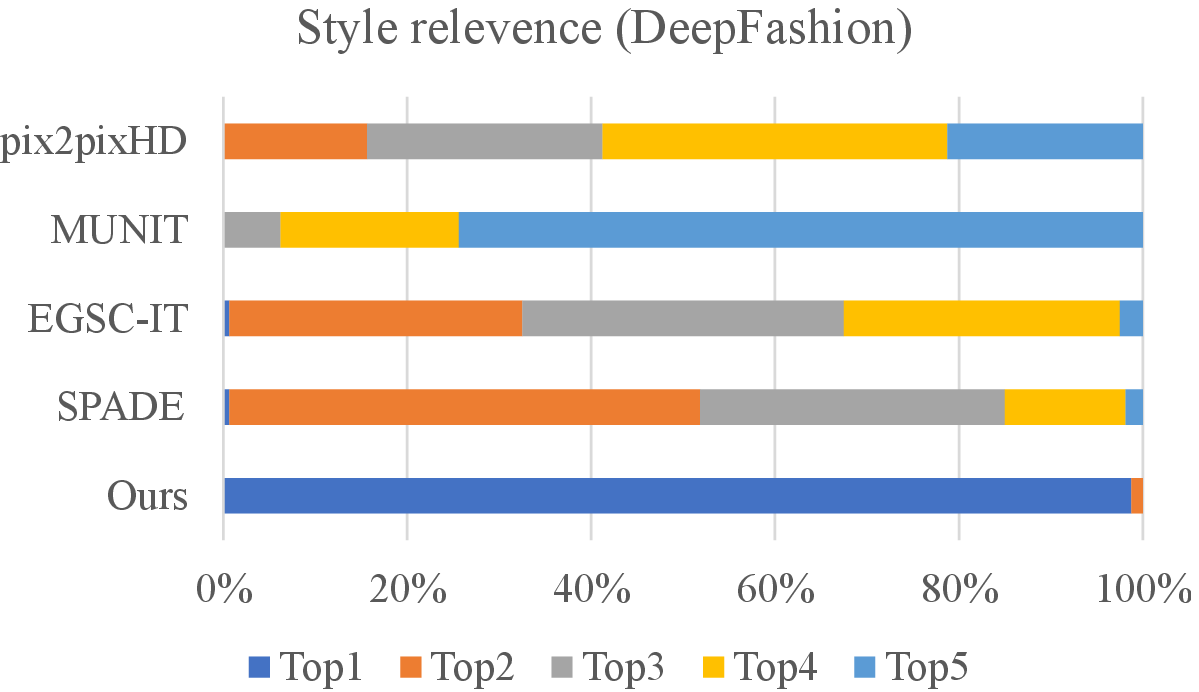}\\
\end{tabular}
}
\caption{\textbf{Detailed user study results for ADE20k, CelebA-HQ and DeepFashion dataset.}}
\label{figure:userstudy_result}
\end{figure*}

Figure~\ref{figure:userstudy_result} shows the detailed results of user study. In ADE20k, there are 67.3\%  and 91.9\% users respectively that prefer the image quality and style relevance for our method. Regarding edge-to-face translation on CelebA-HQ, 91.3\% users prefer our image quality while 90.6\% users believes our method most resembles the exemplar. For pose synthesis on DeepFashion dataset, 90.6\% and 98.8\% users prefer our results  according to the image quality and the style resemblance respectively.

\clearpage
\subsection{Multimodal results for Flickr dataset}
\begin{figure}[h!]
    \center
    \small
    \setlength\tabcolsep{2pt}
    {
    \renewcommand{\arraystretch}{1.0}
    \begin{tabular}{@{}ccccc@{}}
        \includegraphics[width=0.15\columnwidth]{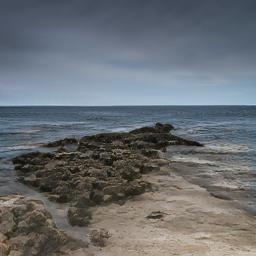}&
        \includegraphics[width=0.15\columnwidth]{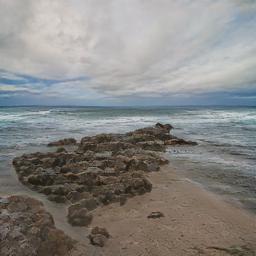}&
        \includegraphics[width=0.15\columnwidth]{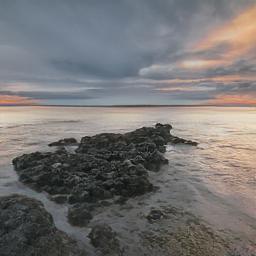}&
        \includegraphics[width=0.15\columnwidth]{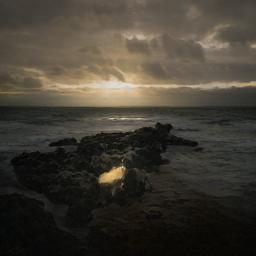}&
        \includegraphics[width=0.15\columnwidth]{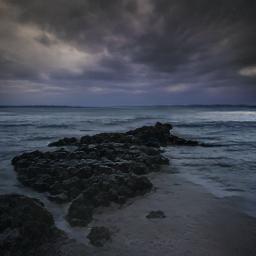}\\
        
        \includegraphics[width=0.15\columnwidth]{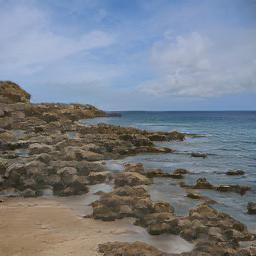}&
        \includegraphics[width=0.15\columnwidth]{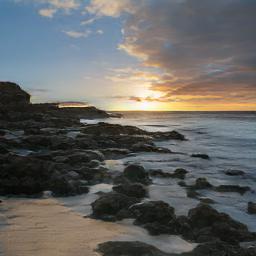}&
        \includegraphics[width=0.15\columnwidth]{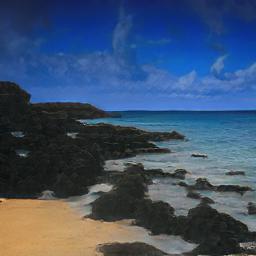}&
        \includegraphics[width=0.15\columnwidth]{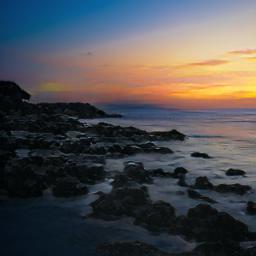}&
        \includegraphics[width=0.15\columnwidth]{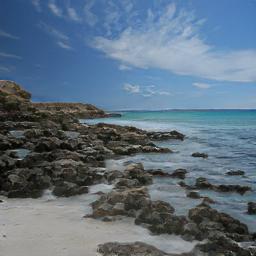}\\
        
        \includegraphics[width=0.15\columnwidth]{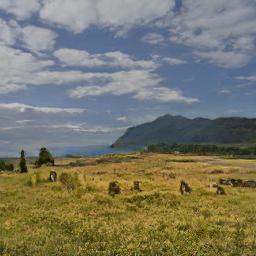}&
        \includegraphics[width=0.15\columnwidth]{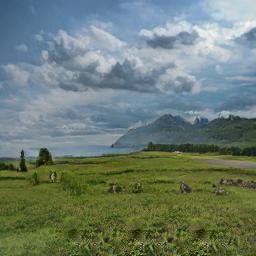}&
        \includegraphics[width=0.15\columnwidth]{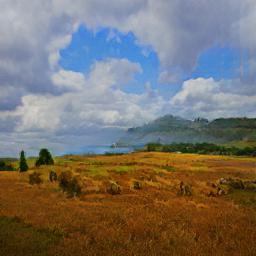}&
        \includegraphics[width=0.15\columnwidth]{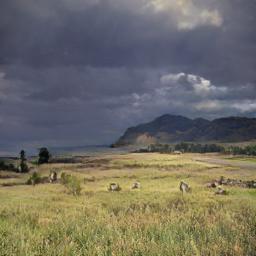}&
        \includegraphics[width=0.15\columnwidth]{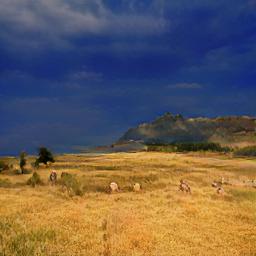}\\
        
        \includegraphics[width=0.15\columnwidth]{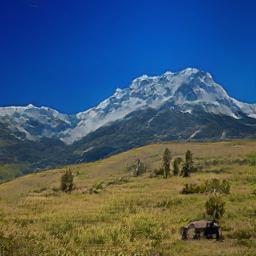}&
        \includegraphics[width=0.15\columnwidth]{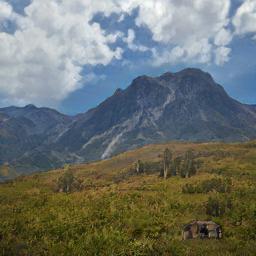}&
        \includegraphics[width=0.15\columnwidth]{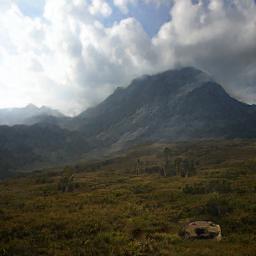}&
        \includegraphics[width=0.15\columnwidth]{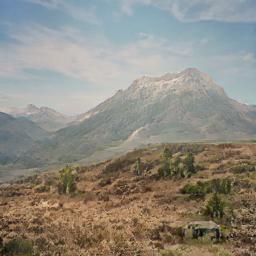}&
        \includegraphics[width=0.15\columnwidth]{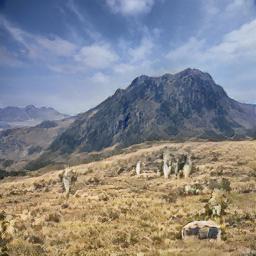}\\
        
        \includegraphics[width=0.15\columnwidth]{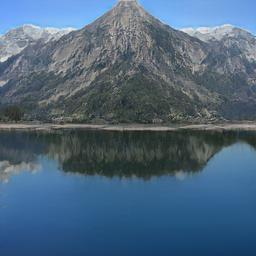}&
        \includegraphics[width=0.15\columnwidth]{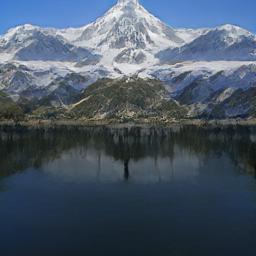}&
        \includegraphics[width=0.15\columnwidth]{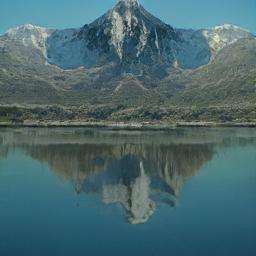}&
        \includegraphics[width=0.15\columnwidth]{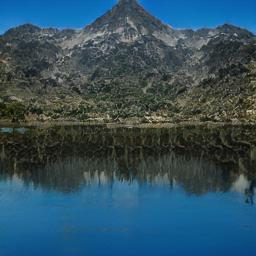}&
        \includegraphics[width=0.15\columnwidth]{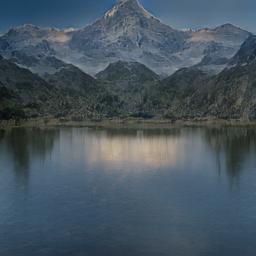}\\
        
        \includegraphics[width=0.15\columnwidth]{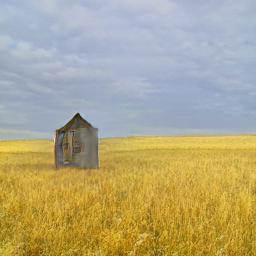}&
        \includegraphics[width=0.15\columnwidth]{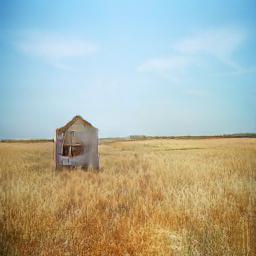}&
        \includegraphics[width=0.15\columnwidth]{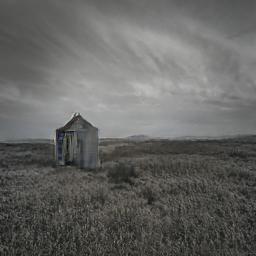}&
        \includegraphics[width=0.15\columnwidth]{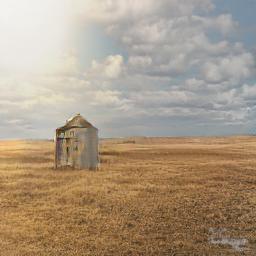}&
        \includegraphics[width=0.15\columnwidth]{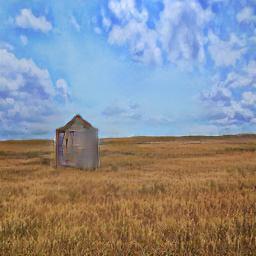}\\
    \end{tabular}
    }
    \caption{\textbf{Multimodal results of Flickr dataset.} We only present the final synthesis results here.}
    \label{figure:multimodal}
\end{figure}

Similar to the practice in~\cite{park2019semantic}, we collect 56,568 landscape images from Flickr. The semantic segmentation masks are computed using a pre-trained  UPerNet101~\cite{xiao2018unified} network. By feeding different exemplar, our method supports multimodal landscape synthesis. Figure~\ref{figure:multimodal} shows highly realistic landscape results using the images in Flicker dataset. 

\clearpage
\subsection{Limitation}
\begin{figure*}[h!]
\center
\small
\setlength\tabcolsep{2pt}
{
\renewcommand{\arraystretch}{1.0}
\begin{tabular}{@{}ccc@{}}
    Input & Exemplar & Synthesis \\
    \includegraphics[width=0.15\columnwidth]{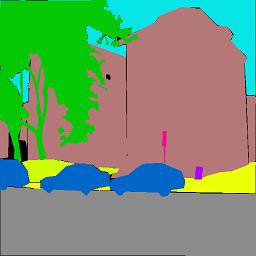}&
    \includegraphics[width=0.15\columnwidth]{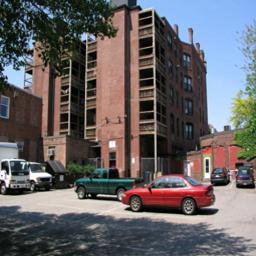}&
    \includegraphics[width=0.15\columnwidth]{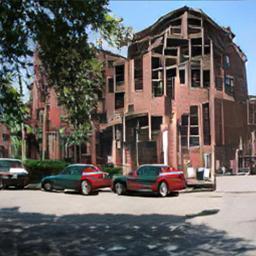}\\
    \includegraphics[width=0.15\columnwidth]{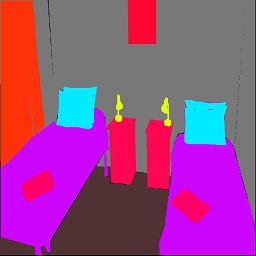}&
    \includegraphics[width=0.15\columnwidth]{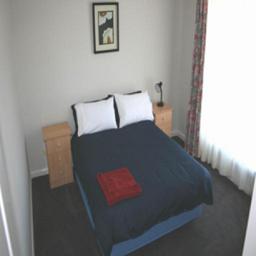}&
    \includegraphics[width=0.15\columnwidth]{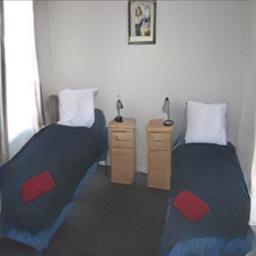}\\
\end{tabular}
}
\caption{\textbf{Limitation.} Our method may produce mixed color artifact due the one-to-many mapping (1st row). Besides, the multiple instances (pillows in the figure) may use the same style in the cases of many-to-one mapping (2nd row).}
\label{figure:multiple instances}
\end{figure*}

As an exemplar-based approach, our method may not produce satisfactory results due to one-to-many and many-to-one mappings as shown in Figure~\ref{figure:multiple instances}. We leave further research tackling these issues as future work.


Another limitation is that the computation of the correlation matrix takes tremendous GPU memory,
which makes our method hardly scale for high resolution images. We leave the solve of this issue in future work.


\end{document}